\documentclass{article}

\usepackage[final,nonatbib]{template}
\usepackage{multirow}

\usepackage[T1]{fontenc}
\usepackage{hyperref}
\hypersetup{hidelinks}
\usepackage{url}
\usepackage[backend=biber,style=numeric,sorting=none,natbib=true]{biblatex}
\AtBeginBibliography{\raggedright}
\usepackage{booktabs}
\usepackage{tabularx}
\usepackage{longtable}
\usepackage{amsfonts}
\usepackage{nicefrac}
\usepackage{microtype}
\usepackage[table]{xcolor}
\usepackage{amsmath}
\usepackage{graphicx}
\usepackage{tikz}
\usetikzlibrary{arrows.meta,positioning,fit,calc}
\usepackage[notransparent]{svg}
\svgpath{{figures/}}
\usepackage{float}
\usepackage{subcaption}
\usepackage{tablefootnote}
\usepackage{eso-pic}
\usepackage[most]{tcolorbox}
\usepackage{placeins}

\definecolor{abstractbg}{HTML}{F3F0FF}
\definecolor{abstractborder}{HTML}{C4B5FD}
\definecolor{abstracttitle}{HTML}{5B21B6}

\newtcolorbox{techreportabstract}{
    colback=abstractbg,
    colframe=abstractborder,
    arc=6pt,
    boxrule=1pt,
    left=18pt, right=18pt,
    top=15pt, bottom=15pt,
    halign=justify,
    before=\vspace{10pt},
    after=\vspace{15pt}
}

\newtcolorbox{insightbox}[1]{
    colback=blue!2,
    colframe=blue!35!black,
    colbacktitle=blue!35!black,
    coltitle=white,
    fonttitle=\bfseries,
    title={#1},
    arc=5pt,
    boxrule=0.8pt,
    left=9pt, right=9pt,
    top=7pt, bottom=7pt,
    before=\vspace{5pt},
    after=\vspace{7pt}
}

\usepackage{fancyhdr}
\usepackage[style=iso]{datetime2}

\fancypagestyle{firstpage}{
    \fancyhf{}
}

\AddToShipoutPictureBG*{%
  \AtPageUpperLeft{%
    \hspace{2.5cm}%
    \raisebox{-2.5cm}{%
      \includegraphics[width=1.0cm]{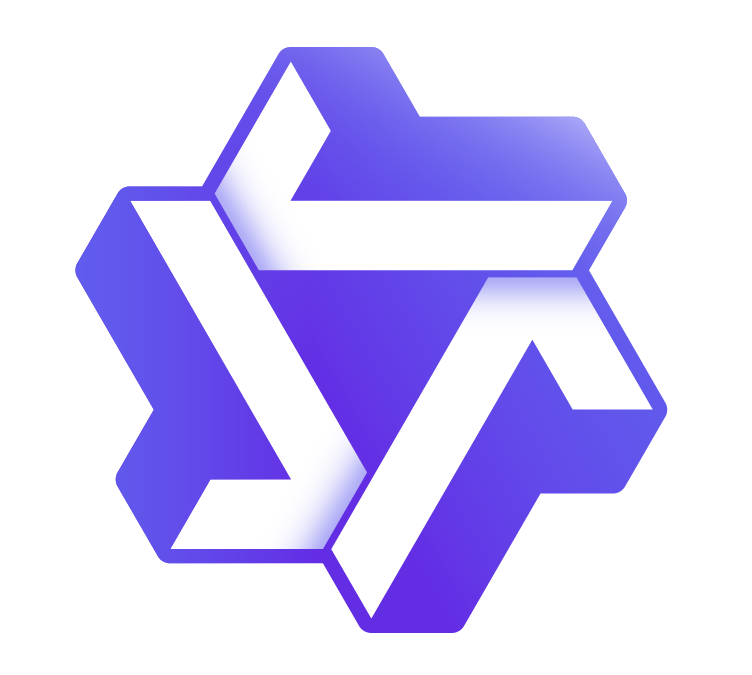}%
    }%
  }%
}

\title{The Evolution of Attention in Large Language Models:\\Mechanisms, Trade-offs, and Emerging Trends}

\author{
Zhentao Tan, Jingyi Shen, Yanbo Li, Yao Liu, Yue Wu, Jieping Ye \\
Alibaba Token Hub, Alibaba Group
}

\date{}

\begin{document}
\raggedbottom

\maketitle
\thispagestyle{firstpage}

\begin{techreportabstract}
    {\centering\large\bfseries Abstract\par}
    \vspace{0.8em}

Self-attention provides large language models with fine-grained, query-dependent access to contextual information, but dense token interactions incur quadratic prefill cost and require a key--value cache that grows with context length. Research has consequently evolved along several interacting directions, including explicit-memory compression, sparse access, recurrent state construction, structured state dynamics, and heterogeneous mechanism composition. This survey analyzes these developments through the shared notion of model-internal contextual memory. We introduce a five-dimensional analytical lens---Memory Representation, Memory Update, Access, Readout, and Integration---that distinguishes what information remains represented, how it changes, what is made eligible for a query, how eligible memory is read, and how one or more completed readouts are transformed or coordinated into the module output. This lens enables historically distinct and overlapping research lines to be compared without reducing them to a single computational model.

We use this perspective to reconstruct mechanism-level developments and examine their architectural adoption through a longitudinal inventory of 59 release-level records spanning 14 major model lineages, complemented by a frozen comparison of 11 high-performing open-weight model endpoints. A three-level synthesis emerges. At the mechanism level, explicit-memory and recurrent-state methods retain different memory interfaces, yet increasingly extend design control across an overlapping set of memory functions. At the architecture level, designs remain heterogeneous, but their coordination increasingly operates along network depth: layer-wise composition distributes complementary memory processing across representational stages, while emerging cross-layer reuse allows selected memory and routing artifacts to persist into later layers. Together, these developments make network depth an emerging dimension along which contextual memory is constructed and managed. Looking forward, they motivate a stateful multidimensional memory-routing hypothesis in which persistent memory is organized across temporal scope, network depth, substrate type, and representation granularity, with coordinated Sparse Write and Sparse Read governing what information is maintained and what stored information contributes to each query. Overall, efficient sequence architecture design is increasingly concerned with the joint organization, lifecycle, and selective use of contextual memory rather than the optimization of an isolated Attention operator.
\end{techreportabstract}

\clearpage
\setcounter{tocdepth}{2}
\tableofcontents
\clearpage

% !TEX root = ../main.tex
\section{Introduction}
\label{sec:introduction}

The Transformer replaced sequential recurrent propagation with self-attention, allowing each token to aggregate contextual information through direct, content-dependent interactions \citep{vaswani2017attention}. This mechanism enabled highly parallel training and established the computational foundation of modern large language models (LLMs) \citep{brown2020gpt3}. Its explicit representation of preceding tokens provides fine-grained, query-dependent access to context and supports in-context learning, long-range dependency modeling, and flexible reuse of information within a sequence. The same interaction pattern, however, creates the principal scaling limitations of dense attention. During prefill, the number of query-key comparisons grows quadratically with sequence length; during autoregressive decoding, the key-value (KV) cache grows linearly with the accumulated context. Moreover, a longer nominal context window does not by itself ensure reliable use of distant evidence: an expanding candidate set can increase retrieval interference and weaken long-range recall \citep{bai2023longbench,hsieh2024ruler}. These limitations have driven the evolution of attention along several interacting architectural directions, including approaches that store token memories more compactly, consult only selected past positions, carry context in recurrent states, structure how those states evolve, and combine different mechanisms within one model.

We organize this literature into four mechanism-centered research lines and one composition-centered line\footnote{In this survey, \emph{Softmax Attention}, \emph{Sparse Attention}, \emph{Linear Attention}, and \emph{State Space Models} denote the four mechanism-centered research lines, while \emph{Hybrid Architecture} denotes the composition-centered line. Other lowercase attention terms describe particular operators, access patterns, layers, heads, or paths rather than survey-level categories. Established method names retain their original capitalization.}:
\begin{itemize}
    \item \textbf{Softmax Attention} stores context in an enumerable collection of memory units, such as token KVs, latent entries, or summaries, and retrieves from them through normalized query--key weights \citep{vaswani2017attention,deepseek2024v2,chu2026ksa}. Recent variants share or compress these units across heads, layers, or time \citep{shazeer2019mqa,ainslie2023gqa,brandon2024crosslayer}. This reduces KV storage and data movement while preserving content-dependent retrieval, although cost still grows with the number and size of retained units.

    \item \textbf{Sparse Attention} keeps explicit content memory but lets each query consult only a subset of the available tokens or blocks \citep{beltagy2020longformer,zaheer2020bigbird,yuan2025nsa}. Selection ranges from fixed patterns to learned or compressed indexes \citep{qwen2026qwen38next,hu2026hils}, with some decisions reused across layers \citep{bai2026indexcache,zan2026longcatsparse}. This reduces score computation and memory traffic, but its effectiveness depends on retaining the evidence relevant to each query.

    \item \textbf{Linear Attention} folds the preceding context into one or more recurrent associative states rather than keeping a growing list of token memories \citep{katharopoulos2020linear}. Later work improves how these states retain and revise information \citep{sun2023retnet,schlag2021fastweight,yang2024gdn} or expands their capacity and temporal coverage \citep{pan2025sse,cabannes2026sdm,guo2025loglinear,wang2026dla}. This avoids a token-growing KV cache during decoding, but individual past tokens are no longer directly retrievable.

    \item \textbf{State Space Models} also carry context in recurrent states, but derive their updates from structured dynamical systems rather than an associative reformulation of attention. Their structured and input-dependent transitions seek to combine parallelizable training with bounded-state recurrent decoding. As with other compressed-state designs, individual past tokens are not preserved as separate retrievable entries \citep{gu2021s4,gu2023mamba,dao2024mamba2}.

    \item \textbf{Hybrid Architecture} retains context through combinations of explicit, sparse, linear, and state-space paths placed across layers, heads, branches, or tokens. They distribute fine-grained token retrieval, local modeling, compressed long-term memory, and computation across different parts of a model. This provides complementary capabilities within one architecture, while introducing additional placement, routing, and coordination decisions \citep{lieber2024jamba,dong2024hymba,mansukhani2024infini,du2025nha}.
\end{itemize}

These categories are intentionally \emph{not} a mutually exclusive partition. Most sparse mechanisms, for example, still apply Softmax after selecting a subset of keys; linear or state-space modules may also use selective operations; and Hybrid Architectures may contain members of every other line. We therefore place each method in the chapter that best matches its principal technical contribution and lineage.

\paragraph{Positioning relative to existing surveys.}
Prior surveys offer several complementary ways to navigate this rapidly expanding literature. General Transformer surveys organize architectural variants and applications, while efficient-Transformer surveys emphasize computational patterns, approximation strategies, and complexity \citep{lin2021surveytransformers,tay2020efficienttransformers}. Long-context surveys examine context extension, length extrapolation, training strategies, retrieval, and application settings \citep{huang2023longcontextsurvey,wang2024beyondlimits,liu2025longcontextsurvey}. More specialized reviews provide detailed treatments of efficient sparse and linear attention \citep{sun2026efficientattention}, state space models \citep{patro2024mamba360}, and KV-cache management and compression \citep{li2025kvcachesurvey}. More recently, an architecture-centric memory survey organizes LLM memory along representation form, update dynamics, and persistence, covering a broader range of implicit and explicit memory mechanisms beyond attention \citep{zhoubian2026memory}. Table~\ref{tab:survey-positioning} summarizes the primary organizing principles of these complementary survey perspectives.

\begin{table}[t]
    \centering
    \small
    \renewcommand{\arraystretch}{1.13}
    \caption{Representative survey perspectives relevant to this work and their primary organizing principles. The table summarizes the dominant perspective of each survey category rather than its complete coverage.}
    \label{tab:survey-positioning}
    \begin{tabularx}{\linewidth}{@{}
        >{\raggedright\arraybackslash}p{0.32\linewidth}
        >{\raggedright\arraybackslash}X@{}}
        \toprule
        \textbf{Survey perspective} & \textbf{Primary organizing principle} \\
        \midrule
        General / efficient Transformers \citep{lin2021surveytransformers,tay2020efficienttransformers} & Architecture families, computational complexity, and approximation patterns. \\
        Long-context LLMs \citep{huang2023longcontextsurvey,wang2024beyondlimits,liu2025longcontextsurvey} & Context extension, length extrapolation, training, retrieval, and application strategies. \\
        Efficient attention \citep{sun2026efficientattention} & Sparse and linear attention algorithms, efficiency, and deployment. \\
        State space models \citep{patro2024mamba360} & SSM lineage, state organization and dynamics, applications, and benchmarks. \\
        KV-cache management \citep{li2025kvcachesurvey} & Cache selection, compression, eviction, quantization, and execution efficiency. \\
        Memory-centric LLM architectures \citep{zhoubian2026memory} & Representation form, update dynamics, and persistence across a broad range of implicit and explicit memory mechanisms. \\
        \bottomrule
    \end{tabularx}
\end{table}

Taken together, these surveys provide complementary architecture-, efficiency-, context-, state-, cache-, and memory-centered views of the field. In comparison with broader memory-centered surveys, this survey focuses specifically on attention and adjacent sequence mixers. We adopt contextual memory as a shared unit of analysis and examine these mechanisms through five \emph{analytical dimensions}: \textbf{Memory Representation}, \textbf{Memory Update}, \textbf{Access}, \textbf{Readout}, and \textbf{Integration}. In brief, they ask what historical information remains represented, how the represented memory changes, what is made eligible for the current query, how eligible memory is read, and how one or more completed readouts are transformed or coordinated into the module output. The five dimensions and the chapter-level research lines serve different purposes: the chapter labels provide a readable historical and technical organization, whereas the analytical dimensions provide a multi-label description of the functions modified by an individual method. This lens thereby complements classifications based on architectural lineage, computational complexity, or application setting.

\paragraph{Scope and literature coverage.}
This survey focuses on model-level mechanisms used by autoregressive LLMs and closely related causal sequence mixers, with model-internal contextual memory as the shared unit of analysis. We cover representative work publicly available through September~22, 2026, identified through targeted literature searches, relevant prior surveys, backward and forward citation tracing, and primary technical documentation. The coverage is structured but not intended as an exhaustive systematic review: we emphasize foundational methods, major architectural transitions, and representative extensions that clarify recurring design choices across Memory Representation, Memory Update, Access, Readout, and Integration. Test-time learning, retrieval from external databases, multimodal-specific memory designs, and implementation-only optimizations remain outside the core taxonomy unless they directly alter the model's contextual-memory semantics.

At the architecture level, the mechanism review is complemented by a curated longitudinal inventory of publicly documented model releases and a frozen comparison of high-performing open-weight model endpoints, both closed on September~22, 2026. The inventory is based primarily on official papers, technical reports, model cards, and released configurations. Models released together are grouped when they share the same language-model attention backbone, whereas separately released versions or documented changes to that backbone form separate records. Records with insufficient architectural disclosure are marked as undisclosed rather than inferred. Natively multimodal models are included only when the relevant autoregressive language backbone is documented; visual encoders, modality interfaces, and other modality-specific components are excluded from the classification. The inventory is purposively curated rather than exhaustive or market-share weighted, and the inventory and frontier comparison are used to characterize documented adoption and coexistence rather than to attribute model quality causally to an attention mechanism.

The principal contributions of this survey are fourfold:
\begin{enumerate}
    \item We introduce a five-dimensional analytical lens---Memory Representation, Memory Update, Access, Readout, and Integration---for comparing how sequence mechanisms retain and use contextual memory. It provides a common vocabulary for Softmax Attention, Sparse Attention, Linear Attention, State Space Models, and Hybrid Architectures without reducing them to one computational model.

    \item We reconstruct the development of Softmax Attention, Sparse Attention, Linear Attention, and State Space Models by examining how their representative methods intervene in Memory Representation, Memory Update, Access, Readout, and Integration, while preserving overlaps among these historically defined research lines.

    \item We connect mechanism-level developments with two complementary analyses of publicly documented LLM architectures. A longitudinal inventory of 59 release-level records spanning 14 major model lineages traces the diversification of attention design, the growing use of layer-wise hybrid composition, and the emergence of cross-layer artifact reuse. A frozen comparison of 11 high-performing open-weight model endpoints provides a cross-sectional view of the attention structures represented near the performance frontier. Together, these analyses show continued architectural heterogeneity and the persistent role of explicit token retrieval in contemporary high-performing models.

    \item We develop a three-level memory-centric synthesis of the surveyed literature. At the mechanism level, explicit-memory and state-based methods retain different memory interfaces while expanding design control across an increasingly overlapping set of memory functions. At the architecture level, layer-wise composition distributes complementary memory processing across representational stages, while cross-layer reuse extends the lifetime of selected memory and routing artifacts across those stages. Together, these developments make network depth an emerging dimension along which contextual memory is constructed and managed. At the forward-looking level, this depth-wise perspective combines with temporal scope, substrate type, and representation granularity to motivate a stateful multidimensional memory-routing hypothesis governed by coordinated Sparse Write and Sparse Read.
\end{enumerate}

The remainder of this survey is organized as follows. Section~\ref{sec:unified-memory-centric-view} introduces the memory-centric analytical lens and its classification principles. Sections~\ref{sec:softmax-attention}-\ref{sec:state-space-models} examine Softmax Attention, Sparse Attention, Linear Attention, and State Space Models, respectively. Section~\ref{sec:hybrid-architectures} analyzes the composition of heterogeneous memory mechanisms across layers, heads, branches, and tokens. Section~\ref{sec:attention-designs-in-publicly-documented-architectures} examines architectural evolution and coordination through a longitudinal model inventory and a frozen comparison of high-performing open-weight models. Section~\ref{sec:landscape-and-future-directions} develops the mechanism-level and architecture-level syntheses and introduces the forward-looking multidimensional memory-routing hypothesis. Finally, Section~\ref{sec:conclusion} concludes the survey.

% !TEX root = ../main.tex
\section{A Unified Memory-Centric View of Attention}
\label{sec:unified-memory-centric-view}

When a model processes a token, it must make information from the preceding context available to the current computation. Different sequence architectures do this in visibly different ways. Softmax Attention retains separately addressable memory units; Sparse Attention limits which of those units are examined; and recurrent mechanisms continually compress the preceding context into one or more states. Looking only at these surface-level operators makes the families appear difficult to compare. At a functional level, however, each can be viewed as a system that maintains and uses internal memory while processing a sequence.

This memory-oriented interpretation is already present in several lines of research. Prior work has described linear attention as associative or fast-weight memory, related attention and state-space recurrences through their memory structure, and studied compressive, bounded, or growing recurrent memories \citep{katharopoulos2020linear,dao2024mamba2,mansukhani2024infini,karami2025lattice,behrouz2025trellis,behrouz2026memorycaching}. Building on this shared perspective, we use \emph{contextual memory} as a common term for the model-internal, input-dependent information that remains available while a sequence is being processed. Depending on the architecture, that information may take the form of token KV representations, compressed tokens or chunks, memory slots, associative matrices, structured recurrent states, or heterogeneous combinations of these forms. The term does not refer to the model's pretrained parameters, an external retrieval corpus, or biological memory.

\begin{figure}[t]
    \centering
    \includegraphics[width=\linewidth]
    {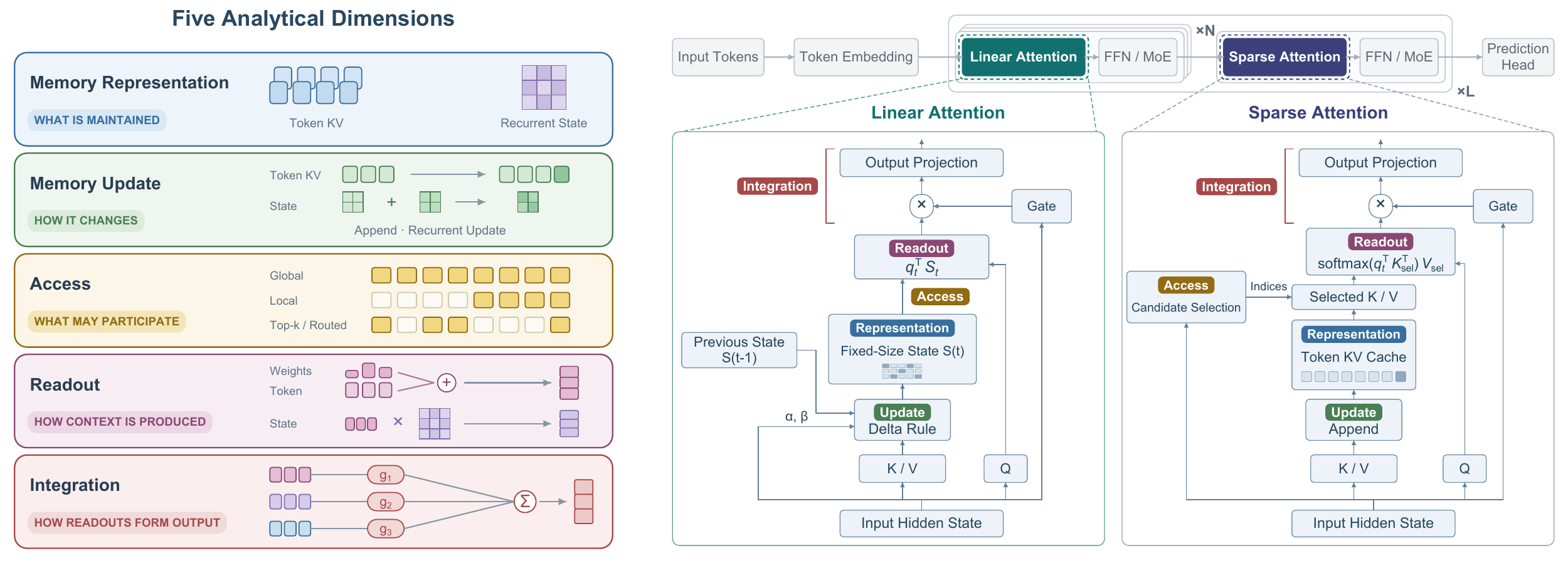}
    \caption{An architectural primer for the five-dimensional memory-centric view. The left panel illustrates Memory Representation, Memory Update, Access, Readout, and Integration through typical explicit-memory and recurrent-state operations. The center and right panels map these dimensions onto representative Linear Attention and Sparse Attention blocks, respectively, while the top row places the two mechanisms in a schematic layer-wise hybrid model. In the Linear Attention example, a delta-style update maintains a fixed-size recurrent state that is read through state contraction; in the Sparse Attention example, candidate selection restricts access to a token KV cache before the core attention readout. Gates and output projections illustrate Integration. The displayed update rule, gates, and block ratio are representative design choices rather than universal properties of these families, and the colored labels denote analytical roles rather than a mandatory execution sequence.}
    \label{fig:five-dimensional-framework}
\end{figure}

Once these mechanisms are viewed in terms of contextual memory, their design differences can be organized around five questions:
\begin{enumerate}
    \item \textbf{Memory Representation:} What information from the past is still represented, and in what form?
    \item \textbf{Memory Update:} How does the current input add to, modify, compress, or overwrite that memory?
    \item \textbf{Access:} Which represented information is eligible to be used for the current query?
    \item \textbf{Readout:} How is the eligible information weighted, decoded, or aggregated?
    \item \textbf{Integration:} How are one or more completed readouts transformed or combined into the module output?
\end{enumerate}
These questions form the analytical lens used throughout this survey. They describe functional roles rather than five components that every architecture must implement separately. A single operation may perform several roles at once, and the roles need not appear as a fixed sequence of implementation stages. Figure~\ref{fig:five-dimensional-framework} provides an architectural primer for this view: it illustrates the five dimensions through representative memory operations and maps them onto example Linear Attention and Sparse Attention blocks within a schematic layer-wise hybrid model. The concrete operations and block arrangement are illustrative rather than universal; the roles are defined formally below.

\subsection{Five Analytical Dimensions}

We now formalize the five questions one at a time. As a running setting, suppose that a model has processed the first $t-1$ tokens of a document and is processing token $t$. It may retain every preceding token as a separate KV entry, retain only selected or compressed entries, or carry the preceding context in a fixed-size recurrent state. Let $x_t$ denote the current input to the memory mechanism, and let $q_t$ denote the representation used to request information relevant to the current position.

\paragraph{Memory Representation.}
The first question is what the model has retained before it attempts to retrieve anything. Memory Representation specifies the form and organization of the maintained information, the granularity at which historical content remains distinguishable, and the amount of information the representation can carry. Softmax Attention retains separately addressable memory units. Multi-query attention (MQA) \citep{shazeer2019mqa}, grouped-query attention (GQA) \citep{ainslie2023gqa}, and multi-head latent attention (MLA) \citep{deepseek2024v2} reduce redundancy across heads or channels while preserving token-level memory units. Linear Attention instead compresses history into an associative recurrent state \citep{katharopoulos2020linear}, whereas SSMs maintain structured recurrent states \citep{gu2021s4,gu2023mamba}.

These examples expose two properties that recur throughout the survey. \emph{History coverage} describes how much of the preceding sequence may influence the maintained memory, whereas \emph{addressability} describes whether a particular historical unit remains separately selectable. A fixed-size state may cover the entire preceding sequence but no longer preserve every token as an individually addressable item. Conversely, an explicit KV cache preserves token-level addressability but grows as more tokens are retained.

To express these alternatives uniformly, let $\rho$ denote a memory schema that specifies the type, organization, granularity, capacity, and persistence of the representation. The maintained memory $\mathcal{M}_t$ belongs to the state space permitted by that schema:
\begin{equation}
    \mathcal{M}_t \in \mathfrak{M}_{\rho}.
    \label{eq:memory-representation-space}
\end{equation}
For example, $\rho$ may describe a growing list of token KVs, a bounded collection of summary slots, one associative matrix, a structured recurrent state, or several heterogeneous memory paths. This notation describes the information made available by the mechanism rather than requiring one physical realization: the memory may be materialized as a decoding cache, constructed in parallel during training, or carried recurrently as a state. The subscript $\rho$ on each operator below indicates that the concrete Update, Access, Readout, and Integration rules depend on the schema; a heterogeneous schema may therefore apply different rules to different memory units

\paragraph{Memory Update.}
Once the form of memory has been identified, the next question is how it changes when the model receives $x_t$. Memory Update covers the operations that write new information, preserve existing information, or remove and revise what was previously stored. In Softmax Attention, the standard update appends a key and value derived from the current input. A bounded or compressed memory may instead merge the new information with an existing summary. Recurrent mechanisms may use additive writes, multiplicative decay, input-dependent retention, delta correction, or explicit erase--write operations \citep{katharopoulos2020linear,yang2024gdn,gu2023mamba}.

Let $\mathcal{M}_t^{-}$ and $\mathcal{M}_t^{+}$ denote the memory immediately before and after the current update. The update is written abstractly as
\begin{equation}
    \mathcal{M}_{t}^{+}
    =
    \operatorname{Update}_{\rho}
    \left(
        \mathcal{M}_{t}^{-},
        x_t
    \right).
    \label{eq:unified-memory-update}
\end{equation}
The same expression therefore covers append-only KV caches, recurrent state transitions, compression into fixed-capacity slots, and controlled erase--write rules. The concrete equations differ across families and are introduced in the corresponding chapters. At this stage, the important distinction is that Update determines how memory changes; it does not determine which parts of the updated memory a particular query will use.

\paragraph{Access.}
That latter decision belongs to Access. Given a represented memory, Access determines which memory units or state interfaces are eligible to participate in the current read. Let $\widehat{\mathcal{M}}_t$ denote the memory visible to the read path. Depending on the mechanism's read--write convention, it may be the state before the current update, the state after the update, or an implementation-specific view constructed during the same computation. Access exposes an eligible memory view $\mathcal{C}_t$:
\begin{equation}
    \mathcal{C}_t
    =
    \operatorname{Access}_{\rho}
    \left(
        q_t,
        \widehat{\mathcal{M}}_t
    \right).
    \label{eq:unified-memory-access}
\end{equation}
Softmax Attention exposes all causally available memory units. Local or block-sparse patterns expose only positions allowed by a prescribed structure \citep{beltagy2020longformer,zaheer2020bigbird}, while learned sparse mechanisms use routers, indexes, or Top-$k$ selection to construct a query-dependent candidate set \citep{yuan2025nsa}. For a recurrent mechanism, Access instead exposes the current state interface: that state may carry information influenced by the complete causal history, but the individual historical tokens are no longer separately selectable. Accordingly, $\mathcal{C}_t$ may contain all represented token memories, a selected set of tokens or blocks, one or more recurrent states, or another mechanism-specific interface. It may also carry local metadata required by Readout, such as routing scores or group-level normalization quantities.

\paragraph{Readout.}
Once Access has established the eligible view $\mathcal{C}_t$, Readout determines how the query extracts contextual information from it. Softmax Attention computes normalized query--key similarities over the eligible memory units and aggregates their values \citep{vaswani2017attention}. Sparse Attention commonly changes the candidate set while retaining the same Softmax Readout over the selected memory units. Linear Attention reads an associative state through a kernelized contraction or a related state operation \citep{katharopoulos2020linear}, whereas an SSM produces a contextual representation through a structured projection of its recurrent state \citep{dao2024mamba2}. Access therefore answers \emph{what may be read}, while Readout answers \emph{how the eligible information is read}.

The resulting contextual representation is written as
\begin{equation}
    r_t
    =
    \operatorname{Readout}_{\rho}
    \left(
        q_t,
        \mathcal{C}_t
    \right).
    \label{eq:unified-memory-read}
\end{equation}
This separation allows two methods to expose different candidate sets while using the same readout rule, or to expose similar state interfaces while decoding them differently.

\paragraph{Integration.}
Integration determines how completed readouts are combined into the representation returned by the memory mechanism. With a single readout, this operation may reduce to an identity mapping or a fixed projection. In multi-head Softmax Attention, the contextual representations produced by the individual heads are concatenated and then mixed by the output projection, allowing information recovered by different heads to contribute jointly to the module output. Parallel branches and heterogeneous memory paths follow the same general principle, but may combine their readouts through summation, gating, routing, or other learned fusion rules.

Specific mechanisms illustrate how this stage can be modified. Gated Attention applies an input-dependent sigmoid gate to each head output before the heads are combined, allowing the contribution of retrieved information to vary across tokens \citep{qiu2025gatedattention}. At the branch level, Infini-attention uses a learned gate to combine the readout from local causal Softmax Attention with the readout from its compressive memory \citep{mansukhani2024infini}. These mechanisms alter how completed readouts contribute to the output without changing which historical memory units were eligible for the preceding read.

When multiple paths participate, $r_t$ may therefore denote a collection such as $\{r_t^{(p)}\}_{p\in\mathcal{P}_t}$ rather than a single vector. The general Integration rule is written as
\begin{equation}
    o_t
    =
    \operatorname{Integration}_{\rho}
    \left(
        r_t;
        x_t
    \right),
    \label{eq:unified-memory-integration}
\end{equation}
where $o_t$ is the output of the memory mechanism. Concatenation followed by the output projection is the inherited Integration rule of standard multi-head Softmax Attention. A method directly modifies Integration when it changes how completed readouts contribute to the output, for example through learned gating, conditional routing, or fusion across heterogeneous memory paths.

The five dimensions are functionally distinguishable, but they are not necessarily orthogonal or separately implemented. Changing Representation can alter what remains accessible; an input-dependent state transition can jointly affect Representation and Update; and a hierarchical sparse index can change both the address representation and the candidate set. Equations~\eqref{eq:memory-representation-space}--\eqref{eq:unified-memory-integration} should therefore be read as a shared analytical vocabulary rather than a universal computational graph.

Table~\ref{tab:framework-notation} collects the terminology, abbreviations, and framework-level symbols that recur across the survey. In this survey, \emph{Softmax Attention} denotes the broader explicit-memory lineage in which contextual information is retained as separately addressable representations and read through normalized query--key interactions. Sparse Attention is discussed separately because candidate selection and access-budget allocation form its central design problem. Later chapters also introduce local notation for mechanism-specific equations. Such local symbols retain the definitions given in their own sections and should not be interpreted as additions to the global framework notation below.

\begin{table}[H]
    \centering
    \footnotesize
    \setlength{\tabcolsep}{5pt}
    \renewcommand{\arraystretch}{1.10}
    \caption{Core terminology, abbreviations, and framework-level notation used throughout the survey. Chapter-specific symbols are defined locally in the corresponding sections.}
    \label{tab:framework-notation}
    \begin{tabularx}{\linewidth}{@{}>{\raggedright\arraybackslash}p{0.26\linewidth}>{\raggedright\arraybackslash}X@{}}
        \toprule
        \textbf{Term or symbol} & \textbf{Meaning in this survey} \\
        \midrule
        \multicolumn{2}{@{}l}{\textit{Core terminology and abbreviations}} \\
        \addlinespace[2pt]
        Contextual memory & Model-internal, input-dependent information retained or made available while processing the current sequence; it excludes pretrained parameters and external retrieval corpora. \\
        KV & Key--value representation associated with a token, compressed entry, or other explicit memory unit. \\
        Softmax Attention & The broader explicit-memory lineage in which contextual information is retained as separately addressable representations and read through normalized query--key interactions. \\
        Sparse Attention & A research line centered on restricting the eligible token-, block-, or entry-level candidate set; most methods retain a Softmax Readout over that set. \\
        Linear Attention & A family that represents history through one or more recurrent associative states rather than retaining the complete token-wise KV history. \\
        SSM & State Space Model; a sequence model that propagates contextual information through a structured recurrent state and state transition. \\
        Hybrid Architecture & An architecture that composes heterogeneous memory mechanisms or access regimes across layers, heads, branches, or tokens. \\
        \addlinespace[3pt]
        \multicolumn{2}{@{}l}{\textit{Framework-level notation}} \\
        \addlinespace[2pt]
        $t$; $x_t$; $q_t$ & Current sequence position; current input to the memory mechanism; and the representation used to query contextual memory. \\
        $\rho$; $\mathfrak{M}_{\rho}$ & Memory schema and the state space admitted by that schema. The schema records properties such as memory type, organization, granularity, capacity, and persistence. \\
        $\mathcal{M}_t$ & Contextual memory represented at position $t$. \\
        $\mathcal{M}_t^{-}$; $\mathcal{M}_t^{+}$ & Memory immediately before and after the current update. \\
        $\widehat{\mathcal{M}}_t$ & Memory view visible to the read path under the mechanism's read--write convention. \\
        $\mathcal{C}_t$ & Eligible memory view produced by Access; it may contain memory units, state interfaces, and local metadata required by Readout. \\
        $r_t$ & Contextual representation produced by Readout before any explicitly modeled Integration step. \\
        $o_t$ & Output of the memory mechanism after Integration. If a chapter omits a separate Integration operation, its local operator output may coincide with this quantity. \\
        $\mathcal{P}_t$; $r_t^{(p)}$ & Set of participating memory paths and the completed readout from path $p$. \\
        \bottomrule
    \end{tabularx}
\end{table}

A small notational qualification is useful when applying this table to later chapters. Mechanism-specific papers often use $o$ for the immediate output of an attention head or recurrent operator. Under the present framework, such a quantity functions as a readout $r_t$ when a subsequent head-, branch-, or path-level Integration step is modeled explicitly; it can coincide with the module output $o_t$ when Integration is the identity or is left implicit. 
%This convention allows later chapters to preserve familiar equations while keeping the functional distinction between Readout and Integration clear.

\subsection{Classification Principles and Family-Level Mapping}
\label{subsec:classification-and-family-mapping}

With the five dimensions and their notation in place, we now use the framework to compare the research lines surveyed in the following chapters. These lines reflect their principal technical questions and historical development rather than being derived mechanically from the five dimensions. The chapter labels therefore indicate the primary context in which a method is discussed, whereas the dimensions provide a multi-label description of the memory functions it modifies.

Table~\ref{tab:family-map} summarizes the dimensions most frequently emphasized in each research line. \textbf{P} marks a dimension commonly treated as a direct design target, whereas \textbf{S} marks one usually inherited or modified only in particular subfamilies. These labels are interpretive rather than exhaustive, and all five dimensions remain applicable to every research line.

\begin{table}[H]
    \centering
    \small
    \renewcommand{\arraystretch}{1.16}
    \caption{Qualitative emphasis of the surveyed research lines across the five analytical dimensions. \textbf{P} denotes a dimension commonly treated as a direct design target in the selected works, while \textbf{S} denotes a dimension inherited from the underlying mechanism or modified in particular subfamilies and extensions. The labels are intended as an interpretive guide rather than an exhaustive quantitative coding of the literature.}
    \label{tab:family-map}
    \begin{tabularx}{\linewidth}{@{}
        >{\raggedright\arraybackslash}p{0.20\linewidth}
        >{\raggedright\arraybackslash}X
        >{\centering\arraybackslash}p{0.075\linewidth}
        >{\centering\arraybackslash}p{0.075\linewidth}
        >{\centering\arraybackslash}p{0.075\linewidth}
        >{\centering\arraybackslash}p{0.075\linewidth}
        >{\centering\arraybackslash}p{0.075\linewidth}@{}}
        \toprule
        \textbf{Research line}
        & \textbf{Memory substrate}
        & \textbf{Rep.}
        & \textbf{Upd.}
        & \textbf{Acc.}
        & \textbf{Read.}
        & \textbf{Int.} \\
        \midrule

        Softmax Attention
        & Explicit memory units
        & \textbf{P} & \textbf{P} & \textbf{S} & \textbf{S} & \textbf{S} \\

        Sparse Attention
        & Explicit token/block KV
        & \textbf{S} & \textbf{S} & \textbf{P} & \textbf{S} & \textbf{S} \\

        Linear Attention
        & Associative recurrent state
        & \textbf{P} & \textbf{P} & \textbf{S} & \textbf{S} & \textbf{S} \\

        State Space Models
        & Structured recurrent state
        & \textbf{P} & \textbf{P} & \textbf{S} & \textbf{S} & \textbf{S} \\

        Hybrid Architectures
        & Heterogeneous memory substrates
        & \textbf{P} & \textbf{S} & \textbf{S} & \textbf{S} & \textbf{S} \\
        \bottomrule
    \end{tabularx}
\end{table}

Across the four mechanism-centered research lines, design effort is concentrated at different stages of memory processing. Softmax Attention primarily develops Memory Representation and Memory Update by reducing redundancy in explicit memory and changing how bounded or compressed memories are maintained \citep{shazeer2019mqa,ainslie2023gqa,bulatov2022rmt,deepseek2024v2}. Sparse Attention instead concentrates on Access: most methods preserve explicit token- or block-level content memory and apply a Softmax Readout over a restricted candidate set \citep{beltagy2020longformer,yuan2025nsa}. Routing keys, hierarchical indexes, compressed addresses, and cross-layer index reuse make Representation and Update secondary innovation axes, but these structures primarily support candidate construction rather than replace the underlying content memory \citep{hu2026hils,bai2026indexcache}. Linear Attention and State Space Models concentrate on Representation and Update because their principal developments concern the organization and evolution of recurrent states \citep{katharopoulos2020linear,yang2024gdn,gu2023mamba,dao2024mamba2}. Access, Readout, and Integration become more explicit mainly in multi-state, routed, gated, or otherwise specialized variants.

Hybrid Architecture requires a different interpretation because it composes multiple memory mechanisms or access regimes rather than defining a single internal memory operator. Memory Representation is therefore its primary family-level axis, while whether a particular design also modifies Memory Update, Access, Readout, or Integration depends on its composition granularity, as analyzed in Section~\ref{sec:hybrid-architectures}.

The resulting pattern highlights complementary design priorities. Softmax Attention reduces redundancy and reorganizes explicit memory while retaining a Softmax-compatible Readout; Sparse Attention allocates a limited Access budget over that memory; and Linear Attention and State Space Models develop the representation and update of recurrent states. Hybrid Architectures instead primarily diversify and allocate heterogeneous memory representations across layers, heads, branches, or tokens. Their effects on the remaining dimensions depend on the composition granularity and are examined in Section~\ref{sec:hybrid-architectures}. The research lines intersect when an individual method modifies several dimensions, shifting the design problem from optimizing an isolated operator toward coordinating Memory Representation, Memory Update, Access, Readout, and Integration under practical compute and storage constraints.
% !TEX root = ../main.tex
\section{Softmax Attention}
\label{sec:softmax-attention}

Softmax Attention is the explicit-memory lineage in which a query compares itself with an enumerable set of memory units, normalizes the resulting scores, and aggregates the corresponding values. For query position $t$ and attention head $h$, let $q_{t,h}$, $k_{i,h}$, and $v_{i,h}$ be the query, key, and value vectors obtained by linearly projecting the layer input. The standard formulation is then
\begin{equation}
\begin{aligned}
    a_{t,i,h}
    &=
    \frac{\exp\!\left(q_{t,h}k_{i,h}^{\top}/\sqrt{d_h}\right)}
    {\sum_{j\in\mathcal{R}_t}\exp\!\left(q_{t,h}k_{j,h}^{\top}/\sqrt{d_h}\right)}, &
    o_{t,h}
    &= \sum_{i\in\mathcal{R}_t}a_{t,i,h}v_{i,h},
\end{aligned}
\label{eq:softmax-attention}
\end{equation}
where $\mathcal{R}_t$ is the set of memory positions eligible for query $t$. In the baseline Transformer, \textbf{Memory Representation} consists of token-wise key--value pairs, \textbf{Memory Update} appends a new pair at each decoding step, \textbf{Access} exposes all positions allowed by the causal mask, \textbf{Readout} is the normalized query--key weighting in Eq.~\eqref{eq:softmax-attention}, and \textbf{Integration} concatenates the head outputs and applies the output projection \citep{vaswani2017attention}. This five-dimensional description is more precise than treating every efficiency modification as a change to ``attention'' in general: two methods can use the same Softmax Readout while changing entirely different memory functions.

Because each retained token adds an independently addressable key--value entry, the cache size and per-step readout cost both grow with context length. Consequently, efforts to reduce this cost while preserving the benefits of explicit memory can be organized around three optimization directions. First, \emph{memory-representation efficiency} preserves token-level addressability but reduces the amount of information stored per token or the number of redundant copies across heads and layers. Second, \emph{sequence-representation compression} changes both Representation and Update by consolidating a growing history into bounded states or lower-resolution remote memories. Third, \emph{readout and integration modulation} leaves the underlying memory largely intact while changing either how eligible values are weighted or how completed head outputs are transformed or combined into the attention-module output. These directions are composable: a model may use grouped-query or latent KV storage, compress remote history, and gate attention outputs at the same time.

The boundary with neighboring families follows the primary intervention. Methods that retain fine-grained KV memory but restrict the query-specific candidate set are treated as Sparse Attention, whereas methods that replace enumerable content-bearing units with a recurrent associative statistic are treated as Linear Attention. This section covers mechanisms that maintain an enumerable set of token, latent, slot, or summary units and use Softmax normalization either directly or within the readout; token-level addressability is typical but not required.

\subsection{Memory-Representation Efficiency}
\label{subsec:softmax-representation-efficiency}

A baseline autoregressive Transformer caches a key and value for each position, KV head, and KV-producing layer. For batch size $B$, cached length $T$, $L_{\mathrm{KV}}$ distinct KV layers, $H_{\mathrm{KV}}$ stored heads per layer, head dimension $d_h$, and $s$ bytes per scalar, the cache size is
\begin{equation}
    M_{\mathrm{KV}}
    = 2 B T L_{\mathrm{KV}} H_{\mathrm{KV}} d_h s,
    \label{eq:softmax-kv-cache-size}
\end{equation}
where the factor of two accounts for keys and values. These factors motivate three representation-efficiency directions: sharing across heads reduces $H_{\mathrm{KV}}$ by allowing multiple query heads to reuse KV states; compression along channels replaces full-width per-token KV states with a narrower latent payload; and sharing across layers reduces $L_{\mathrm{KV}}$ by allowing multiple consumer layers to reuse source-layer KV states. All three preserve token-level addressability and reduce only the coefficient of cache growth, not its linear dependence on $T$.

\subsubsection{Sharing across Heads}
\label{subsubsec:softmax-head-sharing}

Head sharing asks whether every query head requires an independently stored key and value for the same token. Let $H_q$ be the number of query heads and let $g:\{1,\ldots,H_q\}\rightarrow\{1,\ldots,H_{\mathrm{KV}}\}$ assign each query head to a stored KV head. Query head $h$ then evaluates Eq.~\eqref{eq:softmax-attention} with $(k_{i,h},v_{i,h})$ replaced by $(k_{i,g(h)},v_{i,g(h)})$. The access set $\mathcal{R}_t$ and the Softmax Readout remain unchanged; only the multiplicity of token representations differs.

\paragraph{Multi-Head Attention.}
Standard multi-head attention (MHA) uses $H_{\mathrm{KV}}=H_q$ and assigns every query head its own key and value projections. The arrangement offers maximal head-specific representational freedom but requires the cache to store and the decoder to load a distinct KV pair for every head and token \citep{vaswani2017attention}.

\paragraph{Multi-Query Attention.}
Multi-query attention (MQA) sets $H_{\mathrm{KV}}=1$: all query heads retain distinct query projections but read from one shared key head and one shared value head. It therefore removes most head-wise duplication in the cache and reduces memory bandwidth during incremental decoding, at the cost of sharing the same key and value representations across query heads \citep{shazeer2019mqa}.

\paragraph{Grouped-Query Attention.}
Grouped-query attention (GQA) partitions query heads into groups, with one KV head shared inside each group. Varying the number of groups creates a continuum between MHA and MQA, allowing model designers to trade head-specific memory representations for cache and bandwidth efficiency. GQA can also be obtained by adapting an MHA checkpoint, which makes it a practical architectural compromise rather than only a design for training from scratch \citep{ainslie2023gqa}.

\begin{figure}[t]
    \centering
    \includegraphics[width=\textwidth]{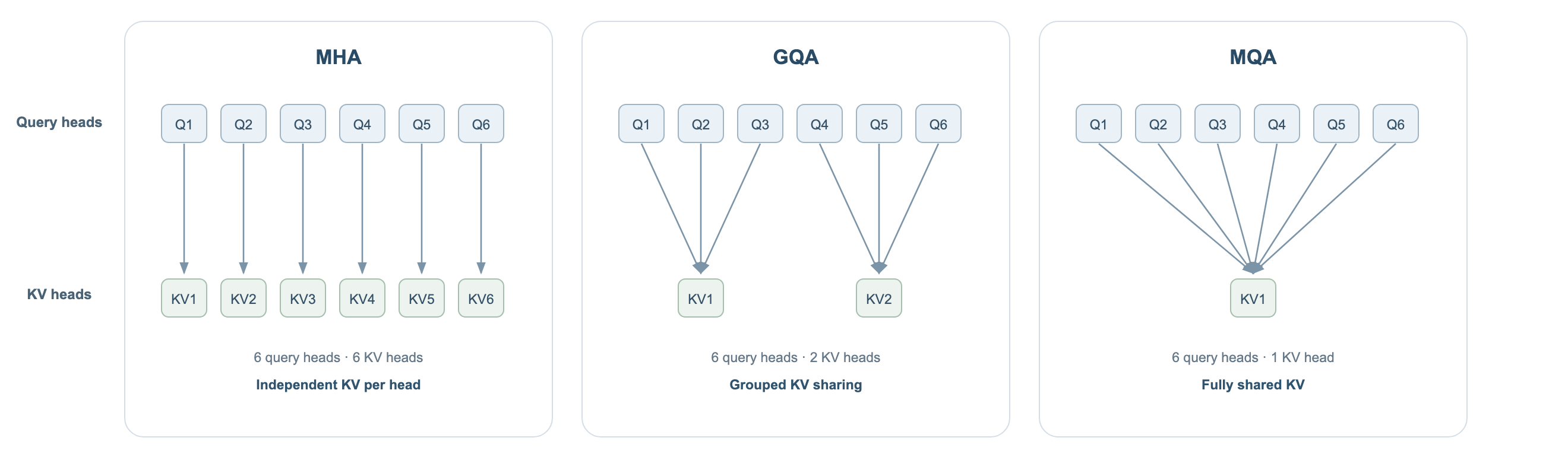}
    \caption{Head-sharing continuum from multi-head attention (MHA) to grouped-query attention (GQA) and multi-query attention (MQA). Each box represents one head; the number of query heads is fixed while the number of independently stored KV heads decreases.}
    \label{fig:softmax-head-sharing}
\end{figure}

The resulting continuum reduces the number of KV representations stored per token without changing the token candidate set, as summarized in Figure~\ref{fig:softmax-head-sharing}.

\subsubsection{Compression along Channels}
\label{subsubsec:softmax-channel-compression}

Channel compression reduces the width of each token memory rather than the number of its copies. Multi-head latent attention (MLA) maps each hidden state to a compact KV latent and derives the head-specific content keys and values from that shared latent. During decoding, the latent can be cached instead of the expanded multi-head tensors, and suitable projection matrices can be algebraically absorbed into the query and output paths so that the full keys and values need not be materialized at every step \citep{deepseek2024v2}.

Rotary positional encoding complicates this factorization because applying RoPE to reconstructed content keys prevents the key up-projection from being absorbed into the query path. MLA therefore separates content and positional components and caches the compact KV latent together with a shared positional component, preserving token-level addressability through a low-rank bottleneck \citep{deepseek2024v2}.

TransMLA studies the complementary deployment problem of converting an existing GQA checkpoint into an MLA-compatible parameterization. It rewrites the shared GQA projections into an equivalent multi-head form, applies low-rank factorization to obtain a common down-projection and head-specific up-projections, and uses continued training to adapt the model to the new latent representation \citep{meng2025transmla}. This distinction is important for evaluation: latent KV is an architectural representation change, not merely a post-hoc compression of already generated cache tensors. Its realized benefit depends on latent width, approximation error, positional treatment, projection fusion, cache layout, and kernel support.

\subsubsection{Sharing across Layers}
\label{subsubsec:softmax-layer-sharing}

Layer sharing targets redundancy along network depth. A conventional Transformer produces and stores new keys and values in every attention layer, so the cache grows approximately linearly with the number of such layers even after MQA or GQA has reduced the number of stored KV heads. Cross-Layer Attention (CLA) computes KV activations only in selected source layers and allows one or more subsequent consumer layers to reuse them \citep{brandon2024crosslayer}. Consumer layers retain their own queries, Softmax weights, output projections, residual paths, and feed-forward transformations; what is shared is the memory content, not the entire attention computation.

If one source layer serves $c$ adjacent layers, the number of independently cached KV layers decreases by approximately a factor of $c$. Head and layer sharing are orthogonal: an MLA, GQA, or MQA source layer can itself be shared by multiple consumers. This also distinguishes cross-layer KV sharing from cross-layer index reuse in sparse attention: the former shares content-bearing KV activations, whereas the latter shares an Access decision while each layer retains different KV content.

\subsection{Sequence-Representation Compression}
\label{subsec:softmax-sequence-compression}

The preceding methods reduce the cost of each token memory while retaining all token identities. Sequence-representation compression instead reorganizes history over time in two ways. Fixed-capacity recurrent memory repeatedly writes accumulated content into a bounded set of persistent states, while dynamic-resolution memory retains recent tokens at fine granularity and represents remote context with fewer, coarser units. The first bounds persistent state but may still use temporary local representations; the second may continue to grow with history, but more slowly than a token-wise cache.

Segment recurrence and compressed segment memory provide important precursors. Transformer-XL \citep{dai2019transformerxl} reuses hidden states from previous segments as an extended context, whereas Compressive Transformer \citep{rae2019compressive} adds a lower-resolution memory for states displaced from the recent cache. Later methods make the update process more explicit, place stricter bounds on the recurrent state, or design learned summary units for very long contexts.

\subsubsection{Fixed-Capacity Recurrent Memory}
\label{subsubsec:softmax-fixed-capacity}

Fixed-capacity methods maintain $K$ persistent memory units and repeatedly update their contents as new tokens or blocks arrive. Historical tokens therefore cease to exist as independently recoverable entries once their information has been consolidated. The central problem shifts from storing every token to controlling what a bounded memory retains, overwrites, and exposes to future queries.

\paragraph{Recurrent latent-state memory.}
Recurrent Memory Transformer (RMT) inserts special memory tokens into each segment. Read-memory tokens carry the state produced by the preceding segment, participate in attention with the current tokens, and are transformed into write-memory tokens that become the state for the next segment \citep{bulatov2022rmt}. The update is therefore performed within the Transformer computation rather than by an external pooling step. TransformerFAM similarly uses a block-wise feedback interface: token queries read the previous Feedback Attention Memory, while memory queries selectively aggregate the current block and return updated latent states to the next block. The mechanism reuses existing attention and feed-forward parameters instead of introducing a separate memory network \citep{hwang2024transformerfam}.

For both models, the persistent memory is a fixed set of hidden-state-like vectors updated at segment or block boundaries. Queries attend to these latent units rather than to the original tokens from earlier segments, so recurrence can propagate information across segment boundaries but does not preserve exact historical addressability: multiple past tokens may be entangled in the same state.

\paragraph{Recurrent KV-slot memory.}
Trellis and Lattice instead maintain explicit key and value slots. The keys define retrieval conditions and the values carry associated content, so current queries can attend directly to a bounded set of memory entries. Trellis uses a two-pass recurrent compression procedure and online-gradient-descent-inspired updates, together with forgetting that regulates the persistence of previous content \citep{behrouz2025trellis}. Lattice formulates cache compression as online optimization, exploits low-rank structure in the KV matrices, and applies state- and input-dependent gates together with an orthogonal update intended to reduce redundant writes; chunkwise parallelization mitigates the sequential cost of recurrence \citep{karami2025lattice}.

The distinction between latent-state and KV-slot memory concerns the persistent interface rather than the update frequency. RMT and TransformerFAM are primarily block- or segment-recurrent, whereas Trellis and Lattice update at finer granularity in their proposed realizations. In both cases, bounded recurrent updates replace append-only token accumulation, and original token identities cease to be separately accessible once consolidated into latent states or slots.

\subsubsection{Dynamic-Resolution Memory}
\label{subsubsec:softmax-dynamic-resolution}

Dynamic-resolution memory assigns different granularities to different portions of history. Recent tokens remain explicit to preserve exact local dependencies, whereas older content is consolidated into fewer summaries or compressed global entries. Unlike fixed-capacity recurrence, the remote memory may continue to grow, but at a slower rate determined by chunk size, summary ratio, hierarchy depth, or consolidation policy.

Kwai Summary Attention (KSA) combines a fine-grained recent path with summary memory for earlier chunks. Sliding Chunk Attention preserves local continuity, while summary attention gives queries access to compressed remote context; the memory state therefore contains both token-level recent units and coarser content-bearing summaries \citep{chu2026ksa}. HCA Core constructs a smaller collection of highly compressed global entries and performs dense Softmax attention over those entries \citep{deepseek2026v4}. The complete DeepSeek-V4 architecture additionally combines local sliding-window and compressed sparse pathways, so the full system is hybrid; HCA Core falls under sequence-representation compression because it replaces remote KV entries with compressed units and attends densely over those units rather than selecting a query-specific subset of the original cache.

Dynamic-resolution memory therefore rewrites the units represented before readout, whereas Sparse Attention retains fine-grained memory and restricts the query-specific candidate set.

Both fixed-capacity and dynamic-resolution policies compress information before future queries are known. Bounded updaters trade retention against overwrite and interference, while dynamic-resolution systems must decide when to consolidate content and what their summaries should preserve. Once token states are overwritten or merged, omitted details are generally unrecoverable. Evaluation should therefore report persistent memory size, local-context budget, compression or update overhead, readout cost, and long-range recall; bounded persistent memory does not eliminate local or update computation, and dynamic-resolution state may still grow with history.

\subsection{Readout and Integration Modulation}
\label{subsec:softmax-readout-integration}

The preceding mechanisms change the memory substrate or its update rule. A separate line of work preserves the underlying memory units but modifies either the Readout in Eq.~\eqref{eq:softmax-attention} or the subsequent Integration stage. Readout modulation transforms query--key scores or combines attention maps before they weight the values, whereas Integration modulation routes or gates completed head outputs before the output projection. These interventions can alter selectivity and information flow without reducing the stored memory or token candidate set.

\subsubsection{Score and Map Modulation}
\label{subsubsec:softmax-readout-modulation}

Talking-Heads Attention applies learned linear projections across heads both before Softmax and after Softmax. It therefore allows different heads to cooperate in forming attention maps, rather than requiring each head to generate and use its weights independently; the same learned mixing matrices are applied to every input \citep{shazeer2020talkingheads}. Dynamically Composable Multi-Head Attention (DCMHA) makes the cross-head composition input-dependent and modulates both pre-Softmax scores and post-Softmax weight matrices, increasing adaptive expressivity at the cost of additional composition and implementation overhead \citep{xiao2024dcmha}.

Differential Attention constructs two query--key paths, obtains two Softmax maps, and subtracts one from the other with a learned scale. Components that activate similarly in both paths can be suppressed, while their differences are retained. The effective weights may consequently be negative and no longer form a single probability distribution, although the component maps are Softmax-normalized \citep{ye2024differential}. Forgetting Transformer (FoX) instead introduces a data-dependent forget gate into the score path, allowing the contribution of historical positions to decay according to both content and distance before Softmax normalization \citep{lin2025forgetting}.

These methods leave the underlying candidate memory and, in their basic forms, the append-only cache unchanged. They alter the weights used to read that memory, but more concentrated or smaller effective weights do not imply fewer stored KV entries or fewer executed score computations.

\subsubsection{Head Routing and Gating}
\label{subsubsec:softmax-integration-modulation}

Mixture of Attention Heads (MoH) treats heads as experts and uses a token-dependent router to select and weight their contributions. Shared heads remain active for all tokens, while routed heads provide conditional specialization \citep{jin2024moh}. Because the controlled object is the contribution of head-level readouts to the module output, the principal change is \textbf{Integration}. A sufficiently specialized implementation may skip inactive heads and obtain conditional computation, but this head-level sparsity is distinct from selecting historical tokens in the Access stage.

Gated Attention introduces input-dependent sigmoid gates after the scaled dot-product attention outputs of individual heads and before their final combination \citep{qiu2025gatedattention}. The gates add a nonlinearity between value aggregation and output projection and allow the model to suppress a head's retrieved information for a particular token. A near-zero post-readout gate, however, does not retroactively eliminate the score computation, Softmax normalization, or KV loads already performed for that head.

MoH and Gated Attention thus both modify Integration, but with different control semantics. MoH allocates contribution through routing and can support discrete conditional execution; Gated Attention continuously rescales head outputs and primarily regulates information flow. Whether either mechanism yields wall-clock savings depends on kernel and runtime support for the induced head-level sparsity.

Readout and Integration modifications should therefore be evaluated separately from memory compression. Numerical sparsity in scores, routing probabilities, or gates reduces computation only when the implementation can skip the corresponding score, load, or head operations before execution; evidence should therefore pair model quality with FLOPs, memory traffic, kernel efficiency, and end-to-end latency.

\subsection{Summary}
\label{subsec:softmax-summary}

Softmax Attention research has pursued several complementary ways to reduce the cost of explicit-memory retrieval while retaining normalized content-based readout. For autoregressive decoding, MQA and GQA reduce head-wise replication, while MLA and cross-layer sharing compress token memory along channel and depth axes while preserving token-level addressability. At longer contexts, lowering the cost of each token is no longer sufficient to control growth with sequence length, motivating bounded recurrent memories and dynamic-resolution summaries that trade exact historical granularity for capacity. Score and map modulation, head routing, and output gating developed as an orthogonal direction for improving selectivity and information flow without directly shrinking the represented memory.

Head sharing is a common design for dense Softmax Attention in several mainstream model families: GQA is used across Llama~3/4 \citep{dubey2024llama3,meta2025llama4} and Qwen2/3 \citep{yang2024qwen2,qwen2025qwen3}, and remains as the periodic gated full-attention path\footnote{Here, \emph{full attention} denotes full-context causal Softmax Attention, in which each query may attend to all causally eligible token positions represented at that layer. The term describes Access scope rather than head or KV parameterization; a full-attention path may use MHA, MQA, GQA, or MLA, and contrasts with local, streaming, or sparsely selected attention.} in Qwen3.5 \citep{qwen2026qwen35}, Qwen3.6 \citep{qwen2026qwen3635ba3b,qwen2026qwen3627b}, and \href{https://huggingface.co/Qwen/Qwen3.8-2.4T-A95B}{Qwen3.8-2.4T-A95B}. Latent KV compression forms a second major lineage. MLA, introduced in DeepSeek-V2 and retained in DeepSeek-V3/V3.2 \citep{deepseek2024v2,deepseek2024v3,deepseek2025v32}, is also used by Kimi K2 \citep{kimi2025k2}, by the Gated MLA layers of Kimi K3 \citep{kimi2026k3}, and as the content representation underlying DeepSeek Sparse Attention in the GLM-5 series \citep{glm2026glm5,glm2026glm52,glm2026glm53}. More aggressive sequence compression appears in DeepSeek-V4's Heavily Compressed Attention \citep{deepseek2026v4}, which applies dense Softmax readout to consolidated KV entries, while output gating accompanies both GQA in recent Qwen hybrids \citep{qwen2026qwen35,qiu2025gatedattention} and MLA in Kimi K3 \citep{kimi2026k3}.

Across these systems, the GQA, latent-KV, compressed-memory, and gating components belong to the developments surveyed here; where recurrent or sparse paths are also present, those paths are treated in later sections. The next section turns to Sparse Attention, where the central question is which represented token- or block-level units are made eligible for each query.

% !TEX root = ../main.tex
\section{Sparse Attention}
\label{sec:sparse-attention}

Dense causal Softmax Attention follows Eq.~\eqref{eq:softmax-attention} and normalizes each query over all causally available positions, so $|\mathcal{R}_t|=\Theta(t)$. Consequently, training and prefill require $\Theta(L^2d)$ attention work, while each decoding step at context length $L$ requires $\Theta(Ld)$ work and reads a KV cache whose size grows as $\Theta(Ld_{\mathrm{KV}})$ per layer. Memory-efficient kernels can reduce intermediate storage but not the number of query--key interactions. Sparse Attention instead restricts Access to a budgeted subset $|\mathcal{I}_t|=b_t\ll t$, targeting computation and data movement while preserving the evidence needed by Readout \citep{child2019sparse,beltagy2020longformer,zaheer2020bigbird,tay2020efficienttransformers}.

Let $\mathcal{U}_t^{(\ell)}$ denote the addressable memory units available before the read at position $t$. A unit may be a token, page, block, chunk, or compressed entry. Sparse Access first constructs a selection-evidence vector $e_{t,:}^{(\ell)}$ (one score per unit) and then chooses an index set under budget $b_t$:
\begin{equation}
    \mathcal{I}_t^{(\ell)}
    = \operatorname{Select}\!\left(
        e_{t,:}^{(\ell)},\mathcal{U}_t^{(\ell)};b_t
      \right).
    \label{eq:sparse-access}
\end{equation}
The selected indices are then used to gather the corresponding entries from content memory. In the common case, the selected content remains ordinary token KVs and the readout is a candidate-local Softmax,
\begin{equation}
    r_t^{(\ell)}
    = \operatorname{softmax}\!\left(
        \frac{q_t^{(\ell)}(K_{\mathcal{I}_t}^{(\ell)})^{\top}}{\sqrt{d}}
      \right)V_{\mathcal{I}_t}^{(\ell)}.
    \label{eq:sparse-local-readout}
\end{equation}
These operators are written for a single layer; we suppress the layer index $(\ell)$ in the family-specific equations that follow and restore it only where cross-layer relationships are discussed.
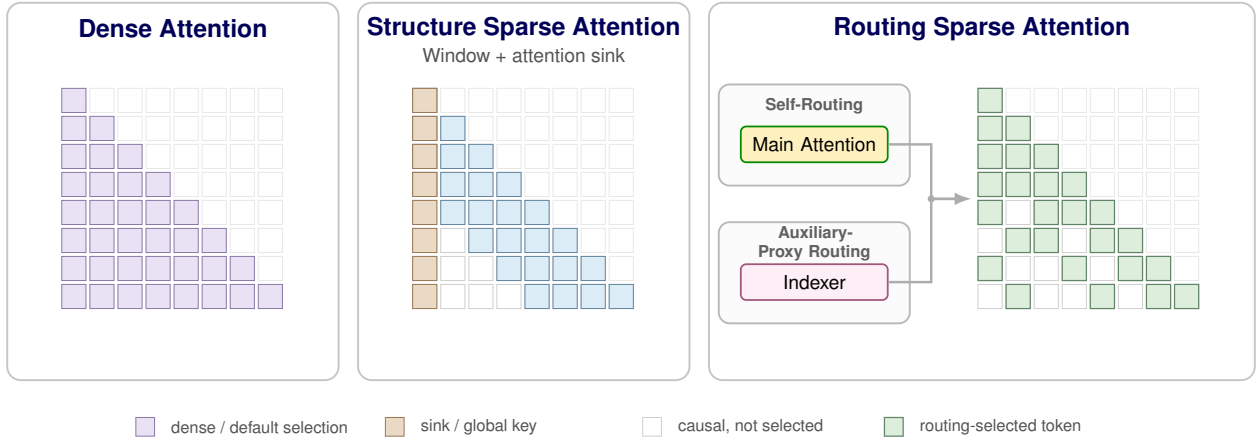
\begin{figure}[t]
    \centering
    \definecolor{densePurpleFill}{HTML}{E9E2F4}
\definecolor{densePurpleLine}{HTML}{8B78A6}
\definecolor{windowBlueFill}{HTML}{DCECF6}
\definecolor{windowBlueLine}{HTML}{6486A0}
\definecolor{sinkCoffeeFill}{HTML}{E9D9C5}
\definecolor{sinkCoffeeLine}{HTML}{9B7955}
\definecolor{routingGreenFill}{HTML}{DDEEDC}
\definecolor{routingGreenLine}{HTML}{628467}
\definecolor{mainAttentionYellow}{HTML}{FFF1BF}

\resizebox{\textwidth}{!}{%
\begin{tikzpicture}[
    x=1cm,
    y=1cm,
    font=\sffamily\scriptsize,
    >={Latex[length=2.2mm]},
    panel/.style={fill=white,draw=gray!45,rounded corners=5pt,line width=0.7pt},
    densecell/.style={fill=densePurpleFill,draw=densePurpleLine,line width=0.35pt},
    windowcell/.style={fill=windowBlueFill,draw=windowBlueLine,line width=0.35pt},
    sink/.style={fill=sinkCoffeeFill,draw=sinkCoffeeLine,line width=0.35pt},
    inactive/.style={fill=white,draw=gray!38,line width=0.3pt},
    future/.style={fill=white,draw=gray!18,line width=0.3pt},
    routingselected/.style={fill=routingGreenFill,draw=routingGreenLine,line width=0.45pt},
    routerouter/.style={fill=gray!3,draw=gray!55,rounded corners=4pt,line width=0.7pt,minimum width=2.45cm,minimum height=1.30cm},
    selfinner/.style={fill=mainAttentionYellow,draw=green!55!black,rounded corners=2pt,line width=0.65pt,minimum width=1.85cm,minimum height=0.46cm,text width=1.65cm,align=center},
    auxinner/.style={fill=magenta!7,draw=magenta!60!black,rounded corners=2pt,line width=0.65pt,minimum width=1.85cm,minimum height=0.46cm,text width=1.65cm,align=center},
    routingline/.style={draw=gray!65,line width=0.9pt},
    mergedarrow/.style={->,draw=gray!65,line width=1.0pt},
    title/.style={font=\sffamily\bfseries\small,text=blue!35!black,align=center},
    routertitle/.style={font=\sffamily\bfseries\tiny,text=gray!75!black,align=center},
    subtitle/.style={font=\sffamily\scriptsize,text=gray!65!black,align=center},
    axis/.style={font=\sffamily\tiny,text=gray!65!black}
]

% Panel backgrounds
\path[panel] (0,0.60) rectangle ++(4.25,4.85);
\path[panel] (4.50,0.60) rectangle ++(4.25,4.85);
\path[panel] (9.00,0.60) rectangle ++(7.00,4.85);

% Titles
\node[title] at (2.125,5.12) {Dense Attention};
\node[title] at (6.625,5.12) {Structure Sparse Attention};
\node[subtitle] at (6.625,4.78) {Window + attention sink};
\node[title] at (12.50,5.12) {Routing Sparse Attention};

% Dense causal map
\begin{scope}[shift={(0.70,4.35)}]
    \foreach \r in {0,...,7} {
        \foreach \c in {0,...,7} {
            \pgfmathsetmacro{\xx}{0.36*\c}
            \pgfmathsetmacro{\yy}{-0.36*\r}
            \ifnum\c>\r
                \path[future] (\xx,\yy) rectangle ++(0.31,-0.31);
            \else
                \path[densecell] (\xx,\yy) rectangle ++(0.31,-0.31);
            \fi
        }
    }
\end{scope}

% Structure-constrained map: four-token local window plus first-column sink
\begin{scope}[shift={(5.20,4.35)}]
    \foreach \r in {0,...,7} {
        \foreach \c in {0,...,7} {
            \pgfmathsetmacro{\xx}{0.36*\c}
            \pgfmathsetmacro{\yy}{-0.36*\r}
            \ifnum\c>\r
                \path[future] (\xx,\yy) rectangle ++(0.31,-0.31);
            \else
                \path[inactive] (\xx,\yy) rectangle ++(0.31,-0.31);
            \fi
        }
    }
    % Sliding window of width four
    \foreach \r in {0,...,7} {
        \foreach \offset in {0,...,3} {
            \pgfmathtruncatemacro{\c}{\r-\offset}
            \ifnum\c<0\else
                \pgfmathsetmacro{\xx}{0.36*\c}
                \pgfmathsetmacro{\yy}{-0.36*\r}
                \path[windowcell] (\xx,\yy) rectangle ++(0.31,-0.31);
            \fi
        }
    }
    % Attention sink at the first key position
    \foreach \r in {0,...,7} {
        \pgfmathsetmacro{\yy}{-0.36*\r}
        \path[sink] (0,\yy) rectangle ++(0.31,-0.31);
    }
\end{scope}

% Routing selector modules, stacked on the left
\node[routerouter] (selfbox) at (10.35,3.75) {};
\node[routertitle,text width=2.15cm] at (10.35,4.13) {Self-Routing};
\node[selfinner] (mainattention) at (10.35,3.63) {Main Attention};

\node[routerouter] (auxbox) at (10.35,1.98) {};
\node[routertitle,text width=2.15cm] at (10.35,2.36) {Auxiliary-Proxy Routing};
\node[auxinner] (indexer) at (10.35,1.86) {Indexer};

% Token-routing map over the full causal triangle
\begin{scope}[shift={(12.45,4.35)}]
    \coordinate (mapleft) at (-0.08,-1.42);
    \foreach \r in {0,...,7} {
        \foreach \c in {0,...,7} {
            \pgfmathsetmacro{\xx}{0.36*\c}
            \pgfmathsetmacro{\yy}{-0.36*\r}
            \ifnum\c>\r
                \path[future] (\xx,\yy) rectangle ++(0.31,-0.31);
            \else
                \path[inactive] (\xx,\yy) rectangle ++(0.31,-0.31);
            \fi
        }
    }
    % The diagonal is always retained.
    \foreach \r in {0,...,7} {
        \pgfmathsetmacro{\xx}{0.36*\r}
        \pgfmathsetmacro{\yy}{-0.36*\r}
        \path[routingselected] (\xx,\yy) rectangle ++(0.31,-0.31);
    }
    % Routed selections give four retained cells per row once possible.
    \foreach \c/\r in {0/1,1/2,0/2,1/3,2/3,0/3,0/4,2/4,3/4,4/5,1/5,2/5,3/6,5/6,0/6,1/7,6/7,4/7} {
        \pgfmathsetmacro{\xx}{0.36*\c}
        \pgfmathsetmacro{\yy}{-0.36*\r}
        \path[routingselected] (\xx,\yy) rectangle ++(0.31,-0.31);
    }
\end{scope}

% Merge the two routing paths before the candidate map
\coordinate (merge) at (11.85,2.93);
\draw[routingline] (mainattention.east) -| (merge);
\draw[routingline] (indexer.east) -| (merge);
\fill[gray!65] (merge) circle (0.045cm);
\draw[mergedarrow] (merge) -- (mapleft);

% Legend

\path[densecell] (1.65,0.12) rectangle ++(0.24,-0.24);
\node[axis,anchor=west] at (1.98,0.00) {dense / default selection};
\path[sink] (4.85,0.12) rectangle ++(0.24,-0.24);
\node[axis,anchor=west] at (5.18,0.00) {sink / global key};
\path[inactive] (8.15,0.12) rectangle ++(0.24,-0.24);
\node[axis,anchor=west] at (8.48,0.00) {causal, not selected};
\path[routingselected] (11.25,0.12) rectangle ++(0.24,-0.24);
\node[axis,anchor=west] at (11.58,0.00) {routing-selected token};

\end{tikzpicture}%
}
    \caption{Attention-map comparison between dense and sparse access. Dense Attention retains every causal pair. Structure Sparse Attention combines a fixed four-token sliding window with a sink connection to the first token. Routing Sparse Attention retains the causal lower-triangular domain, selects the diagonal by default, and chooses up to three additional tokens per row. Self-routing from the main attention and auxiliary-proxy routing through an indexer are alternative ways to produce the same type of selected support; all selected routing cells therefore share one visual style, while unselected causal cells remain unfilled.}
    \label{fig:sparse-attention-overview}
\end{figure}

Sparse Attention is primarily an \textbf{Access} intervention: methods differ in whether support is prescribed structurally, inferred from backbone representations, predicted by an auxiliary index, or reused across steps and layers. \textbf{Memory Representation} and \textbf{Memory Update} become relevant when selection requires address memory---such as pooled summaries, projected keys, or binary codes---or when content itself is compressed, refreshed, or shared. Most methods retain the candidate-local Softmax Readout in Eq.~\eqref{eq:sparse-local-readout}; Readout changes directly only when routing scores also allocate probability mass, while Integration matters mainly in multi-branch designs that combine sparse, local, or compressed paths. The chapter therefore centers on support construction and its execution cost, noting the other dimensions only when a method changes them directly. Figure~\ref{fig:sparse-attention-overview} contrasts dense causal access with representative structure-constrained and routing-based sparse supports.

Sparse Attention methods are organized here by two questions: how candidate-support evidence is produced and how long a selection decision remains valid. Along the evidence-source axis, \emph{Structure-Constrained} methods prescribe a topology or pattern family, \emph{Self-Routing} methods derive evidence from backbone representations or execution state, and \emph{Auxiliary-Proxy Routing} introduces a separate indexing space. Orthogonally, \emph{Temporal and Cross-Layer Reuse} amortizes supports, indexes, or memory representations across time and depth and can be combined with any evidence source. These subsection labels therefore serve as editorial homes rather than mutually exclusive classes; routing granularity and training mode remain cross-cutting attributes.

\subsection{Structure-Constrained Sparse Attention}
\label{subsec:structure-constrained-sparse}

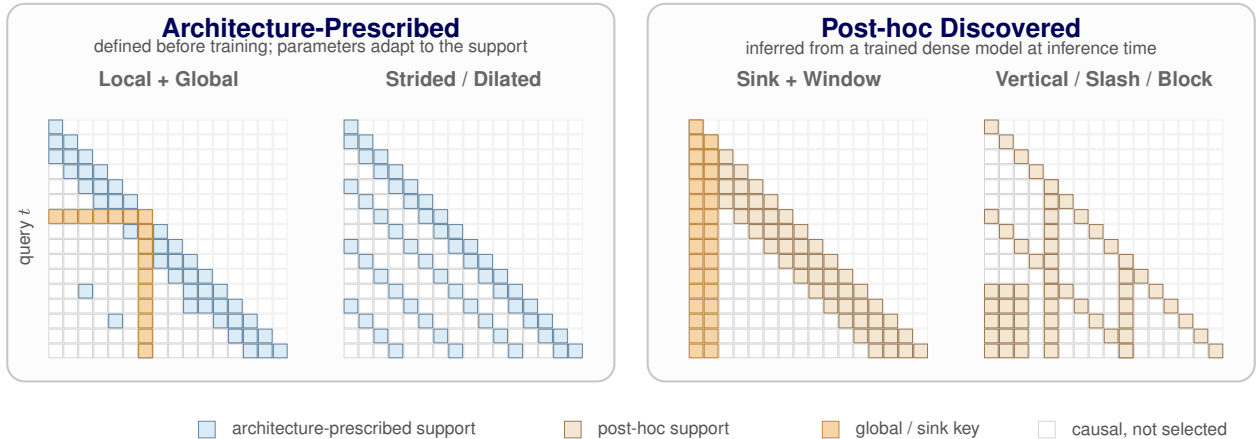
\begin{figure}[t]
    \centering
    \definecolor{archPatternFill}{HTML}{DCECF6}
\definecolor{archPatternLine}{HTML}{6486A0}
\definecolor{postPatternFill}{HTML}{F3E7D3}
\definecolor{postPatternLine}{HTML}{9B7955}
\definecolor{globalKeyFill}{HTML}{F6D5A8}
\definecolor{globalKeyLine}{HTML}{C0863A}

\resizebox{\textwidth}{!}{%
\begin{tikzpicture}[
    x=1cm,
    y=1cm,
    font=\sffamily\scriptsize,
    band/.style={fill=gray!2,draw=gray!45,rounded corners=6pt,line width=0.7pt},
    arch/.style={fill=archPatternFill,draw=archPatternLine,line width=0.25pt},
    post/.style={fill=postPatternFill,draw=postPatternLine,line width=0.25pt},
    globalkey/.style={fill=globalKeyFill,draw=globalKeyLine,line width=0.25pt},
    inactive/.style={fill=white,draw=gray!32,line width=0.2pt},
    future/.style={fill=white,draw=gray!12,line width=0.2pt},
    grouptitle/.style={font=\sffamily\bfseries\small,text=blue!35!black,align=center},
    maptitle/.style={font=\sffamily\bfseries\scriptsize,text=gray!80!black,align=center},
    mapsub/.style={font=\sffamily\tiny,text=gray!65!black,align=center},
    axis/.style={font=\sffamily\tiny,text=gray!65!black,align=center}
]

% Group bands
\path[band] (0.25,0.80) rectangle (7.55,5.35);
\path[band] (7.95,0.80) rectangle (15.25,5.35);
\node[grouptitle] at (3.90,5.06) {Architecture-Prescribed};
\node[mapsub] at (3.90,4.82) {defined before training; parameters adapt to the support};
\node[grouptitle] at (11.60,5.06) {Post-hoc Discovered};
\node[mapsub] at (11.60,4.82) {inferred from a trained dense model at inference time};

% Map A: Local + Global (causal-ized global token)
\node[maptitle] at (2.19,4.45) {Local + Global};
\begin{scope}[shift={(0.75,3.95)}]
    \foreach \r in {0,...,15}{\foreach \c in {0,...,15}{
        \pgfmathsetmacro{\xx}{0.18*\c}\pgfmathsetmacro{\yy}{-0.18*\r}
        \ifnum\c>\r \path[future] (\xx,\yy) rectangle ++(0.165,-0.165);
        \else \path[inactive] (\xx,\yy) rectangle ++(0.165,-0.165); \fi}}
    % sliding window (width three)
    \foreach \r in {0,...,15}{\foreach \o in {0,1,2}{
        \pgfmathtruncatemacro{\c}{\r-\o}
        \ifnum\c<0\else \pgfmathsetmacro{\xx}{0.18*\c}\pgfmathsetmacro{\yy}{-0.18*\r}
        \path[arch] (\xx,\yy) rectangle ++(0.165,-0.165);\fi}}
    % random long-range edges
    \foreach \c/\r in {2/11,9/12,4/13,12/14,6/15}{
        \pgfmathsetmacro{\xx}{0.18*\c}\pgfmathsetmacro{\yy}{-0.18*\r}
        \path[arch] (\xx,\yy) rectangle ++(0.165,-0.165);}
    % global token (g=6): attended by all later positions (column)
    \foreach \r in {6,...,15}{\pgfmathsetmacro{\yy}{-0.18*\r}
        \path[globalkey] (1.08,\yy) rectangle ++(0.165,-0.165);}
    % global token (g=6): attends all earlier positions (causal row)
    \foreach \c in {0,...,6}{\pgfmathsetmacro{\xx}{0.18*\c}
        \path[globalkey] (\xx,-1.08) rectangle ++(0.165,-0.165);}
\end{scope}

% Map B: Strided / Dilated
\node[maptitle] at (5.74,4.45) {Strided / Dilated};
\begin{scope}[shift={(4.30,3.95)}]
    \foreach \r in {0,...,15}{\foreach \c in {0,...,15}{
        \pgfmathsetmacro{\xx}{0.18*\c}\pgfmathsetmacro{\yy}{-0.18*\r}
        \ifnum\c>\r \path[future] (\xx,\yy) rectangle ++(0.165,-0.165);
        \else \path[inactive] (\xx,\yy) rectangle ++(0.165,-0.165); \fi}}
    % local band plus strided/dilated offsets
    \foreach \r in {0,...,15}{\foreach \o in {0,1,4,8,12}{
        \pgfmathtruncatemacro{\c}{\r-\o}
        \ifnum\c<0\else \pgfmathsetmacro{\xx}{0.18*\c}\pgfmathsetmacro{\yy}{-0.18*\r}
        \path[arch] (\xx,\yy) rectangle ++(0.165,-0.165);\fi}}
\end{scope}

% Map C: Sink + Window (Lambda-shape)
\node[maptitle] at (9.89,4.45) {Sink + Window};
\begin{scope}[shift={(8.45,3.95)}]
    \foreach \r in {0,...,15}{\foreach \c in {0,...,15}{
        \pgfmathsetmacro{\xx}{0.18*\c}\pgfmathsetmacro{\yy}{-0.18*\r}
        \ifnum\c>\r \path[future] (\xx,\yy) rectangle ++(0.165,-0.165);
        \else \path[inactive] (\xx,\yy) rectangle ++(0.165,-0.165); \fi}}
    % recent window (width four)
    \foreach \r in {0,...,15}{\foreach \o in {0,1,2,3}{
        \pgfmathtruncatemacro{\c}{\r-\o}
        \ifnum\c<0\else \pgfmathsetmacro{\xx}{0.18*\c}\pgfmathsetmacro{\yy}{-0.18*\r}
        \path[post] (\xx,\yy) rectangle ++(0.165,-0.165);\fi}}
    % attention sink (first two keys)
    \foreach \r in {0,...,15}{\foreach \c in {0,1}{
        \ifnum\c>\r\else \pgfmathsetmacro{\xx}{0.18*\c}\pgfmathsetmacro{\yy}{-0.18*\r}
        \path[globalkey] (\xx,\yy) rectangle ++(0.165,-0.165);\fi}}
\end{scope}

% Map D: Profiled vertical / slash / block
\node[maptitle] at (13.44,4.45) {Vertical / Slash / Block};
\begin{scope}[shift={(12.00,3.95)}]
    \foreach \r in {0,...,15}{\foreach \c in {0,...,15}{
        \pgfmathsetmacro{\xx}{0.18*\c}\pgfmathsetmacro{\yy}{-0.18*\r}
        \ifnum\c>\r \path[future] (\xx,\yy) rectangle ++(0.165,-0.165);
        \else \path[inactive] (\xx,\yy) rectangle ++(0.165,-0.165); \fi}}
    % diagonal anchor plus a separated single-width slash line
    \foreach \r in {0,...,15}{\foreach \o in {0,6}{
        \pgfmathtruncatemacro{\c}{\r-\o}
        \ifnum\c<0\else \pgfmathsetmacro{\xx}{0.18*\c}\pgfmathsetmacro{\yy}{-0.18*\r}
        \path[post] (\xx,\yy) rectangle ++(0.165,-0.165);\fi}}
    % vertical lines
    \foreach \r in {4,...,15}{\pgfmathsetmacro{\yy}{-0.18*\r}
        \path[post] (0.72,\yy) rectangle ++(0.165,-0.165);}
    \foreach \r in {9,...,15}{\pgfmathsetmacro{\yy}{-0.18*\r}
        \path[post] (1.62,\yy) rectangle ++(0.165,-0.165);}
    % remote block (lower-left)
    \foreach \r in {11,...,15}{\foreach \c in {0,1,2}{
        \pgfmathsetmacro{\xx}{0.18*\c}\pgfmathsetmacro{\yy}{-0.18*\r}
        \path[post] (\xx,\yy) rectangle ++(0.165,-0.165);}}
\end{scope}

% Shared axis hint
\node[axis,rotate=90] at (0.48,2.55) {query $t$};

% Legend
\path[arch] (2.55,0.32) rectangle ++(0.20,-0.20);
\node[axis,anchor=west] at (2.83,0.22) {architecture-prescribed support};
\path[post] (6.95,0.32) rectangle ++(0.20,-0.20);
\node[axis,anchor=west] at (7.23,0.22) {post-hoc support};
\path[globalkey] (10.05,0.32) rectangle ++(0.20,-0.20);
\node[axis,anchor=west] at (10.33,0.22) {global / sink key};
\path[inactive] (12.65,0.32) rectangle ++(0.20,-0.20);
\node[axis,anchor=west] at (12.93,0.22) {causal, not selected};

\end{tikzpicture}%
}
    \caption{Two origins of structure-constrained support, shown as causal attention maps (rows are queries, columns are keys). Architecture-prescribed patterns---local windows with global or random edges, and strided/dilated schedules---are fixed before training so parameters adapt to them. Post-hoc patterns---sink-plus-window rules and offline-profiled vertical/slash/block families---are inferred from a trained dense model and applied at inference. The global token is drawn in causal-ized form: it attends all earlier positions (its causal row) and is attended by all later ones (its column), never accessing future keys; a sink instead contributes only a column.}
    \label{fig:sparse-structure-pattern-origin}
\end{figure}

Structure-constrained methods restrict Access using positional relations, distance schedules, or a predefined topology. Their distinguishing property is not a particular mask geometry, but that the admissible support is limited by a pattern established independently of unrestricted query-specific search. Let $\mathcal{P}$ denote either a pattern designed before training or a pattern family discovered from a trained dense model. The instantiated support can be written as
\begin{equation}
    \mathcal{P}\in
    \{\mathcal{P}_{\mathrm{design}},\mathcal{P}_{\mathrm{observed}}\},
    \qquad
    \mathcal{I}_t
    = \operatorname{Instantiate}(\mathcal{P};t,z_t),
    \qquad |\mathcal{I}_t|\ll t,
    \label{eq:structured-sparse-instantiation}
\end{equation}
where $z_t$ is optional input-dependent information used only to instantiate positions within the permitted family. This distinction separates \emph{architecture-prescribed} patterns, fixed before training so that parameters adapt to the restricted support, from \emph{post-hoc discovered} patterns, inferred from an already-trained dense checkpoint and applied at inference. Figure~\ref{fig:sparse-structure-pattern-origin} visualizes these two origins through representative causal attention maps.

\subsubsection{Architecture-Prescribed Patterns}
\label{subsubsec:architecture-prescribed-sparse}

Architecture-prescribed patterns determine the attention topology before pretraining or task training, allowing the model parameters to adapt to the restricted support. A common construction combines a local backbone with a small number of long-range edges. Longformer uses a sliding window together with task-dependent global tokens that can read the full sequence and be read by all positions \citep{beltagy2020longformer}. BigBird combines local, global, and random connections to provide local continuity, global aggregation, and additional long-range graph connectivity \citep{zaheer2020bigbird}. These local--global composites preserve a predictable memory-access pattern, although special nodes and heterogeneous edge types complicate both model design and implementation.

A second line organizes long-range access directly by distance. Sparse Transformer factorizes dense attention into complementary local and strided or fixed patterns, reducing the interaction count to approximately $O(L\sqrt{L})$ for suitable factorizations \citep{child2019sparse}. LongNet uses progressively increasing segment sizes and dilation rates to form multi-scale Dilated Attention, thereby extending the receptive field while keeping the attention work linear in sequence length under its sparse schedule \citep{ding2023longnet}. PowerAttention similarly uses power-of-two offsets to allocate a fixed budget across multiple distance scales \citep{chen2025powerattention}. Distance-structured patterns are regular and budgetable, but their ability to retrieve a remote item depends on the prescribed offsets and on information propagation through depth.

The benefit of architecture-prescribed sparsity is training--inference consistency: the model learns under the same constrained Access used at deployment. The cost is limited input adaptivity. A fixed topology can retain irrelevant regions and omit evidence whose location or density does not match the structural prior. Enlarging windows or adding global edges improves coverage but directly consumes the compute and bandwidth savings that motivate sparsity.

\subsubsection{Post-hoc Discovered Patterns}
\label{subsubsec:posthoc-sparse-patterns}

Post-hoc methods analyze a trained dense model and convert recurring attention behavior or extrapolation failures into an inference-time rule. StreamingLLM identifies the importance of initial ``attention sink'' tokens and retains a small sink set together with a rolling recent window \citep{xiao2023streamingllm}. LM-Infinite similarly uses an initial-plus-recent $\Lambda$-shaped mask and caps relative-position distances to improve zero-shot length extrapolation \citep{han2023lminfinite}. Their runtime Access is globally fixed rather than content-selected: the rule preserves empirically important positions but does not retrieve arbitrary middle-context tokens.

MInference occupies an intermediate point between fixed and fully content-adaptive routing. It profiles dense attention maps to assign each head an A-shape, vertical--slash, or block-sparse pattern family, and then instantiates columns, diagonals, or blocks from the current Q/K values at inference time \citep{jiang2024minference}. The family is fixed by offline analysis, whereas the selected positions retain limited input dependence. This reduces the online search space and supports structured kernels, but the dense checkpoint was not optimized under the resulting mask.

Post-hoc discovery is attractive because it can accelerate an existing model without sparse pretraining. Its central limitation is training--inference mismatch: deleting connections and changing the Softmax normalization domain can alter a computation that was learned as dense. Fixed rules minimize selection overhead but offer little input adaptivity; pattern families recover some adaptivity at the price of runtime statistics and candidate construction. Comparisons should therefore report not only nominal sparsity, but also retained attention mass or support recall, end-to-end latency, and quality across tasks whose relevant evidence has different positional structure.

\subsection{Self-Routing Sparse Attention}
\label{subsec:self-routing-sparse}

\begin{figure}[H]
    \centering
    \includegraphics[width=\textwidth]{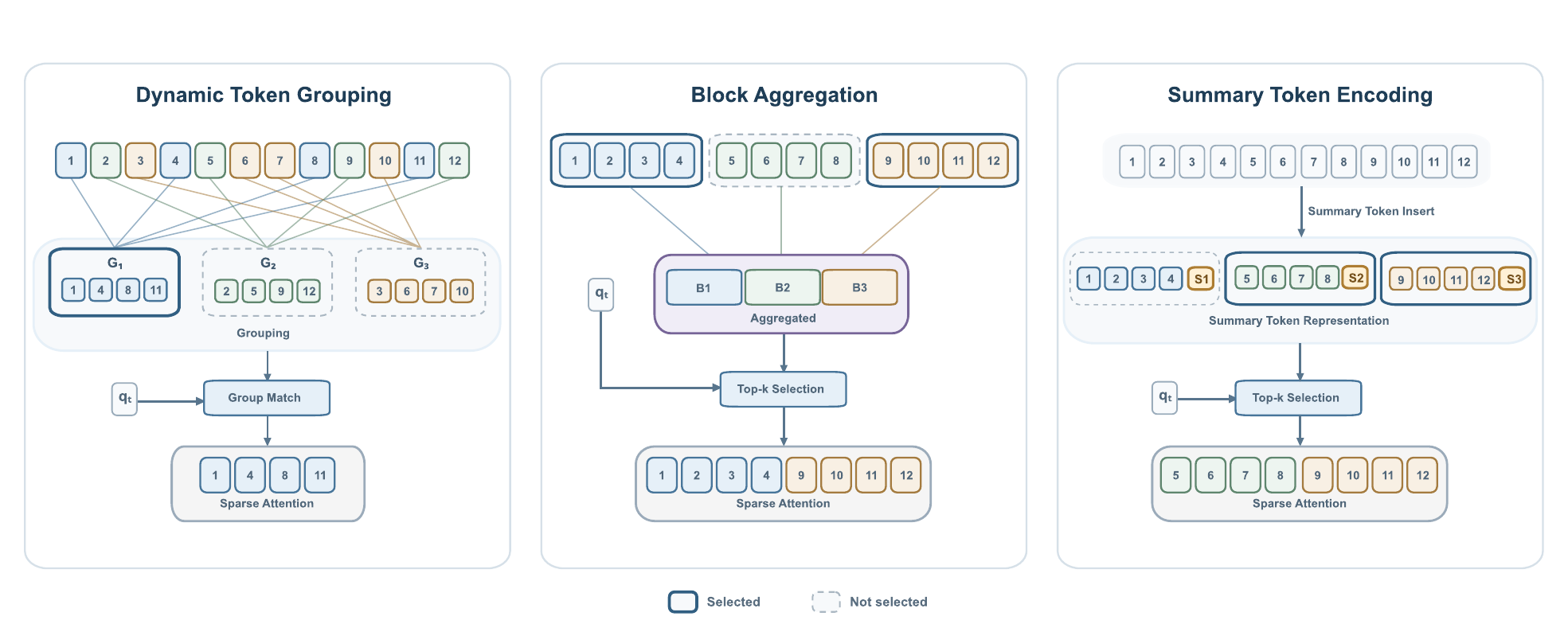}
    \caption{Representative evidence constructions for self-routing Sparse Attention: dynamic token grouping, block aggregation, and in-sequence summary-token encoding.}
    \label{fig:sparse-self-routing-evidence}
\end{figure}

Self-routing constructs support from the main model's own representations or execution state by scoring candidates in the same query--key representation space used by the main Softmax Readout, either directly or through deterministic pooling of backbone keys. Candidate evidence may additionally use sampled queries, pooled statistics, online-Softmax state, or a draft model's attention before being passed to the budgeted selector in Eq.~\eqref{eq:sparse-access}. The methods below differ in how they construct this evidence and in how tightly selection is coupled to attention execution. Figure~\ref{fig:sparse-self-routing-evidence} summarizes three representative evidence constructions developed below. Auxiliary-proxy routing, discussed in Section~\ref{subsec:auxiliary-proxy-routing}, instead forms selection scores from separately parameterized index queries and keys that the main attention does not read.

\subsubsection{Dynamic Token Grouping}
\label{subsubsec:dynamic-token-grouping}

Dynamic token grouping constructs a content-dependent sparse pattern without first evaluating the full query--key score matrix. A lightweight partitioner maps every query and key independently to one or more groups according to representation similarity. Tokens with matching group assignments are then gathered, causal masking is applied within each group, and ordinary token-level Softmax is evaluated only on those pairs. The groups may contain positions that are far apart in the original sequence, so this mechanism can connect semantically related tokens without prescribing where those tokens must occur.

Reformer realizes this idea with locality-sensitive hashing over shared query--key representations: tokens that collide under a hash attend within the same or adjacent buckets, and multiple hash rounds improve recall \citep{kitaev2020reformer}. Routing Transformer is a particularly representative learned variant because it exposes the complete grouping pipeline---learned partition, content-dependent assignment, and within-group readout---while retaining standard dot-product attention after routing \citep{roy2021routing}.

Routing Transformer maintains $G$ shared centroids $\{\mu_c\}_{c=1}^{G}$ and normalizes queries, keys, and centroids onto a common sphere. Conceptually, each query and key is assigned to its most similar centroid:
\begin{equation}
    a_t^Q
    = \operatorname*{arg\,max}_{c\in\{1,\ldots,G\}}
      \widetilde{q}_t\mu_c^{\top},
    \qquad
    a_j^K
    = \operatorname*{arg\,max}_{c\in\{1,\ldots,G\}}
      \widetilde{k}_j\mu_c^{\top},
    \label{eq:routing-transformer-assignment}
\end{equation}
where $\widetilde{q}_t$ and $\widetilde{k}_j$ are normalized routing vectors. The centroids are updated online by mini-batch spherical $k$-means, allowing the partition to follow the representation geometry learned by the model. For causal self-attention, the routing support of query $t$ consists of earlier keys assigned to the same centroid:
\begin{equation}
    \mathcal{I}_t^{\mathrm{route}}
    = \left\{j<t\;\middle|\;a_j^K=a_t^Q\right\}.
    \label{eq:routing-transformer-support}
\end{equation}
The query then performs exact Softmax attention within this support:
\begin{equation}
    r_t^{\mathrm{route}}
    = \operatorname{softmax}\!\left(
        \frac{\widetilde{q}_t
        (\widetilde{K}_{\mathcal{I}_t^{\mathrm{route}}})^{\top}}
        {\sqrt{d}}
      \right)V_{\mathcal{I}_t^{\mathrm{route}}}.
    \label{eq:routing-transformer-readout}
\end{equation}
In practice, Routing Transformer assigns each centroid roughly $L/G$ queries and keys with the highest centroid similarity rather than allowing unconstrained cluster sizes; this balances parallel work but can give a token more than one membership. Its models also allocate separate heads to local attention so nearby evidence is preserved when clustering misses it. The resulting cost is $O(LGd+L^2d/G)$ under balanced groups, minimized near $G=\sqrt{L}$ to $O(L^{3/2}d)$. This content adaptivity comes with assignment, sorting, group-size control, and load-balancing overhead, while non-contiguous members produce irregular gathers that are harder to map to accelerator tiles. These costs motivate later methods to fix the physical candidate unit as a page or contiguous block and focus on constructing an accurate low-cost score for each unit.

\subsubsection{Block Aggregation}
\label{subsubsec:self-routing-block-aggregation}

Once candidates are fixed as contiguous blocks, routing reduces to scoring each block and expanding the selected blocks to their constituent KVs. A key distinction is whether those scores are derived from an unchanged dense checkpoint or from representations optimized under sparse support. This is not a strict chronological progression: training-free designs preserve retrofitability, whereas trainable designs trade additional optimization for a smaller dense-to-sparse mismatch.

For an existing dense checkpoint that must be deployed without adaptation, block relevance can be estimated directly from its current query--key geometry. Quest bounds the largest possible query--key product within each page, while XAttention derives block scores from antidiagonal statistics \citep{tang2024quest,xu2025xattention}. Their appeal is immediate applicability without modifying model parameters; their limitation is that they approximate the attention behavior of a model never optimized for restricted support, so the dense-to-sparse mismatch can become more consequential as the retained budget shrinks.

When training from scratch or continued adaptation is feasible, the model can instead learn to make block-level evidence reliable under sparse support. MoBA partitions KVs into contiguous blocks and represents each block $\mathcal{B}_u$ by the mean of its backbone keys:
\begin{equation}
    \bar{k}_u
    = \frac{1}{|\mathcal{B}_u|}
      \sum_{i\in\mathcal{B}_u} k_i.
    \label{eq:moba-block-mean-key}
\end{equation}
For query $q_t$, it then computes the relevance of block $u$ by a query-to-mean-key inner product:
\begin{equation}
    e_{t,u}^{\mathrm{MoBA}}
    = q_t\bar{k}_u^{\top}.
    \label{eq:moba-block-score}
\end{equation}
MoBA selects the Top-$k$ historical blocks according to $e_{t,u}^{\mathrm{MoBA}}$ and always adds the current block $u_t$:
\begin{equation}
    \mathcal{S}_t^{\mathrm{MoBA}}
    = \operatorname{TopK}_{u<u_t}\!\left(
        e_{t,u}^{\mathrm{MoBA}};k
      \right)\cup\{u_t\}.
    \label{eq:moba-block-selection}
\end{equation}
Let $\mathcal{I}_t^{\mathrm{MoBA}}$ contain the causally visible token positions in the selected blocks. The final readout applies exact token-level Softmax to their original KVs:
\begin{equation}
    o_t^{\mathrm{MoBA}}
    = \operatorname{softmax}\!\left(
        \frac{q_t(K_{\mathcal{I}_t^{\mathrm{MoBA}}})^{\top}}{\sqrt{d}}
      \right)V_{\mathcal{I}_t^{\mathrm{MoBA}}}.
    \label{eq:moba-readout}
\end{equation}
Because scoring reuses the backbone's Q/K space, sparse training can adapt those representations to make the coarse block score useful \citep{lu2025moba}. FlashMoBA preserves this routing rule but makes smaller blocks practical: tiled Top-$k$ avoids materializing the full query--block score matrix, and a gather-and-densify kernel packs irregularly selected queries into on-chip tiles for FlashAttention-style computation \citep{xiao2025flashmoba}.

Native Sparse Attention (NSA), introduced by DeepSeek-AI as a hardware-aligned and natively trainable sparse architecture, uses block aggregation within a three-branch design. Let $\ell_c$, $s_c$, and $\ell_s$ denote the compression-block length, compression stride, and selection-block length, respectively. NSA first maps each completed compression block to a learned compressed key:
\begin{equation}
    \widetilde{K}_t^{\mathrm{cmp}}
    = \left\{\phi_K\!\left(k_{i s_c+1:i s_c+\ell_c}\right)
      \;\middle|\;
      0\le i\le \left\lfloor\frac{t-\ell_c}{s_c}\right\rfloor\right\}.
    \label{eq:nsa-compressed-keys}
\end{equation}
Here $\phi_K$ is a learned compression MLP with intra-block positional encoding; compressed values $\widetilde{V}_t^{\mathrm{cmp}}$ are constructed analogously. The compressed branch computes its query--key attention probabilities as
\begin{equation}
    p_t^{\mathrm{cmp}}
    = \operatorname{softmax}\!\left(
        \frac{q_t(\widetilde{K}_t^{\mathrm{cmp}})^{\top}}{\sqrt{d}}
      \right).
    \label{eq:nsa-compressed-score}
\end{equation}
NSA reuses these probabilities as routing evidence. When compression and selection blocks have different boundaries, the score of selection block $u$ aggregates the probabilities of the overlapping compression blocks:
\begin{equation}
    p_t^{\mathrm{slc}}[u]
    = \sum_{m=0}^{\ell_s/s_c-1}
      \sum_{r=0}^{\ell_c/s_c-1}
      p_t^{\mathrm{cmp}}\!\left[
        \frac{\ell_s}{s_c}u-m-r
      \right],
    \label{eq:nsa-selection-block-score}
\end{equation}
where out-of-range terms are taken as zero. For grouped-query attention, the scores of the $H_g$ query heads sharing one KV head are summed before selection:
\begin{equation}
    \widehat{p}_t^{\mathrm{slc}}[u]
    = \sum_{h=1}^{H_g}p_{t,h}^{\mathrm{slc}}[u].
    \label{eq:nsa-group-score}
\end{equation}
The fine-grained branch then retains the Top-$n$ selection blocks according to the aggregated score:
\begin{equation}
    \mathcal{S}_t^{\mathrm{NSA}}
    = \operatorname{TopK}\!\left(
        \widehat{p}_t^{\mathrm{slc}};n
      \right).
    \label{eq:nsa-block-selection}
\end{equation}
The original-token KVs in $\mathcal{S}_t^{\mathrm{NSA}}$ form the selected branch, while a sliding-window branch preserves recent local context. Input-dependent sigmoid gates weight the compressed, selected, and sliding-window branch outputs, whose weighted sum forms the final NSA readout. Thus, unlike MoBA's single selected-block path, NSA uses the compressed summaries both as readable memory and as evidence for retrieving full-resolution content \citep{yuan2025nsa}.

Other trainable variants mainly change how block evidence is summarized. InfLLM-V2 combines multi-stage mean/max pooling with GQA-group aggregation and allows dense--sparse switching; DashAttention learns hierarchical pooling with $\alpha$-entmax to allocate a variable support; and COBS compresses second-order statistics to estimate block attention mass \citep{zhao2025infllmv2,huang2026dashattention,tian2026cobs}. Across these designs, adaptation can reduce dense-to-sparse mismatch, but summary maintenance, block-boundary errors, overfetch, and additional Readout paths remain part of the end-to-end cost.

\subsubsection{Summary-Token Encoding}
\label{subsubsec:summary-token-routing}

Summary-token methods insert learned tokens into the sequence and use their backbone representations as address memory. Landmark Attention places a landmark after each context block and uses a grouped Softmax in which a token's effective probability depends on both its own key and its block landmark; the selected landmarks determine which original blocks are loaded \citep{mohtashami2023landmark}. Simplified Sparse Attention inserts gist tokens, trains them under a restricted mask to encode their chunks, and then attends over raw KVs from the selected chunks together with the selected gist representations. Its hierarchical extension adds meta-gists to reduce the cost of scanning all chunk summaries \citep{mao2026ssa}.

HiLS targets the attention mass of each distant chunk rather than only its mean or maximum token score \citep{hu2026hils}. For query $q_t$ and chunk $\mathcal{B}_u$, let $s_{t,j}=q_tk_j^{\top}/\sqrt{d}$; the exact unnormalized chunk mass is
\begin{equation}
    Z_{t,u}
    = \sum_{j\in\mathcal{B}_u}\exp(s_{t,j}).
    \label{eq:hils-exact-chunk-mass}
\end{equation}
Computing $Z_{t,u}$ for every chunk would require all token-level query--key products. HiLS therefore appends a landmark token to each chunk and uses its learned query $q'_u$ to construct a compact summary. The landmark query first induces an intra-chunk distribution, whose weighted key average becomes the summary key:
\begin{equation}
    \alpha_{u,j}
    = \frac{\exp\!\left(q'_u k_j^{\top}/\sqrt{d}\right)}
           {\sum_{r\in\mathcal{B}_u}\exp\!\left(q'_u k_r^{\top}/\sqrt{d}\right)},
    \qquad
    k'_u
    = \sum_{j\in\mathcal{B}_u}\alpha_{u,j}k_j.
    \label{eq:hils-chunk-summary}
\end{equation}
The entropy of this aggregation distribution supplies a chunk-dependent bias:
\begin{equation}
    b'_u
    = -\sum_{j\in\mathcal{B}_u}\alpha_{u,j}\log\alpha_{u,j}.
    \label{eq:hils-aggregation-entropy}
\end{equation}
For the current query, HiLS combines summary relevance and aggregation entropy into a linear surrogate for the chunk's LogSumExp score:
\begin{equation}
    \widehat{s}_{t,u}
    = \frac{q_t(k'_u)^{\top}}{\sqrt{d}}+b'_u
    \approx \log Z_{t,u},
    \qquad
    \widehat{Z}_{t,u}=\exp(\widehat{s}_{t,u})\approx Z_{t,u}.
    \label{eq:hils-surrogate-chunk-mass}
\end{equation}
It then retrieves the $K$ distant chunks with the largest surrogate scores:
\begin{equation}
    \mathcal{S}_t^{\mathrm{HiLS}}
    = \operatorname{TopK}_{u\in\mathcal{C}_t}\!\left(
        \widehat{s}_{t,u};K
      \right),
    \label{eq:hils-chunk-selection}
\end{equation}
where $\mathcal{C}_t$ contains historical chunks outside the local window. HiLS computes exact token-level Softmax within each selected chunk, but uses the surrogate masses to allocate probability across chunks. If $Z_{t,\mathrm{loc}}$ is the exact mass of the local window and $c(j)$ denotes the chunk containing token $j$, then for a token in a selected distant chunk,
\begin{equation}
    w_{t,j}^{\mathrm{HiLS}}
    = \frac{\exp(s_{t,j})}{Z_{t,c(j)}}
      \frac{\widehat{Z}_{t,c(j)}}{\widehat{\mathcal{Z}}_t},
    \qquad
    \widehat{\mathcal{Z}}_t
    = Z_{t,\mathrm{loc}}
      + \sum_{u\in\mathcal{S}_t^{\mathrm{HiLS}}}\widehat{Z}_{t,u}.
    \label{eq:hils-hierarchical-readout}
\end{equation}
The local window is treated as one group with its exact mass $Z_{t,\mathrm{loc}}$, and the final output is the weighted sum of values over the local window and selected chunks. Because $\widehat{Z}_{t,u}$ participates in the forward attention weights rather than only in the discrete Top-$K$ decision, the language-modeling loss can train the landmark summaries end to end. This example illustrates why Access and Readout must be separated: a summary score may merely choose candidates, or it may also determine how probability mass is distributed across candidate groups. Summary tokens remain semantically aligned with the backbone, but they require a specialized sequence layout and training procedure, consume additional token capacity, and can become an information bottleneck when one vector must represent a long or heterogeneous chunk.

\subsection{Auxiliary-Proxy Sparse Routing}
\label{subsec:auxiliary-proxy-routing}

\begin{figure}[t]
    \centering
    \includegraphics[width=\textwidth]{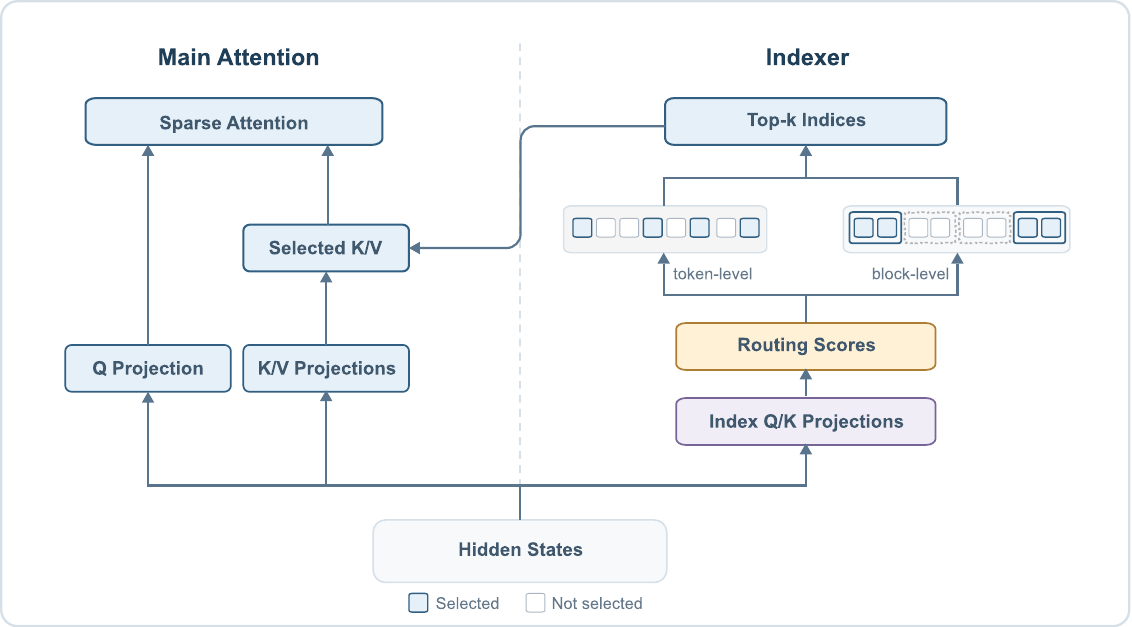}
    \caption{Auxiliary-proxy Sparse Attention within one layer. Shared hidden states feed the Main Attention and Indexer paths; token- or block-level index selections determine which K/V entries reach Sparse Attention.}
    \label{fig:sparse-proxy-indexer}
\end{figure}

Auxiliary-proxy routing also selects support from the current content, but introduces an address representation whose parameters and training objective are distinct from the main attention projections. A lightweight indexer typically forms the proxy query and key through separate projections of backbone hidden states, for example $q_t^I=h_tW_Q^I$ and $k_u^I=h_uW_K^I$ at token granularity; block-level methods may instead pool these proxy keys or construct one compressed proxy entry per block. The index-specific parameters $W_Q^I$ and $W_K^I$ are learned independently of the main attention projections. The resulting proxy scores $e_{t,u}^I=s_I(q_t^I,k_u^I)$ determine which original K/V entries are selected. Depending on the method, these scores may be discarded after Access or may also influence Readout. Figure~\ref{fig:sparse-proxy-indexer} makes the internal split explicit: the indexer maps query and candidate states into a compact address space, constructs selection evidence, and applies a budgeted selector; only the returned indices address the original K/V memory. For block routing, evidence can be formed by scoring pooled proxy keys or by aggregating token scores, so block-granular Access does not by itself imply a block-granular evidence scan. Decoupling makes the dimensionality, number of heads, update schedule, supervision, and routing granularity independently tunable. It also creates a second memory whose fidelity to the content-memory relevance relation must be learned and maintained.

\subsubsection{Token-Granularity Routing}
\label{subsubsec:proxy-token-routing}

Token-granularity proxies maintain a separately addressable entry for each historical token. TokenButler projects hidden states and cached keys into a low-dimensional importance space, distills the proxy from dense attention distributions, and amortizes predictions through intervals and neighbor fetching \citep{akhauri2025tokenbutler}.

DeepSeek Sparse Attention (DSA) from DeepSeek-V3.2 introduces a lightweight Lightning Indexer for routing \citep{deepseek2025v32}. The indexer has an MQA-style structure in which several index query heads share a single index key per token (left column of Figure~\ref{fig:sparse-proxy-aggregation}). For query $t$ and a preceding token $u$, its proxy score is
\begin{equation}
    e_{t,u}^{I}
    = \sum_{h=1}^{H_I} w_{t,h}^{I}\,
      \operatorname{ReLU}\!\left(
        q_{t,h}^{I}(k_u^{I})^{\top}
      \right),
    \label{eq:dsa-index-score}
\end{equation}
where $H_I$ is the number of index heads, $q_{t,h}^{I}$ is the query of index head $h$, $k_u^{I}$ is the index key shared by all heads, and the query-dependent scalar weights $w_{t,h}^{I}$ fuse the per-head scores into one token-level score. The Top-$k$ tokens under $e_{t,:}^{I}$ form a single support shared by all attention heads, because the main attention runs MLA in its MQA mode, where each latent KV entry serves every query head; the index scores only determine this support and do not enter Readout. The indexer is supervised by a KL loss toward the main attention scores summed over heads and renormalized over tokens: a short dense warmup trains only the indexer over the full context, after which the loss is restricted to the selected tokens while the whole model adapts to sparse support, and the indexer input is detached so that the indexer and the backbone learn only from the KL and language-modeling losses, respectively.

LongCat Sparse Attention (LSA) extends the DSA Lightning Indexer to relieve two of its deployment bottlenecks, the full-length index scan and the discontinuous KV access induced by token-level selection. Streaming-Aware Indexing reserves part of the support budget for a fixed sink and sliding window while using the remainder for dynamic token selection, improving the contiguity of KV access. Hierarchical Indexing first recalls candidate blocks with approximate scores and then performs fine-grained token selection within those blocks, reducing index computation while retaining token-level final support \citep{zan2026longcatsparse}.

SAS learns a continuous context-ranking gate and incorporates the gate score into attention logits, allowing the language-model objective to optimize both ranking under a finite budget and the subsequent Readout \citep{li2026sas}. TokenButler, DSA, and SAS thus respectively illustrate plug-in distillation, native mask-oriented indexing, and end-to-end coupling between proxy evidence and readout weights.

Binary proxies reduce the storage and comparison cost of each address entry. HashAttention learns mappings from queries and key--value pairs to a Hamming space and retrieves pivotal tokens by binary similarity before applying attention to the selected original KVs \citep{desai2024hashattention}. HATA trains query and key codes to preserve the relative ordering required for Top-$k$ selection and co-designs binary scoring with a sparse-attention kernel \citep{gong2025hata}. Binary coding lowers per-entry cost, but it does not change the $O(L)$ number of token-level index entries. The router must still scan or retrieve candidates and then gather KVs from potentially scattered positions.

Token routing minimizes block overfetch and can recover isolated evidence. Its limiting resource at very long context may nevertheless be index bandwidth rather than arithmetic. A low-dimensional or binary proxy is useful only if its reduced comparison cost exceeds the cost of reading the full proxy sequence, performing Top-$k$, and gathering irregular content memory.

\subsubsection{Block-Granularity and Compressed-Entry Routing}
\label{subsubsec:proxy-block-routing}

Block-granularity routing uses a contiguous unit for final selection and content access. SeerAttention learns a lightweight predictor over pooled Q/K representations and distills block importance from dense attention; the resulting AttnGate chooses original-token KV blocks \citep{gao2024seerattention}. SeerAttention-R extends the mechanism to decoding, shares selection within GQA groups, and maintains a compressed key index \citep{gao2025seerattentionr}. SpotAttention learns a calibrated block distribution and uses dual Top-$p$ rules to allocate different budgets across queries and layers \citep{ahmad2026spotattention}. These methods improve access regularity while retaining the full-resolution KVs inside selected blocks, but all three train the router as a plug-in on a frozen pretrained model, so the backbone itself never adapts to block-sparse support.

\begin{figure}[H]
    \centering
    \includegraphics[width=\textwidth]{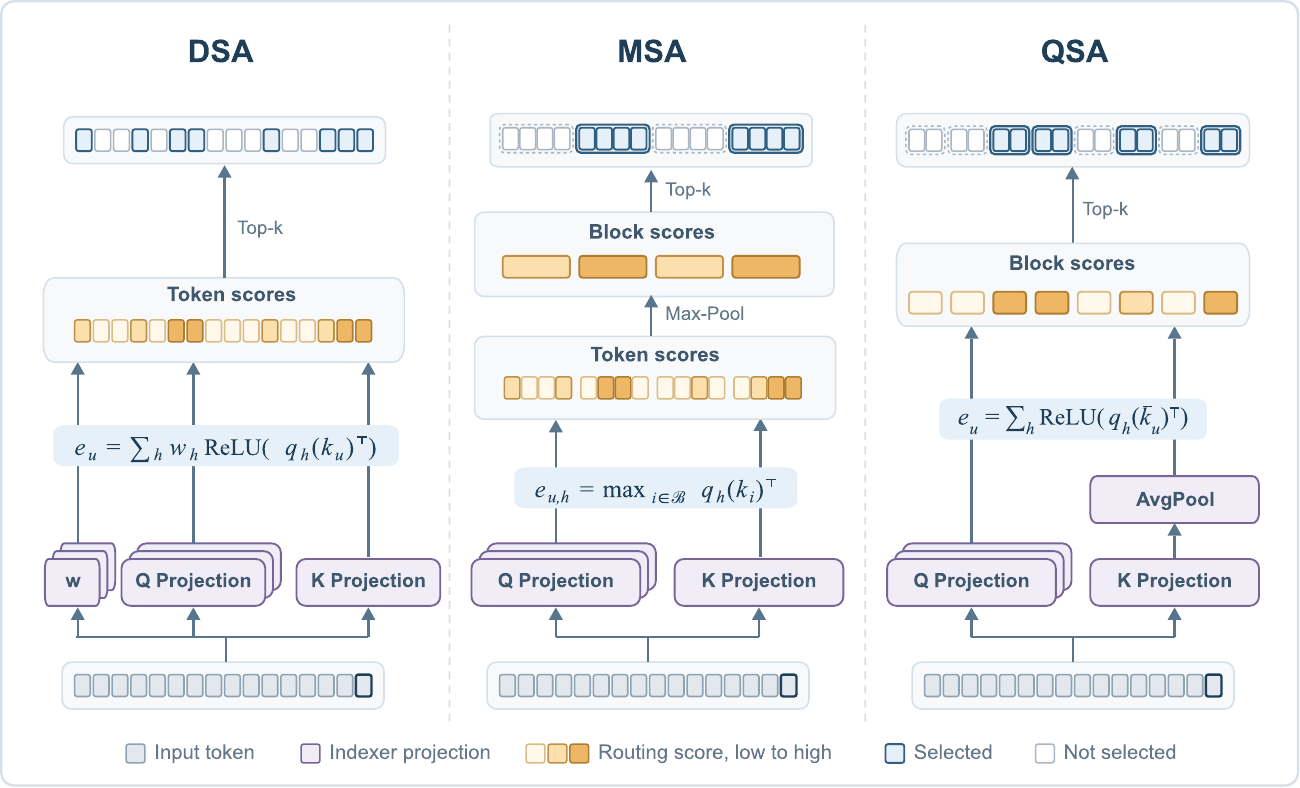}
    \caption{Indexer structures of DSA, MSA, and QSA, which differ mainly in where aggregation enters the index-score computation. The in-figure formulas are simplified; Eqs.~\eqref{eq:dsa-index-score}--\eqref{eq:qsa-block-score} give the full definitions.}
    \label{fig:sparse-proxy-aggregation}
\end{figure}

MiniMax Sparse Attention (MSA) instead trains the backbone under its block-sparse support. MSA is a block-sparse extension of GQA used in MiniMax-M3, a natively multimodal MoE model with about 428B total and 23B activated parameters that applies MSA in 57 of its 60 layers \citep{minimax2026msa,minimax2026m3}. Its Index Branch shares the MQA-style structure of the DSA Lightning Indexer (middle column of Figure~\ref{fig:sparse-proxy-aggregation}) but differs in the role of the index heads: instead of fusing them into one support shared by all attention heads, MSA assigns one index head to each GQA group, so each index head selects support independently for its own group. Index scores are computed at token level, between the index query and every visible index key, whereas selection operates at block level: for index head $h$, each contiguous KV block $u$ is scored by the maximum of its token scores,
\begin{equation}
    e_{t,u,h}^{I}
    = \max_{i\in\mathcal{B}_u,\; i\le t}
      \frac{q_{t,h}^{I}(k_i^{I})^{\top}}{\sqrt{d_I}},
    \label{eq:msa-block-score}
\end{equation}
where $\mathcal{B}_u$ is the set of token positions in block $u$ and $d_I$ is the index dimension. The Top-$k$ blocks under $e_{t,:,h}^{I}$ are kept together with the block containing the query, over which the query heads of the corresponding GQA group compute exact Softmax attention. Supervision follows the DSA recipe---a KL loss toward the Main Branch attention on the selected tokens, preceded by a full-attention warmup and confined to the index projections by stop-gradients---except that the teacher is averaged over the query heads of each group.

MSA thus uses block-granular Access but a token-granular evidence scan, trading the token-level localization of DSA for regular per-group KV reads. Because the index scan still touches every visible token, its cost continues to grow with context length and ultimately bounds the attainable speedup; regular KV reads do not by themselves reduce the computation and bandwidth consumed by the proxy.

Compressed-entry routing changes the address representation more aggressively. Qwen Sparse Attention (QSA) compresses the index key sequence into micro-block representations before scoring, thereby shrinking the index scan itself rather than only the final selection \citep{qwen2026qwen38next,qwen2026qwen38flashnext}. In Qwen3.8-Flash-Next (Table~\ref{tab:top-ranked-open-weight-attention}), QSA replaces the full-attention layers during continued pretraining, with one QSA layer following every three Gated DeltaNet layers. Its indexer shares the MQA-style structure of MSA, but where MSA pools token scores, QSA pools the index keys themselves: the shared key sequence is average-pooled over non-overlapping micro-blocks before positional encoding, so each micro-block receives one content summary and one block position. For query $t$ and micro-block $u$,
\begin{equation}
    \bar{k}_u^{I}
    = \operatorname{AvgPool}\!\left(\{k_i^{I}\}_{i\in\mathcal{B}_u}\right),
    \qquad
    e_{t,u}^{I}
    =
    \begin{cases}
        \displaystyle\sum_{h=1}^{H_I}\operatorname{ReLU}\!\left(q_{t,h}^{I}(\bar{k}_u^{I})^{\top}\right), & \max\mathcal{B}_u\le t,\\
        -\infty, & \text{otherwise},
    \end{cases}
    \label{eq:qsa-block-score}
\end{equation}
so per-head scores are summed without the query-dependent weights of Eq.~\eqref{eq:dsa-index-score}, and only fully observed micro-blocks are scored. Even though the core attention uses GQA, QSA does not select per group as MSA does: the fused score $e_{t,u}^{I}$ yields a single selection that all GQA groups share. The Top-$k$ micro-blocks under $e_{t,:}^{I}$ are expanded to their original token positions and joined with the tokens of the current incomplete micro-block, over which core attention reads original-token KVs. Training follows the two stages of DSA---indexer-only dense distillation, then sparse training of the whole model with the KL restricted to the selected blocks---except that the token-level teacher distribution is max-pooled within each micro-block and renormalized to match the block-level scores. Because the index scan shrinks by the micro-block size while content stays at token resolution, QSA compresses addresses without compressing content. In short, DSA and MSA both score every visible token and differ in whether index heads are fused into token scores or token scores are max-pooled into per-group block scores, whereas QSA pools the keys before scoring and is the only one of the three that shortens the index scan.

Compressed Sparse Attention (CSA) goes further and compresses content as well \citep{deepseek2026v4}. Introduced in DeepSeek-V4, where it alternates with Heavily Compressed Attention (HCA) layers that compress more aggressively and attend densely, CSA merges the KV entries of every $m$ tokens into one entry by learned, overlapping weighted pooling, applies the same compression to the index keys, and scores the compressed blocks with the DSA indexer:
\begin{equation}
    \bar{c}_u
    = \sum_{i\in\mathcal{B}_{u-1}\cup\mathcal{B}_u} \pi_{u,i}\odot c_{u,i},
    \qquad
    e_{t,u}^{I}
    = \sum_{h=1}^{H_I} w_{t,h}^{I}\,
      \operatorname{ReLU}\!\left(q_{t,h}^{I}(\bar{k}_u^{I})^{\top}\right),
    \label{eq:csa-compressed-score}
\end{equation}
where $c_{u,i}$ is a per-token projection serving as both key and value, $\pi_{u,i}$ are channel-wise softmax weights over the current and preceding blocks, and $\bar{k}_u^{I}$ is the compressed index key. The Top-$k$ compressed entries, together with an uncompressed sliding window that covers the query's incomplete block, form the eligible content view used by Readout. Because this view holds compressed entries rather than token KVs, CSA reduces index length, attention work, and resident content memory at the cost of item-level fidelity. QSA and CSA together show why Memory Representation must distinguish address compression from content compression.

Block selection offers fewer routing units and more regular gathers, but it reads irrelevant tokens whenever importance is concentrated inside a block. If evidence is still computed at token resolution, the proxy scan remains linear in the number of tokens. If both the index and content are compressed by a factor $m$, the scanned sequence can shrink from $L$ to approximately $L/m$, but the reduction is purchased with coarser routing resolution and, for compressed content, reduced recoverability of individual details.

\subsection{Temporal and Cross-Layer Reuse}
\label{subsec:sparse-reuse}

Once content-adaptive Access can produce a useful support, a complementary optimization is to avoid recomputing similar decisions at every decoding step and layer. Two distinct objects can be reused. Temporal support reuse uses a history of prior query--support relations to propose candidates for a new decoding query, after which a method may rerank or refresh them. Cross-layer reuse lets a non-anchor layer inherit selected positions from an anchor layer; more aggressive designs also let it read K/V content produced by a source layer. These mechanisms change when Access decisions and content states are refreshed rather than defining a new evidence representation. They can therefore be layered onto structure-, self-, or proxy-routed candidate generation and are treated here as cross-cutting lifecycle attributes rather than a mutually exclusive family.

\subsubsection{Temporal Support Reuse}
\label{subsubsec:temporal-support-reuse}

Temporal support reuse exploits the observation that nearby or semantically similar decoding queries often retrieve many of the same historical tokens. Rather than accepting an earlier support as the current answer, these methods use past selections only to construct a smaller candidate set. The current query then recomputes exact QK scores over those candidates, selects its own support, and applies ordinary sparse Softmax to the original KVs. Temporal reuse therefore reduces the cost of discovering candidates without reusing stale attention weights or discarding the full KV cache.

ReTopK is a representative recall-before-rerank design \citep{yao2026retopk}. For each query head, it maintains a bounded FIFO cache of normalized historical queries $\bar{q}_\tau=q_\tau/\|q_\tau\|_2$ together with their selected supports $\widehat{\mathcal{I}}_\tau$. Given a new query, it first computes cosine similarity to each cached query,
\begin{equation}
    \rho_{t,\tau}
    = \bar{q}_t\bar{q}_\tau^{\top},
    \label{eq:retopk-query-similarity}
\end{equation}
and denotes the indices of the $R$ most similar cache entries by $\mathcal{R}_t$. Their historical supports are merged with a recent window $\mathcal{L}_t$ to form the candidate set
\begin{equation}
    \mathcal{A}_t
    = \mathcal{L}_t
      \cup \bigcup_{\tau\in\mathcal{R}_t}\widehat{\mathcal{I}}_\tau.
    \label{eq:retopk-candidate-recall}
\end{equation}
The recent window keeps newly appended tokens eligible even though they cannot yet appear in a historical support. ReTopK next evaluates the current query against only the recalled candidate keys and reranks them:
\begin{equation}
    \widehat{\mathcal{I}}_t
    = \operatorname{TopK}_{i\in\mathcal{A}_t}\!\left(
        \frac{q_tk_i^{\top}}{\sqrt{d}};K
      \right).
    \label{eq:retopk-reranking}
\end{equation}
The selected indices are then used in the candidate-local Softmax of Eq.~\eqref{eq:sparse-local-readout}. Thus, ReTopK reuses only historical support identities for candidate recall; all retained scores, attention probabilities, and value aggregation are recomputed from the current query.

The main failure mode is support drift: similar queries need not require identical evidence, and errors can accumulate when supports produced by reuse are inserted back into the cache. ReTopK therefore falls back to full-history exact Top-$K$ when the maximum cached-query similarity is below a threshold and periodically performs an exact refresh. Its efficiency depends on the cache capacity, number $R$ of recalled supports, recent-window width, fallback rate, and refresh interval. If these quantities remain bounded, reranking touches at most the union of $R$ supports and the recent window rather than the full history; if fallback is frequent or the union becomes large, the advantage correspondingly shrinks.

\subsubsection{Cross-Layer Reuse}
\label{subsubsec:cross-layer-support-reuse}

The simplest cross-layer strategy shares selected positions. TidalDecode observes persistence in selected positions across adjacent layers: a small set of selection layers performs full attention and generates indices, while intervening layers reuse them \citep{yang2024tidaldecode}. Kascade similarly computes exact Top-$k$ only at calibrated anchor layers and uses dynamic programming to choose where those anchors should be placed \citep{deshmukh2025kascade}.

IndexCache applies this principle directly to the learned indexers of DSA. It assigns every layer one of two roles: an \emph{F} (Full) layer runs its indexer over the full history and caches fresh Top-$k$ indices, whereas an \emph{S} (Shared) layer omits its indexer and inherits the indices from the nearest preceding F layer. If $c_\ell\in\{\mathrm{F},\mathrm{S}\}$ denotes the role of layer $\ell$, the support is
\begin{equation}
    \mathcal{I}_t^{(\ell)}
    =
    \begin{cases}
        \operatorname{TopK}\!\left(e_{t,:}^{I,(\ell)};k\right),
            & c_\ell=\mathrm{F},\\
        \mathcal{I}_t^{(f(\ell))},
            & c_\ell=\mathrm{S},
    \end{cases}
    \qquad
    f(\ell)=\max\{j<\ell:c_j=\mathrm{F}\}.
    \label{eq:indexcache-layer-reuse}
\end{equation}
The first layer is always Full, and each later Full layer overwrites the temporary index cache, so reuse adds no persistent index buffer beyond the one already required by DSA. The training-free variant greedily chooses which indexers to retain by measuring language-modeling loss on a calibration set, because uniformly spaced Full layers can remove unusually important indexers. The training-aware variant instead distills each retained indexer from the attention distributions of all layers that will share its output, encouraging one support to serve the whole layer group rather than only its anchor layer \citep{bai2026indexcache}. GLM-5.2 deploys this mechanism under the name IndexShare, with one indexer shared by every four sparse-attention layers; GLM-5.3 retains the same base architecture and therefore the same cross-layer indexing design \citep{glm2026glm52,glm2026glm53}.

More extensive reuse shares content memory as well as discrete indices. YOIO/CLSA starts from a self-decoder/cross-decoder architecture: the self-decoder writes one cross-layer KV memory, a single-head query-aware indexer computes token-level Top-$k$ routing once over that memory, and every cross-decoder layer reads its own query-dependent values from the same selected support. A multi-layer distillation target averaged over decoder layers and attention heads trains the shared indexer to preserve tokens that are jointly useful across the stack \citep{sun2026yoio}.

Compressed Sparse Attention 2 (CSA2) was introduced by DeepSeek-AI in the technical report \emph{DeepSeek-V4.1-Flash: Pushing the Limits of KV Cache Compression} as the successor to the CSA mechanism used in DeepSeek-V4 \citep{deepseek2026v41flash}. It targets two remaining long-context costs jointly: storing separate global KVs at every layer and repeatedly scanning the full index at every layer. CSA2 therefore shares main KVs and indexer keys across depth while allowing the schedule for refreshing sparse indices to be controlled separately.

Every CSA2 layer is statically assigned one of three modes. A Full layer computes fresh main KVs and indexer keys, scans all causally visible positions, and produces fresh Top-$k$ indices. In the decoder, the first Full layer also builds a larger candidate pool $\mathcal{P}_t^{\mathrm{CSA2}}$ that is reused by later indexing layers. A Reindex layer reuses the main KVs and indexer keys from the most recent Full layer, but forms its own index query and refreshes the Top-$k$ support within that shared candidate pool. A Reuse layer skips index scoring and inherits the latest support produced by either a Full or Reindex layer. If $m_\ell\in\{\mathrm{F},\mathrm{R},\mathrm{U}\}$ is the mode of layer $\ell$ and $r(\ell)$ is its nearest preceding index-producing layer, the reuse schedule is
\begin{equation}
    \mathcal{I}_t^{(\ell)}
    =
    \begin{cases}
        \operatorname{TopK}_{i\le t}\!\left(e_{t,i}^{I,(\ell)};k\right),
            & m_\ell=\mathrm{F},\\
        \operatorname{TopK}_{i\in\mathcal{P}_t^{\mathrm{CSA2}}}\!\left(e_{t,i}^{I,(\ell)};k\right),
            & m_\ell=\mathrm{R},\\
        \mathcal{I}_t^{(r(\ell))},
            & m_\ell=\mathrm{U}.
    \end{cases}
    \label{eq:csa2-layer-modes}
\end{equation}
All three modes still compute their own main query and local sliding-window KVs. Full layers refresh shared content, address entries, and support; Reindex layers refresh only the support against reused address entries; and Reuse layers refresh none of them. YOIO/CLSA therefore provides the simpler shared-memory baseline before CSA2 introduces selective reindexing and hierarchical candidate narrowing; this conceptual progression also matches their publication order.

HySparse alternates occasional full-attention layers with groups of sparse layers. It extracts block-level maximum attention scores from a full layer, reuses them as selection evidence for later sparse layers, shares the full layer's KVs, and preserves local information through a separate sliding-window branch \citep{zhao2026hysparse}. HySparse2 refines the sparse-attention path in three ways: it replaces block-level retrieval with token-level selection to avoid block overfetch, reuses both the full layer's KVs and selected indices across the following sparse layers, and forces a recent window into the selected token set. Local and retrieved tokens therefore use one unified sparse-attention operation over a shared KV cache rather than separate global and sliding-window branches \citep{wei2026hysparse2}. This design retains fine-grained retrieval while amortizing index computation and KV storage across each group of layers.

Reuse amortizes selection and can also reduce duplicated memory when index and content states are shared. The corresponding loss is specificity. Temporal reuse can lag behind changes in the current query; cross-layer reuse can suppress differences in what shallow and deep layers need to retrieve. Selection or reindex frequency, reuse span, representation compatibility, and correction policy must therefore be reported alongside accuracy and throughput. A claimed speedup should also separate savings in router computation from savings in index traffic, KV traffic, and resident memory.

\subsection{Summary}
\label{subsec:sparse-attention-summary}

Across these families, the design trajectory reflects a shift in \emph{who decides where to attend}. The earliest designs fixed the support structurally---local, strided, or dilated patterns set before training \citep{child2019sparse,beltagy2020longformer,zaheer2020bigbird,ding2023longnet}---and later sink-plus-window rules or profiled pattern families applied structural constraints to trained dense checkpoints \citep{xiao2023streamingllm,jiang2024minference}. Self-routing restored content adaptivity by letting the backbone score candidates from its own representations, progressing from hashing or clustering \citep{kitaev2020reformer,roy2021routing} and training-free block estimation on a dense checkpoint \citep{tang2024quest,xu2025xattention} to training the model \emph{under} its own sparsity, where learned block summaries drive selection \citep{yuan2025nsa,lu2025moba}. Auxiliary-proxy routing then moved address computation into a separate, distilled index, tunable in dimensionality and granularity and coarsened from tokens to blocks and compressed entries to shrink the index itself \citep{gao2024seerattention,deepseek2025v32}. As context windows reach one million tokens and beyond, this per-layer index scan becomes the bottleneck, pushing the frontier from constructing a support to managing its lifecycle---sharing address and content memory and reusing selections across steps and layers \citep{deepseek2026v4,deepseek2026v41flash,sun2026yoio}.

This trajectory has steadily narrowed the gap to dense attention. Training-free sparsity is ultimately an approximation and inherits a training--inference mismatch, so its quality degrades once the retained budget drops below the evidence a task needs. Methods trained under their own sparse support behave differently: because the parameters adapt to a finite candidate set, natively trained selective attention and distilled-then-continued indexers now report quality on par with dense attention---and, at long context, occasionally above it---while cutting both prefill and per-step decoding cost \citep{yuan2025nsa,lu2025moba,deepseek2025v32}. This parity is conditional rather than universal: it holds when the router is trained end-to-end, the budget matches the task's evidence density, and the reported speedup accounts for evidence construction, index traffic, and gathering rather than only the final sparse matrix multiplication. Within those conditions, trained sparse attention is a credible option for long-context deployment rather than merely a lossy fallback, although its margin over dense attention remains task-, budget-, and implementation-dependent.

Accordingly, sparse attention has moved from research prototypes into production LLMs, where deployed designs use auxiliary-proxy routing at token (DSA), block (MSA), or compressed-entry (QSA and CSA) granularity, and a subset of recent releases adds cross-layer reuse. DeepSeek-V3.2 pairs each latent token with a low-dimensional index key scored by a Lightning Indexer, aligned to dense attention by distillation and then trained under sparse support \citep{deepseek2025v32}; DeepSeek-V4 aggregates index keys and KV entries into a compressed global memory combined with a sliding window \citep{deepseek2026v4}; and DeepSeek-V4.1-flash organizes layers into Full, Reindex, and Reuse modes that share indexer keys and KVs along depth \citep{deepseek2026v41flash}. Qwen3.8-flash-next compresses token routing keys into micro-block indices that expand to original-token KVs \citep{qwen2026qwen38flashnext}, MiniMax scores token-level index keys and max-pools them into block selections \citep{minimax2026msa}, and later GLM designs incorporate cross-layer index sharing \citep{glm2026glm52}. LongCat-2.0 and LongCat-Flash-Lite-Sparse deploy LSA, combining streaming-aware support, hierarchical index scoring, and cross-layer index reuse \citep{zan2026longcatsparse,meituan2026longcat2}. The emerging recipe---a learned, compact index, optionally compressed across positions, aligned through native training or distillation and continued training, and increasingly reused across depth---shows that the mechanism-level developments surveyed here now underpin mainstream long-context architectures.

% !TEX root = ../main.tex
\section{Linear Attention}
\label{sec:linear-attention}

Softmax Attention preserves contextual history as an enumerable collection of token- or chunk-level key--value representations, whereas Sparse Attention primarily restricts which of these explicit memory units are eligible for a query. Linear Attention changes the memory substrate itself: by exploiting factorizable query--key interactions, it accumulates historical key--value associations into one or more recurrently maintained associative states without retaining the complete token-wise KV history. The canonical formulation maintains a context-length-independent state, while later variants enlarge or partition the associative space, preserve multiple states over different temporal ranges, or construct block-level associative summaries. Linear Attention therefore trades item-wise historical addressability for recurrent compression, shifting the central design problem from selecting among stored tokens to determining how compressed associative memory is represented, updated, accessed, and read.

Because the term \emph{Linear Attention} covers several related formulations, we focus here on causal mechanisms that represent contextual history through one or more recurrently maintained associative states. Starting from the kernel-factorized formulation of Linear Transformer, we examine how this memory is accumulated, retained, corrected, expanded, accessed, and read. This perspective also includes gated Linear RNNs whose principal operation is associative state editing. Methods developed primarily from structured state-space dynamics are discussed in Section~\ref{sec:state-space-models}, while architectures that combine Linear Attention with explicit attention or other heterogeneous sequence mixers are examined in Section~\ref{sec:hybrid-architectures}.

Linear Transformer provides the canonical operator. Suppose a kernelized similarity admits the separable form
\(
K(q_t,k_i)=\phi(q_t)^{\top}\phi(k_i)
\).
For normalized causal Linear Attention, historical key--value information can be accumulated into an associative matrix and a normalization state,
\begin{equation}
    S_t=S_{t-1}+\phi(k_t)v_t^{\top},
    \qquad
    z_t=z_{t-1}+\phi(k_t),
    \label{eq:linear-state-basic}
\end{equation}
and read as
\begin{equation}
    o_t=
    \frac{\phi(q_t)^{\top}\widehat S_t}
         {\phi(q_t)^{\top}\widehat z_t+\varepsilon},
    \label{eq:linear-read-basic}
\end{equation}
where \((\widehat S_t,\widehat z_t)\) denotes the state visible under the model's causal read--write convention and \(\varepsilon\) represents an optional numerical stabilizer. The usual normalized form assumes a feature map and kernel for which the denominator is well defined; later recurrent variants may instead use unnormalized state contractions, output normalization, or learned gates. During autoregressive decoding, the canonical formulation maintains only \((S_t,z_t)\), while parallel or chunkwise algorithms can expose the same recurrence during training \citep{katharopoulos2020linear,schlag2021fastweight}.

Within the framework introduced in Section~\ref{sec:unified-memory-centric-view}, the development of Linear Attention begins with a fundamental change in Memory Representation and Memory Update. Its canonical formulation replaces an enumerable token-wise history with an associative matrix state and recurrently incorporates new key--value associations into this compressed representation. Later methods refine the state transition through retention, correction, erasure, and controlled writing, while enlarging or reorganizing the maintained memory through dense dimensional expansion, sparse address spaces, multiple temporal states, and block-level associative summaries. These developments also reshape the other stages of memory processing. A differentiated state organization requires Access to determine which maintained units are exposed to the current query, Readout to decode and aggregate the eligible state information, and Integration to coordinate completed readouts when multiple heads, states, or memory paths contribute to the module output. The five-dimensional view therefore reveals a coupled design trajectory: changes in how associative memory is represented and updated progressively alter what can be accessed, how it can be read, and how the resulting contextual information is incorporated into the network.

Following this trajectory, we first examine Memory Update within a single associative state, from additive accumulation to increasingly controlled forms of retention, correction, erasure, and writing. We then turn to Memory Representation and study how associative memory is enlarged or reorganized through dense dimensional expansion, sparse address spaces, temporal state collections, and block-level summaries. On this basis, we analyze how differentiated memory units introduce new Access and Readout choices, including routing among maintained states, normalized state decoding, and aggregation across multiple summaries. We finally consider how completed readouts are integrated within a layer and how memory-derived signals may be coordinated across network depth. Throughout the section, the functional analysis is paired with the corresponding computational implications, including persistent state growth, routing and merging overhead, parallel training form, and the maturity of the available empirical evidence.

\subsection{Memory Update: From Accumulation to Controlled State Editing}
\label{subsec:linear-memory-update}

For comparison across single-state methods, we use an analytical decomposition into retain/decay, erase/correct, and write/commit. This decomposition identifies distinguishable update roles; it does not assert that every method implements three independent modules or applies them in an identical computational order. For a model with $H$ attention heads, the complete memory is $S_t\in\mathbb{R}^{H\times d_k\times d_v}$, and head $h$ maintains $S_t^h\in\mathbb{R}^{d_k\times d_v}$. To simplify notation, the head index is omitted below and $S_t\in\mathbb{R}^{d_k\times d_v}$ denotes a single-head state. Its update is summarized as
\begin{equation}
    \bar S_{t-1}=\alpha_t\odot S_{t-1},
    \qquad
    S_t=\bar S_{t-1}-E_t+W_t,
    \label{eq:linear-update-decomposition}
\end{equation}
where $\bar S_{t-1}$ is the retained old state, $E_t$ is the association erased or corrected from that state, and $W_t$ is the current write.

\subsubsection{Retain / Decay}
\label{subsubsec:linear-retain-decay}

Retain / Decay controls how much of the old state remains before new information is written. Existing methods differ in both the source and the granularity of $\alpha_t$. The coefficient can be a fixed parameter after training or can be generated dynamically from the current input through $f_{\alpha}(x_t)$. Its granularity can be a head-wise scalar $\alpha_t^h\in\mathbb{R}$ or a channel-wise vector $\alpha_t^h\in\mathbb{R}^{d_k}$ acting on the $d_k$ state rows. The development begins with complete retention without decay, then introduces fixed decay with predefined time scales, and finally moves to input-dependent, fine-grained retention. Table~\ref{tab:linear-retain} summarizes this progression.

\begin{table}[H]
    \centering
    \small
    \caption{Evolution of Retain / Decay mechanisms in Linear Attention.}
    \label{tab:linear-retain}
    \begin{tabularx}{\textwidth}{@{}>{\raggedright\arraybackslash}p{0.25\textwidth}>{\raggedright\arraybackslash}p{0.12\textwidth}>{\raggedright\arraybackslash}p{0.25\textwidth}>{\raggedright\arraybackslash}X@{}}
        \toprule
        \textbf{Retain category} & \textbf{Period} & \textbf{Unified form} & \textbf{Representative methods} \\
        \midrule
        No decay & 2020--2025 & $\alpha_t^h=1$ & Linear Transformer, DeltaNet, DeltaProduct \\
        Fixed head-wise scalar & 2023--2024 & $\alpha^h\in\mathbb{R}$ & RetNet, Lightning Attention-2 \\
        Fixed channel-wise vector & 2024 & $\alpha^h\in\mathbb{R}^{d_k}$ & RWKV-5 \\
        Input-dependent head-wise scalar & 2024 & $\alpha_t^h=f_{\alpha}(x_t)\in\mathbb{R}$ & GDN \\
        Input-dependent channel-wise vector & 2023--2026 & $\alpha_t^h=f_{\alpha}(x_t)\in\mathbb{R}^{d_k}$ & GLA, RWKV-6/7, HGRN2, KDA, GDN2, EDA \\
        \bottomrule
    \end{tabularx}
\end{table}

\paragraph{No decay.}
Linear Transformer, DeltaNet, and DeltaProduct set $\alpha_t^h=1$, so historical state does not decay actively with time \citep{katharopoulos2020linear,schlag2021fastweight,yang2024deltanet,siems2025deltaproduct}. This preserves the accumulated history, but early information continues to occupy the finite state space, and the model cannot vary the retention strength of different historical components according to time or input content.

\paragraph{Fixed decay.}
Fixed decay introduces stable time scales through predefined retention coefficients. RetNet and the fixed-decay Lightning Attention-2 formulation use head-wise scalar decay, so different heads cover histories of different lengths; RWKV-5 uses channel-wise time decay, allowing different state channels within the same head to retain information at different rates \citep{sun2023retnet,minimax2025text01,peng2024eaglefinch}.\footnote{RWKV-4 also uses fixed channel-wise time decay, but it maintains normalized vector-valued accumulators rather than the matrix-valued associative state assumed in this subsection; it is therefore omitted from the comparison tables \citep{peng2023rwkv}.} Compared with undecayed accumulation, fixed decay continually reduces the weight of earlier history and forms multiscale memory, but its forgetting rate does not change with input content at inference time.

\paragraph{Input-dependent decay.}
Input-dependent decay makes the retention coefficient a function of the current input. GLA uses a channel-wise vector gate, whereas GDN uses an input-dependent head-wise scalar, allowing different tokens to regulate historical retention globally or by channel \citep{yang2023gla,yang2024gdn}. HGRN2 also uses an input-dependent channel-wise forget gate and assigns a retention lower bound that increases with layer depth. At layer $\ell$,
\begin{equation}
    g_t^{(\ell)}=\sigma\!\left(f_g(x_t^{(\ell)})\right),
    \qquad
    \alpha_t^{(\ell)}=\boldsymbol{\beta}_{\ell}+(\mathbf{1}-\boldsymbol{\beta}_{\ell})\odot g_t^{(\ell)}
    \in(0,1)^{d_k},
    \label{eq:hgrn2-retain}
\end{equation}
where $g_t^{(\ell)}$ is generated from the current input and the channel-wise lower bound $\boldsymbol{\beta}_{\ell}\in[0,1)^{d_k}$ increases elementwise and monotonically with depth. This constraint allows lower layers to update local information more rapidly while higher layers retain longer historical time scales \citep{qin2024hgrn2}. RWKV-6/7, KDA, GDN2, and EDA also generate input-dependent channel-wise decay $\alpha_t\in(0,1)^{d_k}$ to control retention in individual state rows \citep{peng2024eaglefinch,peng2025rwkv7,zhang2025kimilinear,hatamizadeh2026gdn2,li2026eda}. Retention thus develops from a fixed temporal prior into token-dependent and channel-aware control, with cross-layer constraints further organizing different time scales.

\subsubsection{Erase / Correct}
\label{subsubsec:linear-erase-correct}

Erase / Correct describes which associations are removed or revised in the old state before new information is written. Let $X_{t-1}^e\in\mathbb{R}^{d_k\times d_v}$ be the source state actually read by the erase operation, let $R_t^e,U_t^e\in\mathbb{R}^{d_k\times m_t^e}$ be the read directions and state-update directions, and let $m_t^e$ be the effective erase rank. The total erase term is
\begin{equation}
    E_t=U_t^e\left((R_t^e)^{\top}X_{t-1}^e\right)
    \in\mathbb{R}^{d_k\times d_v}.
    \label{eq:linear-erase-general}
\end{equation}
Erase mechanisms evolve from no explicit removal, to delta correction in which read and update directions are coupled, and then to designs that use different parameters for the two directions or combine multiple erase components. Table~\ref{tab:linear-erase} summarizes this progression.

\begin{table}[H]
    \centering
    \small
    \caption{Evolution of Erase / Correct mechanisms in Linear Attention.}
    \label{tab:linear-erase}
    \begin{tabularx}{\textwidth}{@{}>{\raggedright\arraybackslash}p{0.24\textwidth}>{\raggedright\arraybackslash}p{0.13\textwidth}>{\raggedright\arraybackslash}p{0.31\textwidth}>{\raggedright\arraybackslash}X@{}}
        \toprule
        \textbf{Erase category} & \textbf{Period} & \textbf{Unified form} & \textbf{Representative methods} \\
        \midrule
        No explicit erase & 2020--2024 & $m_t^e=0,\ E_t=0$ & Linear Transformer, RetNet, GLA, RWKV-5/6, HGRN2, Lightning Attention-2 \\
        Coupled corrective erase & 2021--2025 & $m_t^e=1,\ U_t^e=\beta_tk_t,\ R_t^e=k_t$ & DeltaNet, GDN, KDA, DeltaProduct \\
        Decoupled erase & 2025--2026 & $U_t^e$ and $R_t^e$ need not be aligned and/or $m_t^e>1$ & RWKV-7, GDN2, EDA \\
        \bottomrule
    \end{tabularx}
\end{table}

\paragraph{No explicit erase.}
When $m_t^e=0$, no independent erase term is constructed and $E_t=0$. Linear Transformer \citep{katharopoulos2020linear}, RetNet \citep{sun2023retnet}, GLA \citep{yang2023gla}, RWKV-5/6 \citep{peng2024eaglefinch}, HGRN2 \citep{qin2024hgrn2}, and Lightning Attention-2 \citep{minimax2025text01} do not read the old content at the current address and remove it directionally before writing. In the unified decomposition, HGRN2 contains only $\bar S_{t-1}=D_tS_{t-1}$ and no address-specific removal, so it also belongs to this category.

\paragraph{Coupled corrective erase.}
DeltaNet uses the same direction to read and modify the old association at the current address. For $k_t\in\mathbb{R}^{d_k}$ and scalar correction rate $\beta_t\in\mathbb{R}$,
\begin{equation}
    X_{t-1}^e=\bar S_{t-1},
    \qquad m_t^e=1,
    \qquad U_t^e=\beta_tk_t,
    \qquad R_t^e=k_t,
    \label{eq:linear-coupled-erase}
\end{equation}
which gives $E_t=\beta_tk_t(k_t^{\top}\bar S_{t-1})$. Because $R_t^e$ and $U_t^e$ are both determined by $k_t$, the old association is read and modified along aligned directions. GDN and KDA retain this rank-one correction, while DeltaProduct performs multiple corrections of the same type within one token \citep{schlag2021fastweight,yang2024deltanet,yang2024gdn,zhang2025kimilinear,siems2025deltaproduct}.

\paragraph{Decoupled erase.}
RWKV-7, GDN2, and EDA no longer require every erase component to use identical read and state-update directions. GDN2 defines a key-side gate $b_t\in\mathbb{R}^{d_k}$ and uses
\begin{equation}
    X_{t-1}^e=\bar S_{t-1},
    \qquad m_t^e=1,
    \qquad U_t^e=k_t,
    \qquad R_t^e=b_t\odot k_t.
    \label{eq:gdn2-erase}
\end{equation}
Here $b_t\odot k_t$ determines the content read from the old state, while $k_t$ determines the state-update direction, producing an asymmetric rank-one erase \citep{hatamizadeh2026gdn2}.

EDA performs an independent pre-erase followed by delta correction. Let $e_t,k_t\in\mathbb{R}^{d_k}$ be the pre-erase and correction directions and $\gamma_t,\beta_t\in\mathbb{R}$ their strengths. Expanding the two operations gives
\begin{equation}
    X_{t-1}^e=\bar S_{t-1},
    \quad
    U_t^e=\begin{bmatrix}\gamma_te_t&\beta_tk_t\end{bmatrix},
    \quad
    R_t^e=\begin{bmatrix}e_t&k_t-\gamma_t(e_t^{\top}k_t)e_t\end{bmatrix},
    \quad m_t^e=2.
    \label{eq:eda-erase}
\end{equation}
The first component performs pre-erasure along $e_t$, and the second performs correction along $k_t$ after pre-erasure; together they form a rank-two total erase \citep{li2026eda}.

RWKV-7 uses a normalized removal key $\widehat{k}_t\in\mathbb{R}^{d_k}$ and a vector-valued removal rate $b_t\in\mathbb{R}^{d_k}$,
\begin{equation}
    X_{t-1}^e=S_{t-1},
    \qquad m_t^e=1,
    \qquad R_t^e=\widehat{k}_t,
    \qquad U_t^e=b_t\odot\widehat{k}_t.
    \label{eq:rwkv7-erase}
\end{equation}
The read direction is defined by $R_t^e$, whereas $U_t^e$ adjusts the state-update direction through channel-wise removal strength \citep{peng2025rwkv7,zhang2025kimilinear}.

\subsubsection{Write / Commit}
\label{subsubsec:linear-write-commit}

Write / Commit specifies the location, content, and strength with which current information is committed. Let $A_t^w\in\mathbb{R}^{d_k\times r_t^w}$ and $C_t^w\in\mathbb{R}^{d_v\times r_t^w}$ denote write directions and contents, where $r_t^w$ is the effective write rank. The write term is
\begin{equation}
    W_t=A_t^w(C_t^w)^{\top}\in\mathbb{R}^{d_k\times d_v}.
    \label{eq:linear-write-general}
\end{equation}
Write evolves from direct writing without additional regulation, to corrective writing controlled by an update rate, and then to decoupled writing with independent control of write direction or content. Table~\ref{tab:linear-write} summarizes the progression.

\begin{table}[H]
    \centering
    \small
    \caption{Evolution of Write / Commit mechanisms in Linear Attention.}
    \label{tab:linear-write}
    \begin{tabularx}{\textwidth}{@{}>{\raggedright\arraybackslash}p{0.24\textwidth}>{\raggedright\arraybackslash}p{0.13\textwidth}>{\raggedright\arraybackslash}p{0.31\textwidth}>{\raggedright\arraybackslash}X@{}}
        \toprule
        \textbf{Write category} & \textbf{Period} & \textbf{Unified form} & \textbf{Representative methods} \\
        \midrule
        Direct write & 2020--2024 & $A_t^w=k_t,\ C_t^w=v_t$ & Linear Transformer, RetNet, GLA, RWKV-5/6, Lightning Attention-2 \\
        Corrective write & 2021--2025 & $A_t^w=\beta_tk_t,\ C_t^w=v_t$ & DeltaNet, GDN, KDA, DeltaProduct \\
        Erase--write decoupling & 2025--2026 & Separate control of erase and write address or content & RWKV-7, GDN2, EDA \\
        \bottomrule
    \end{tabularx}
\end{table}

\paragraph{Direct write.}
The earliest Linear Attention methods directly add a rank-one key--value association,
\begin{equation}
    r_t^w=1,
    \qquad A_t^w=k_t,
    \qquad C_t^w=v_t,
    \qquad W_t=k_tv_t^{\top}.
    \label{eq:linear-direct-write}
\end{equation}
Linear Transformer establishes this form \citep{katharopoulos2020linear}. RetNet, GLA, RWKV-5/6, and Lightning Attention-2 use the same direct outer-product write or an equivalent recurrence \citep{sun2023retnet,yang2023gla,peng2024eaglefinch,minimax2025text01}. HGRN2 sets $A_t^w=\mathbf{1}-\alpha_t$ and $C_t^w=i_t$, where $i_t\in\mathbb{R}^{d_v}$ is candidate content, giving $W_t=(\mathbf{1}-\alpha_t)i_t^{\top}$; equivalently, one may relabel its write address as $k_t=\mathbf{1}-\alpha_t$ \citep{qin2024hgrn2}.

\paragraph{Corrective write.}
Delta-style methods use a scalar update rate to control commit strength,
\begin{equation}
    r_t^w=1,
    \qquad A_t^w=\beta_tk_t,
    \qquad C_t^w=v_t,
    \qquad W_t=\beta_tk_tv_t^{\top}.
    \label{eq:linear-corrective-write}
\end{equation}
The coefficient $\beta_t$ determines how strongly the current association enters the state. DeltaNet establishes this rate-controlled commit; GDN and KDA retain scalar-controlled writing; and DeltaProduct generates multiple write components within one token \citep{schlag2021fastweight,yang2024deltanet,yang2024gdn,zhang2025kimilinear,siems2025deltaproduct}.

\paragraph{Erase--write decoupling.}
Recent methods decouple the erase and write roles through separate address or content controls. GDN2 uses
\begin{equation}
    r_t^w=1,
    \qquad A_t^w=k_t,
    \qquad C_t^w=w_t\odot v_t,
    \label{eq:gdn2-write}
\end{equation}
where $w_t\in\mathbb{R}^{d_v}$ is a value-side gate controlling the committed content channel by channel \citep{hatamizadeh2026gdn2}. EDA retains the corrective write $A_t^w=\beta_tk_t$, $C_t^w=v_t$, and $W_t=\beta_tk_tv_t^{\top}$, while decoupling it from the independently addressed pre-erase \citep{li2026eda}. RWKV-7 uses a distinct add key with $A_t^w=k_t$, $C_t^w=v_t$, and $W_t=k_tv_t^{\top}$ \citep{peng2025rwkv7,zhang2025kimilinear}.

\subsubsection{Representative Recurrent Update Rules}
\label{subsubsec:linear-update-rules}

Retain, Erase, and Write jointly determine the full recurrence. Table~\ref{tab:linear-update-rules} lists representative methods chronologically and marks their Decay, Erase, and Write types. All equations use the single-head orientation $S_t\in\mathbb{R}^{d_k\times d_v}$ and retain only the core update terms. Let $I\in\mathbb{R}^{d_k\times d_k}$ and $D_t=\operatorname{Diag}(\alpha_t)$; when decay is scalar, $D_t=\alpha_t I$.

\begin{table}[t]
    \centering
    \scriptsize
    \setlength{\tabcolsep}{3pt}
    \caption{Representative recurrent update rules.}
    \label{tab:linear-update-rules}
    \begin{tabularx}{\textwidth}{@{}>{\raggedright\arraybackslash}p{0.105\textwidth}>{\raggedright\arraybackslash}p{0.135\textwidth}>{\raggedright\arraybackslash}p{0.14\textwidth}>{\raggedright\arraybackslash}p{0.12\textwidth}>{\centering\arraybackslash}p{0.04\textwidth}>{\raggedright\arraybackslash}X@{}}
        \toprule
        \textbf{Method} & \textbf{Decay} & \textbf{Erase} & \textbf{Write} & \textbf{Year} & \textbf{Unified update rule} \\
        \midrule
        Linear Transformer \citep{katharopoulos2020linear} & No decay & No explicit erase & Direct & 2020 & $S_t=S_{t-1}+k_tv_t^{\top}$ \\
        DeltaNet \citep{schlag2021fastweight,yang2024deltanet} & No decay & Coupled corrective & Corrective & 2021 & $S_t=(I-\beta_tk_tk_t^{\top})S_{t-1}+\beta_tk_tv_t^{\top}$ \\
        RetNet \citep{sun2023retnet} & Fixed head-wise & No explicit erase & Direct & 2023 & $S_t=\alpha^hS_{t-1}+k_tv_t^{\top}$ \\
        GLA \citep{yang2023gla} & Input-dependent channel-wise & No explicit erase & Direct & 2023 & $S_t=D_tS_{t-1}+k_tv_t^{\top}$, $D_t=\operatorname{Diag}(f_{\alpha}(x_t))$ \\
        RWKV-5 \citep{peng2024eaglefinch} & Fixed channel-wise & No explicit erase & Direct & 2024 & $S_t=\operatorname{Diag}(\alpha^h)S_{t-1}+k_tv_t^{\top}$ \\
        RWKV-6 \citep{peng2024eaglefinch} & Input-dependent channel-wise & No explicit erase & Direct & 2024 & $S_t=D_tS_{t-1}+k_tv_t^{\top}$, $D_t=\operatorname{Diag}(f_{\alpha}(x_t))$ \\
        HGRN2 \citep{qin2024hgrn2} & Input-dependent channel-wise & No explicit erase & Direct & 2024 & $S_t=D_tS_{t-1}+(\mathbf{1}-\alpha_t)i_t^{\top}$ \\
        GDN \citep{yang2024gdn} & Input-dependent head-wise & Coupled corrective & Corrective & 2024 & $S_t=\alpha_t(I-\beta_tk_tk_t^{\top})S_{t-1}+\beta_tk_tv_t^{\top}$ \\
        Lightning Attention-2 \citep{minimax2025text01} & Fixed head-wise & No explicit erase & Direct & 2024 & $S_t=\alpha^hS_{t-1}+k_tv_t^{\top}$ \\
        DeltaProduct \citep{siems2025deltaproduct} & No decay & Coupled corrective & Corrective & 2025 & $S_t=(\mathcal{T}_{t,R}\circ\cdots\circ\mathcal{T}_{t,1})(S_{t-1})$ \\
        RWKV-7 \citep{peng2025rwkv7} & Input-dependent channel-wise & Decoupled & Erase--write decoupled & 2025 & $S_t=[D_t-(b_t\odot\widehat{k}_t)\widehat{k}_t^{\top}]S_{t-1}+k_tv_t^{\top}$ \\
        KDA \citep{zhang2025kimilinear} & Input-dependent channel-wise & Coupled corrective & Corrective & 2025 & $S_t=(I-\beta_tk_tk_t^{\top})D_tS_{t-1}+\beta_tk_tv_t^{\top}$ \\
        GDN2 \citep{hatamizadeh2026gdn2} & Input-dependent channel-wise & Decoupled & Erase--write decoupled & 2026 & $S_t=[I-k_t(b_t\odot k_t)^{\top}]D_tS_{t-1}+k_t(w_t\odot v_t)^{\top}$ \\
        EDA \citep{li2026eda} & Input-dependent channel-wise & Decoupled & Erase--write decoupled & 2026 & $S_t=(I-\beta_tk_tk_t^{\top})(I-\gamma_te_te_t^{\top})D_tS_{t-1}+\beta_tk_tv_t^{\top}$ \\
        \bottomrule
    \end{tabularx}
\end{table}

The formulas use unified notation to emphasize Update structure and omit method-specific normalization, feature generation, and output gating. In HGRN2, $\alpha_t\in(0,1)^{d_k}$ is the channel-wise forget gate and $i_t\in\mathbb{R}^{d_v}$ is candidate content; $\mathbf{1}-\alpha_t$ also forms its direct-write direction. In DeltaProduct, $R$ is the number of delta transformations performed for one token and $\mathcal{T}_{t,r}$ is the $r$th update. The method-specific symbols and decoupling relations for RWKV-7, GDN2, and EDA are defined above.

\subsubsection{Loss-Based Optimization View}
\label{subsubsec:linear-loss-optimization}

The same recurrent rules admit a complementary comparison through equivalent learning objectives that reproduce their state updates. Prior work has interpreted recurrent associative-memory updates as online optimization over an internal state, including fast-weight delta rules, test-time training and regression, and optimization-based analyses of gated Linear Attention \citep{schlag2021fastweight,sun2024ttt,wang2026ttr,zhang2025kimilinear}. We therefore compare representative methods by identifying a per-token objective whose unit-step gradient update yields the corresponding recurrence,
\begin{equation}
    S_t=S_t^{-}-\nabla_{S_t^{-}}\mathcal{L}_t(S_t^{-}),
    \label{eq:linear-loss-update-template}
\end{equation}
where the optimization state $S_t^{-}$ can be the previous state, a decayed state, or an intermediate state produced by an earlier optimization step.

\begin{table}[H]
    \centering
    \scriptsize
    \setlength{\tabcolsep}{3pt}
    \renewcommand{\arraystretch}{1.18}
    \caption{Survey-derived equivalent per-token objectives corresponding to representative update rules.}
    \label{tab:linear-loss-optimization}
    \begin{tabularx}{\textwidth}{@{}>{\raggedright\arraybackslash}p{0.125\textwidth}>{\raggedright\arraybackslash}p{0.43\textwidth}>{\raggedright\arraybackslash}X@{}}
        \toprule
        \textbf{Method} & \textbf{Per-token objective $\mathcal{L}_t$} & \textbf{Resulting update} \\
        \midrule
        Linear Transformer \citep{katharopoulos2020linear}
        & $-\left\langle S_{t-1}^{\top}k_t,v_t\right\rangle$
        & $S_t=S_{t-1}+k_tv_t^{\top}$ \\

        RetNet \citep{sun2023retnet}
        & $-\left\langle S_{t-1}^{\top}k_t,v_t\right\rangle+\frac{1}{2}\left\|\sqrt{1-\alpha^h}\,S_{t-1}\right\|_F^2$
        & $S_t=\alpha^hS_{t-1}+k_tv_t^{\top}$ \\

        GLA \citep{yang2023gla}
        & $-\left\langle S_{t-1}^{\top}k_t,v_t\right\rangle+\frac{1}{2}\left\|\sqrt{I-D_t}\,S_{t-1}\right\|_F^2$
        & $S_t=D_tS_{t-1}+k_tv_t^{\top}$ \\

        HGRN2 \citep{qin2024hgrn2}
        & $-\left\langle S_{t-1}^{\top}(\mathbf{1}-\alpha_t),i_t\right\rangle+\frac{1}{2}\left\|\sqrt{I-D_t}\,S_{t-1}\right\|_F^2$
        & $S_t=D_tS_{t-1}+(\mathbf{1}-\alpha_t)i_t^{\top}$ \\
        \midrule
        DeltaNet \citep{schlag2021fastweight,yang2024deltanet}
        & $\frac{\beta_t}{2}\left\|S_{t-1}^{\top}k_t-v_t\right\|_2^2$
        & $S_t=(I-\beta_tk_tk_t^{\top})S_{t-1}+\beta_tk_tv_t^{\top}$ \\

        GDN \citep{yang2024gdn}
        & $\frac{\beta_t}{2}\left\|\widetilde S_{t-1}^{\top}k_t-v_t\right\|_2^2$, optimized at $\widetilde S_{t-1}=\alpha_tS_{t-1}$
        & $S_t=\alpha_t(I-\beta_tk_tk_t^{\top})S_{t-1}+\beta_tk_tv_t^{\top}$ \\

        KDA \citep{zhang2025kimilinear}
        & $\frac{\beta_t}{2}\left\|\widetilde S_{t-1}^{\top}k_t-v_t\right\|_2^2$, optimized at $\widetilde S_{t-1}=D_tS_{t-1}$
        & $S_t=(I-\beta_tk_tk_t^{\top})D_tS_{t-1}+\beta_tk_tv_t^{\top}$ \\

        DeltaProduct \citep{siems2025deltaproduct}
        & $\mathcal{L}_{t,r}=\frac{\beta_{t,r}}{2}\left\|S_{t,r-1}^{\top}k_{t,r}-v_{t,r}\right\|_2^2$, $r=1,\ldots,R$
        & $S_t=(\mathcal{T}_{t,R}\circ\cdots\circ\mathcal{T}_{t,1})(S_{t-1})$ \\

        EDA \citep{li2026eda}
        & $\mathcal{L}_t^e=\frac{\gamma_t}{2}\left\|\widetilde S_{t-1}^{\top}e_t\right\|_2^2$, then $\mathcal{L}_t^w=\frac{\beta_t}{2}\left\|(S_t^e)^{\top}k_t-v_t\right\|_2^2$, with $\widetilde S_{t-1}=D_tS_{t-1}$
        & $S_t=(I-\beta_tk_tk_t^{\top})(I-\gamma_te_te_t^{\top})D_tS_{t-1}+\beta_tk_tv_t^{\top}$ \\
        \bottomrule
    \end{tabularx}
\end{table}

For DeltaProduct, each optimization step is
$\mathcal{T}_{t,r}(S)=(I-\beta_{t,r}k_{t,r}k_{t,r}^{\top})S+\beta_{t,r}k_{t,r}v_{t,r}^{\top}$, with $S_{t,0}=S_{t-1}$. For EDA, the first objective produces $S_t^e=(I-\gamma_te_te_t^{\top})\widetilde S_{t-1}$, which is then used as the optimization state of the second objective.

Table~\ref{tab:linear-loss-optimization} reveals two broad stages in the development of single-state updates. Early additive methods maximize the correlation between the state prediction and the current value. Their differences arise from the regularization applied to the old state: Linear Transformer has no forgetting penalty, RetNet introduces isotropic fixed decay, GLA replaces it with an input-dependent channel-wise penalty, and HGRN2 further couples the write address to the complementary forget gate. These objectives produce direct writes whose magnitude does not depend on the value already predicted at the current address.

Delta-style methods replace correlation maximization with reconstruction-error minimization, making the write explicitly residual dependent. DeltaNet performs one correction on the previous state. GDN and KDA move the same correction to a decayed optimization state, with GDN using a head-wise scalar and KDA using channel-wise decay. DeltaProduct increases optimization depth by fitting multiple key--value targets sequentially within one token. EDA instead expands the objective sequence: it first fits a zero target at an independently selected erase address and then fits the current value at the write address. The progression therefore changes not only the form of the loss, but also the state at which it is optimized, the number of optimization steps, and whether erase and write share the same address and target.

\subsection{Capacity Expansion}
\label{subsec:linear-capacity-expansion}

By 2025, decay and delta-style editing enabled a single state to use bounded memory more effectively, but the complete history was still compressed into one $d_k\times d_v$ matrix. Better update rules reduce conflicts but cannot provide an unlimited number of independent storage locations; directly enlarging a dense state also increases every token's read and write cost. Capacity Expansion therefore relaxes the assumption that each head maintains only one dense state and enlarges available memory through sparse selection.

Standard Linear Attention maintains one matrix state $S_t\in\mathbb{R}^{d_k\times d_v}$ per head. Capacity Expansion generalizes it to $M$ states,
\begin{equation}
    \mathcal{S}_t=
    \left\{S_t^{(m)}\in\mathbb{R}^{d_k\times d_v}\right\}_{m=1}^{M},
    \qquad
    \mathcal{S}_t\in\mathbb{R}^{M\times d_k\times d_v}.
    \label{eq:linear-capacity-expanded-state}
\end{equation}
Flattening the state and key dimensions gives an equivalent row matrix of shape $(Md_k)\times d_v$. SSE and SDM both increase capacity by adding available state rows, but SSE retains a group structure whereas SDM uses a larger flat memory table. Under this unified row-memory view, routing determines which groups or rows are eligible for the current query and therefore belongs to Access, while weighting and aggregating the selected rows into one contextual representation belong to Readout. If a grouped formulation instead produces separate group- or state-specific readouts before combining them, that subsequent combination belongs to Integration. The distinction therefore depends on the analytical granularity: groups may be treated as internal partitions of one expanded state or as distinct readout paths; for SSE and SDM in this subsection, we use the former interpretation unless separate completed readouts are made explicit.

% \begin{figure}[H]
%     \centering
%     \includegraphics[width=\textwidth]{figures/linear-capacity-expansion.png}
%     \caption{A unified row-memory view of SSE and SDM. SSE preserves a grouped organization, whereas SDM performs sparse routing directly in a larger flat row space.}
%     \label{fig:linear-capacity-expansion}
% \end{figure}

\subsubsection{Sparse State Expansion (SSE)}
\label{subsubsec:linear-sse}

Sparse State Expansion (SSE) enlarges the associative memory by maintaining multiple groups of state rows \citep{pan2025sse}. Each group provides a separate region of the expanded state space, while a learned router activates only a small number of groups for the current token. Within the active groups, separate coefficients determine how strongly individual rows participate in writing and reading. The resulting organization is hierarchical: routing first identifies relevant groups and then distributes the update or readout within those groups.

This hierarchy gives SSE a strong structural prior. Group-level selection reduces the number of state regions considered by each token and provides a comparatively regular execution pattern. Because read and write operate within the same group organization, the structure can also help preserve a relationship between where information is stored and where later queries search for it. At the same time, predefined group boundaries restrict direct competition across the full memory space. Rows in an inactive group cannot participate in the current operation, even when their contents might be relevant, and the fixed partition may limit how memory units reorganize as their roles evolve during training.

SSE therefore expands capacity through a balance between specialization and regularity. Increasing the number of groups enlarges the persistent memory without requiring every token to interact with every row, but the effective benefit depends on whether the router distributes information across groups, avoids repeatedly overloading a small subset, and selects groups that remain useful for later retrieval.

\subsubsection{Sparse Delta Memory (SDM)}
\label{subsubsec:linear-sdm}

Sparse Delta Memory (SDM) removes the predefined group boundaries and organizes the expanded state as a flat collection of recurrent memory rows \citep{cabannes2026sdm}. Product-Key Memory routers retrieve a limited number of rows for writing and reading. Rows selected for writing receive delta-style updates, while unselected rows preserve their previous contents. A separate read router identifies the rows that contribute to the current output.

The flat organization gives SDM greater addressing freedom than a fixed grouped structure. Any row can, in principle, compete for the current write or read operation, and the write and read supports can be generated independently. This separation allows the system to distinguish where incoming information should be stored from where a later query should search. It also makes the enlarged memory less dependent on a manually imposed partition of the row space.

The same flexibility introduces a more demanding coordination problem. A row may be written frequently but rarely retrieved, while another may be retrieved before receiving sufficiently relevant updates. Independently learned routers can also develop incompatible address conventions, weakening the connection between storage and later recall. Moreover, the practical efficiency of a large flat state depends on fast retrieval, balanced row utilization, and memory access patterns that do not offset the savings obtained from sparse activation.

\subsubsection{Comparison and Open Design Space}
\label{subsubsec:linear-capacity-comparison}

Table~\ref{tab:sse-sdm-comparison} compares SSE and SDM. Both expand a dense matrix into row memory and use sparse read/write operations to control per-token cost; SSE uses fixed groups for hierarchical routing, whereas SDM performs independent retrieval directly in a flat row space.

\begin{table}[H]
    \centering
    \small
    \caption{Comparison of SSE and SDM under a unified row-memory view.}
    \label{tab:sse-sdm-comparison}
    \begin{tabularx}{\textwidth}{@{}>{\raggedright\arraybackslash}p{0.20\textwidth}>{\raggedright\arraybackslash}p{0.38\textwidth}>{\raggedright\arraybackslash}X@{}}
        \toprule
        \textbf{Aspect} & \textbf{SSE} & \textbf{SDM} \\
        \midrule
        Unified state & \multicolumn{2}{l}{$S_t\in\mathbb{R}^{N\times d_v}$, $N=Md_k$; each row is a $d_v$-dimensional memory unit} \\
        State organization & $M$ fixed groups with $d_k$ rows per group & Flat row table without group boundaries \\
        Write selection & Group Top-$K_g$ plus within-group coefficients & Independent PKM write router selecting Top-$W$ rows \\
        Read selection & Read over selected groups and row support & Independent PKM read router selecting Top-$R$ rows \\
        Read/write relation & Shared group structure; strongly related supports & Independent read and write routers \\
        Capacity expansion & Increase the number of groups & Directly increase the number of flat rows \\
        Routing characteristic & Hierarchical and regular, with a group prior & Flat and flexible, but more dependent on fast approximate retrieval \\
        \bottomrule
    \end{tabularx}
\end{table}

Fixed groups reduce the search range and regularize routing, but a predefined partition can restrict dynamic reorganization among memory units. Flat routing provides greater addressing freedom, but must sustain low-cost, high-recall retrieval over a much larger row space. A hierarchical combination of grouped and flat routing could provide both coarse partitioning and fine-grained global selection.

Coordination between Read and Write also remains unresolved. Shared or related supports keep written content aligned with later retrieval but may prevent a query from recalling other memory regions. Fully independent routers are more flexible, yet can produce uneven write utilization, rows that are rarely read, or memory units retrieved before they are adequately trained. Explicit read--write consistency objectives, load balancing, and state-utilization regularization may improve this trade-off.

As row count increases, routing cost, cross-layer memory organization, and state lifetime become increasingly important. Potential directions include reusing routing decisions across neighboring layers, allocating or reclaiming rows according to usage frequency, and allowing different layers to access shared collections of state memory. Capacity Expansion therefore requires the joint design of Memory Representation, sparse Access, Readout, and update dynamics rather than merely increasing state size.

\subsection{Temporal Expansion}
\label{subsec:linear-temporal-expansion}

Capacity Expansion increases rows within the associative space, but history may still be compressed into one temporally uniform memory. When recent details and distant information remain mixed in the same state, different temporal ranges share the same representation resolution and cannot be independently emphasized during readout. Temporal Expansion instead preserves multiple states along the token dimension. If $S_t^{(m)}$ summarizes a contiguous historical interval, then
\begin{equation}
    \mathcal{S}_t
    =
    \left\{
        S_t^{(m)}
    \right\}_{m=1}^{M_t}.
    \label{eq:linear-temporal-expanded-state}
\end{equation}
The query first produces a state-specific Readout from each retained state, after which Integration combines these readouts,
\begin{equation}
    o_t
    =
    \sum_{m=1}^{M_t}
    \lambda_{t,m}
    \operatorname{Read}
    \!\left(
        q_t,S_t^{(m)}
    \right),
    \label{eq:linear-temporal-expanded-readout}
\end{equation}
where $M_t$ is the number of retained segment states and $\lambda_{t,m}$ controls the contribution of state $m$ during Integration.

The central questions are how tokens are partitioned into segments, whether old states are retained or merged as the context grows, and how the current query combines information from different temporal intervals. Log-Linear Attention \citep{guo2025loglinear} uses a deterministic multiscale hierarchy, Dynamic Linear Attention \citep{wang2026dla} uses content-dependent segmentation and merging, whereas Multi-Head Linear Attention \citep{zhang2026mhla} preserves a flat collection of fixed-granularity token-block summaries. These mechanisms all expand memory along the token dimension, but differ in temporal organization, state-count control, and cross-state Integration.

\begin{figure}[H]
    \centering
    \includegraphics[width=\textwidth]{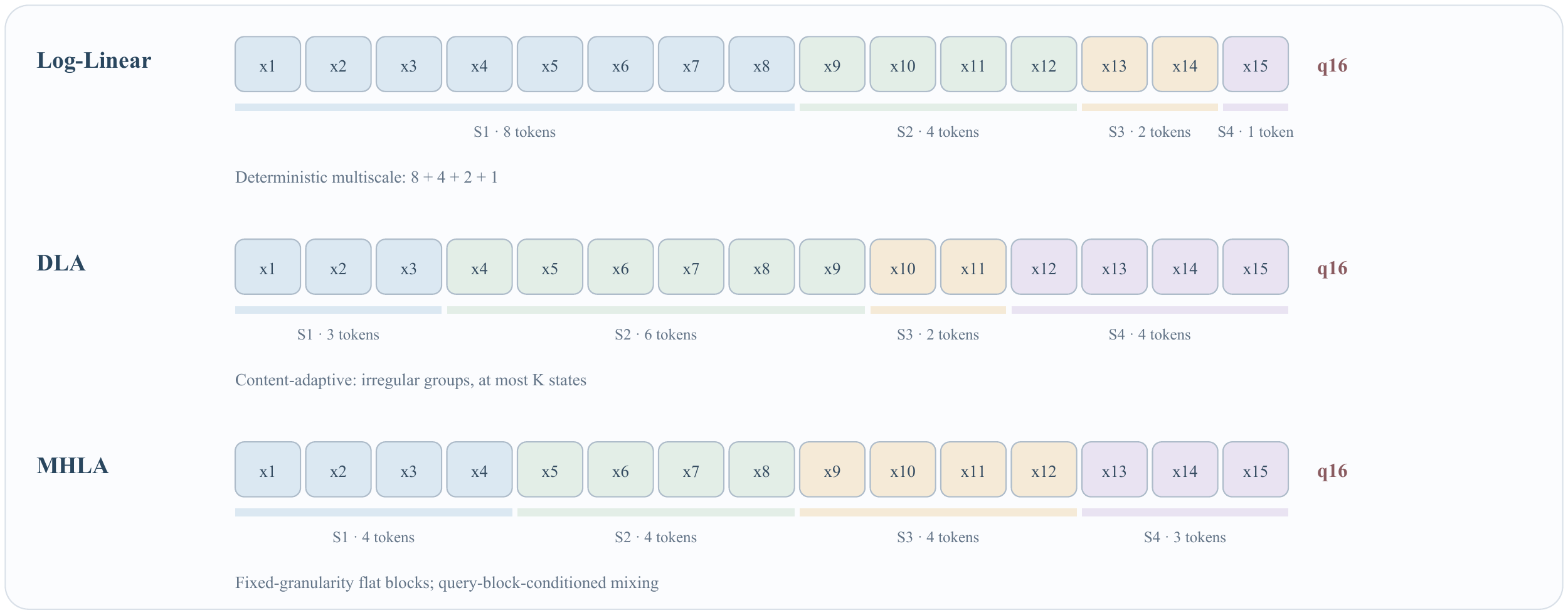}
    \caption{Comparison of temporal grouping strategies in Log-Linear Attention, DLA, and MHLA. Log-Linear Attention deterministically organizes history into multiscale intervals, DLA forms irregular content-adaptive segments under a bounded state budget, and MHLA preserves a flat collection of fixed-granularity token blocks with query-block-conditioned mixing.}
    \label{fig:linear-temporal-expansion}
\end{figure}

\subsubsection{Log-Linear Attention}
\label{subsubsec:log-linear-attention}

Log-Linear Attention uses a deterministic multiscale organization to avoid compressing the entire history into one state \citep{guo2025loglinear}. A Fenwick-tree or power-of-two schedule maintains $O(\!\log t)$ states. As new tokens arrive, shorter segments are merged according to a binary-carry rule, producing contiguous historical groups such as $8+4+2+1$. Recent information is preserved in short segments at high resolution, while distant history is compressed into longer segments, and the number of states grows only logarithmically with sequence length. Its limitation is that segmentation depends entirely on position, so a fixed merge can cross an important semantic boundary.

\subsubsection{Dynamic Linear Attention (DLA)}
\label{subsubsec:dynamic-linear-attention}

DLA replaces fixed temporal partitioning with content-dependent segmentation \citep{wang2026dla}. It uses a State Information Score to determine whether the current token should continue to accumulate into an existing state or create a new segment state when the information changes substantially. When the number of states reaches an upper bound $K$, DLA merges adjacent states with the lowest information density. Relative to Log-Linear Attention, it allocates more of the limited state budget to regions with dense semantic change while keeping memory bounded by $M_t\leq K$. The cost is that boundaries and merges depend on dynamic scores, and history that has already been merged generally cannot recover its original temporal resolution.

\subsubsection{Multi-Head Linear Attention}
\label{subsubsec:multi-head-linear-attention}

Multi-Head Linear Attention (MHLA) partitions the token sequence into fixed blocks and preserves a separate key--value summary for each block \citep{zhang2026mhla}. Its token-level ``heads'' therefore differ from conventional feature heads, which partition the channel dimension. Rather than merging all block contributions into one global state, MHLA allows each query block to form a learned mixture of the retained summaries, followed by the usual query--key interaction within each block. From the temporal-memory perspective, MHLA is a flat expansion with approximately uniform chunk resolution: it preserves greater distinction among historical intervals than a single recurrent state, but unlike Log-Linear Attention and DLA, it does not intrinsically merge older summaries or bound their number. Its efficiency therefore depends on controlling both the number of retained blocks and the cost of inter-block mixing.

\subsubsection{Comparison and Development Trends}
\label{subsubsec:linear-temporal-comparison}

Log-Linear Attention, DLA, and MHLA all preserve multiple associative summaries over different token intervals, but organize temporal resolution differently. Log-Linear Attention uses a deterministic multiscale hierarchy, DLA creates and merges segments according to content, and MHLA retains a flat collection of fixed-granularity blocks. Table~\ref{tab:linear-temporal-comparison} compares their segmentation criteria, state-management rules, and readout mechanisms.

\begin{table}[H]
    \centering
    \scriptsize
    \setlength{\tabcolsep}{2.5pt}
    \caption{Comparison of representative Temporal Expansion mechanisms.}
    \label{tab:linear-temporal-comparison}
    \begin{tabularx}{\textwidth}{
        @{}
        >{\raggedright\arraybackslash}p{0.15\textwidth}
        >{\raggedright\arraybackslash}p{0.26\textwidth}
        >{\raggedright\arraybackslash}p{0.26\textwidth}
        >{\raggedright\arraybackslash}X
        @{}
    }
        \toprule
        \textbf{Aspect}
        &
        \textbf{Log-Linear Attention}
        &
        \textbf{DLA}
        &
        \textbf{MHLA}
        \\
        \midrule

        Segmentation basis
        &
        Position-dependent Fenwick or power-of-two schedule
        &
        Content-dependent State Information Score
        &
        Fixed token-block partition
        \\

        Segment form
        &
        Multiscale contiguous intervals
        &
        Unequal-length semantic intervals
        &
        Flat, usually equal-granularity blocks
        \\

        Number of states
        &
        $M_t=O(\!\log t)$
        &
        $M_t\leq K$
        &
        Determined by the number of retained blocks
        \\

        State management
        &
        Hierarchical binary-carry merging
        &
        Dynamic creation and information-based merging
        &
        Independent block summaries without intrinsic merging
        \\

        Temporal resolution
        &
        Fine for recent and coarse for distant history
        &
        Allocated according to semantic change
        &
        Approximately uniform across blocks
        \\

        Cross-state Integration
        &
        Query-dependent integration of temporal-scale readouts
        &
        Integration of readouts from dynamically constructed segment states
        &
        Query-block-dependent integration of block-summary readouts
        \\

        Principal advantage
        &
        Predictable complexity and multiscale coverage
        &
        Higher resolution around semantic transitions
        &
        Preserves independently distinguishable block summaries
        \\

        Principal limitation
        &
        Fixed merges may cross semantic boundaries
        &
        Depends on learned boundary and merging decisions
        &
        State count and mixing cost can grow with the number of blocks
        \\

        \bottomrule
    \end{tabularx}
\end{table}

The three methods illustrate complementary approaches to temporal memory organization. Log-Linear Attention varies representation resolution according to temporal distance, DLA allocates resolution according to content change, and MHLA preserves a uniform flat partition while differentiating historical blocks during readout. These choices expose a broader trade-off among temporal resolution, bounded memory growth, segmentation adaptivity, and query-dependent retrieval. They are also potentially composable: fixed block summaries could be hierarchically merged or dynamically consolidated, while richer query-conditioned routing could be applied over the resulting bounded collection of temporal states.

\subsection{Auxiliary Coordination and Routing Enhancements}
\label{subsec:linear-auxiliary-enhancements}

The preceding subsections concern the construction and organization of associative memory itself. Memory Update methods change how information is retained, corrected, erased, or written within a recurrent state. Capacity Expansion increases the number of available state units, while Temporal Expansion preserves multiple states that summarize different historical intervals. A separate set of recent methods instead focuses on how Linear Attention computations are coordinated across broader structural axes such as feature heads and network layers. These methods do not form a single mechanism family: they are grouped here because their principal contribution lies in coordinating an identifiable Linear Attention computation rather than introducing another general state-update or state-organization rule.

\paragraph{Feature-head coordination.}
Softmax Linear Attention (SLA) introduces token-dependent competition across the feature heads of a Linear Attention layer \citep{xu2026sla}. A key-side gate controls the relative strength with which the current token is written into different head states, while a query-side gate controls the relative contribution of those heads to the current output. Unlike the mechanisms in Section~\ref{subsec:linear-memory-update}, SLA does not primarily redesign the recurrence within each state; instead, it allocates write strength across a bank of existing states and controls how their completed outputs are integrated. Its direct effects therefore lie in Memory Update and Integration. Because the gates are continuous and need not exclude a head from computation, they should not automatically be interpreted as sparse Access decisions.

\paragraph{Cross-depth coordination.}
Cross-Layer Value Routing (CLVR) operates along network depth \citep{cerruti2026lineararchitectures}. Rather than changing the number of recurrent states or the retain--correct--write rule within a layer, it exposes an internal value associated with the current write operation to subsequent layers through the residual stream. Later layers can therefore use a memory-derived signal that would otherwise remain internal to the layer that produced it.

CLVR differs from Capacity and Temporal Expansion because it does not primarily create additional persistent memory units. It also differs from the Hybrid Architectures examined in Section~\ref{sec:hybrid-architectures}, because it preserves the host Linear Attention recurrence rather than combining heterogeneous sequence mixers or independent memory paths. Within the five-dimensional framework, CLVR is best understood as cross-layer coordination around the memory operator. It affects how memory-derived information enters later computation, but the routed value is not necessarily a completed readout in the strict sense used to define Integration in Section~\ref{sec:unified-memory-centric-view}. CLVR should therefore be distinguished from both ordinary residual propagation and explicit fusion of completed memory-path outputs.

Taken together, SLA and CLVR expose two distinct coordination axes. SLA redistributes writing and output contribution across feature heads, whereas CLVR propagates a memory-derived signal across network depth. Their relationship to the preceding categories is complementary rather than hierarchical. State-update methods determine how an individual associative state changes; Capacity and Temporal Expansion determine which memory units are maintained; and the mechanisms discussed here determine how state computations or internal memory signals are coordinated across feature partitions and network layers. This distinction also explains why these methods may affect several analytical dimensions without constituting new top-level memory substrates.

\subsection{Summary}
\label{subsec:linear-summary}

Linear Attention replaces an enumerable token-wise KV history with recurrently maintained associative memory. Its development begins with changes to Memory Representation and Memory Update, and subsequently extends to the organization, access, readout, and coordination of multiple memory units.

\noindent\textbf{The evolution of state updating seeks to compress historical information more effectively within a finite associative state.}
The canonical additive update continually superposes new key--value associations, which can lead to interference as the state accumulates information. Retention mechanisms regulate how strongly previous content persists, delta-style correction revises the association stored at a particular address, and more differentiated erase--write mechanisms provide finer control over what is removed and committed. These developments primarily refine Memory Update while determining which historical distinctions remain preserved in Memory Representation.

\noindent\textbf{As sequence length increases, state expansion relaxes the reliance on a single uniformly compressed memory.}
Capacity Expansion enlarges the associative space through additional dimensions, groups, or memory rows, whereas Temporal Expansion maintains multiple states that summarize different historical intervals. These changes directly enrich Memory Representation, but also introduce corresponding Access, Readout, and Integration questions: the model must determine which maintained units are eligible for the current query, how each eligible state is decoded, and how the resulting state-specific readouts are combined. The effective benefit of expansion therefore depends not only on nominal state size, but also on routing accuracy, read--write coordination, temporal resolution, and the recoverability of stored information.

\noindent\textbf{Other emerging directions coordinate Linear Attention computations across broader structural axes.}
SLA allocates writing across feature heads and integrates their outputs, whereas CLVR propagates memory-derived signals across network depth. Both complement state updating and expansion without defining new memory substrates. Together with the capacity- and temporal-expansion mechanisms discussed above, they show that Linear Attention is developing from a compact associative operator toward a more organized memory system whose practical value depends on jointly balancing compression, capacity, temporal resolution, access cost, readout quality, and implementation efficiency.
\section{State Space Models}
\label{sec:state-space-models}

State Space Models (SSMs) process a sequence by evolving a hidden state as each input arrives and reading that state to produce a contextual representation. Their central design problem is to preserve useful features of the input history through dynamics that can be learned and computed efficiently. Structured transitions and input-dependent control address the state-dynamics problem \citep{gu2021s4,gu2023mamba,dao2024mamba2}, while later variants enrich the read--write interfaces to the retained state \citep{gu2026mamba3,wang2026mimomamba}.

\begin{figure}[!hbp]
    \centering
    \includegraphics[width=16.5cm]{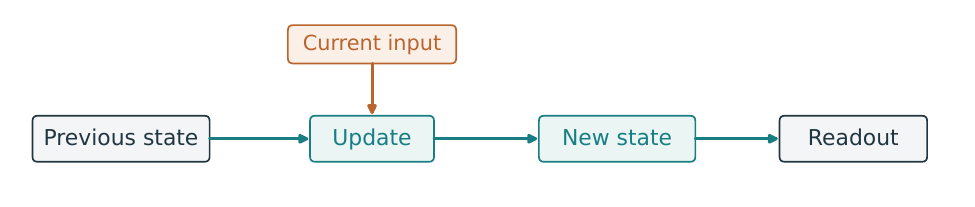}
    \caption{A state-space update at one sequence position. The new state supplies both the current readout and the memory carried to the next position.}
    \label{fig:ssm-v7-a}
\end{figure}

In the memory-centric framework of Section~\ref{sec:unified-memory-centric-view}, these designs primarily shape Memory Representation and Memory Update, while later variants also make Readout more explicit. Figure~\ref{fig:ssm-v7-a} illustrates the basic update at one sequence position: the previous state and current input determine a new state, from which the layer reads a contextual representation. We begin with structured time-invariant dynamics and selective control, including their sequence-processing algorithms, and then examine how input--output interfaces, head organization, and update geometry extend the basic state-space layer.

\subsection{State Dynamics and Selective Control}
\label{subsec:ssm-dynamics-v2}

An SSM state is a set of activations carried between sequence positions. Model weights encode what is learned across training examples, whereas this state records information about the sequence currently being processed. The transition determines which state components persist and how they interact; the input and readout maps connect that state to the surrounding network. Time-invariant and input-conditioned SSMs differ in whether these operators are shared across positions or generated in response to the current input.

\subsubsection{Structured Time-Invariant Dynamics}
\label{subsubsec:ssm-lti-v2}

A basic discrete state-space kernel can be written as
\begin{equation}
    h_t=\bar A h_{t-1}+\bar Bx_t,
    \qquad
    r_t=Ch_t,
    \label{eq:ssm-kernel-v2}
\end{equation}
where $x_t\in\mathbb{R}^{m}$ is the current input, $h_t\in\mathbb{R}^{n}$ is the memory state, and $r_t\in\mathbb{R}^{p}$ is the contextual readout. The transition $\bar A$ retains and mixes existing state components, $\bar B$ writes the new input, and $C$ extracts information for the current computation. The bars denote discrete-time parameters, obtained, for example, by discretizing a continuous-time system. Equation~\eqref{eq:ssm-kernel-v2} isolates the state-space kernel; a complete layer can also include a direct input--output branch, nonlinearities, gates and output projections.

For a time-invariant kernel, learned operators are shared across positions and act on a continually changing sequence state. Starting from $h_0=0$, unrolling the recurrence gives
\begin{equation}
    r_t=\sum_{i=1}^{t}C\bar A^{t-i}\bar Bx_i.
    \label{eq:ssm-convolution-v2}
\end{equation}
Each input is written through $\bar B$, propagated for the elapsed number of steps, and read through $C$. The coefficient $C\bar A^{t-i}\bar B$ depends on the lag $t-i$, yielding a convolutional view for whole-sequence processing and a recurrent view for incremental decoding. The two views expose the same dynamics to different computational settings: training can process a sequence collectively, while decoding updates the state as new tokens arrive.

HiPPO provides an analytic foundation for constructing history-preserving dynamics \citep{gu2020hippo}. It represents an input history through coefficients of a polynomial approximation under a chosen measure over the past. Intuitively, the state tracks a set of features of the historical signal, with the measure determining how different parts of the past contribute and how the coefficients evolve over time. Different measures yield dynamics with different time dependencies; here the construction provides operators and initializations for the structured SSMs that follow. The approximation viewpoint makes the role of state dimension concrete: it controls the number of coefficients available to represent the history.

LSSL introduced trainable state-space layers \citep{gu2021lssl}. S4 made long-sequence computation practical through a structured transition parameterization, commonly initialized from HiPPO, while learning dynamics and input--output projections from task data \citep{gu2021s4}. In a suitable basis, S4 represents the transition using diagonal and low-rank components, providing structure for efficient kernel computation. In Equation~\eqref{eq:ssm-convolution-v2}, the transition participates in every lag coefficient. Its structure therefore affects both which temporal features the layer can express and how efficiently the convolution kernel can be constructed and applied. These developments turned a prescribed historical approximation into a trainable sequence-processing component: training adapts the operators, and inference applies them while updating the state for each new sequence.

S4D simplified the transition parameterization to diagonal structure \citep{gu2022s4d}. In a diagonal system, each state coordinate propagates through its own transition coefficient, while the input and readout maps combine these coordinates with the network channels. S5 instead organized the layer as one multi-input multi-output system and used parallel scans \citep{smith2022s5}. These choices illustrate how temporal dynamics and the organization of the recurrent unit jointly shape an SSM layer.

A scan exploits associative composition of successive state transformations. Two updates can be composed into a transformation over their combined interval, and these interval transformations can be combined hierarchically. This permits parallel evaluation of the sequence state updates. Transition structure determines the cost of representing and composing those transformations; it is central to making the scan practical. Convolution and scan thus offer distinct sequence-processing routes, while recurrent decoding carries the current state forward one position at a time.

\subsubsection{Input-Conditioned State Dynamics}
\label{subsubsec:ssm-selective-v2}

A time-invariant kernel applies the same temporal mixing rule to every input sequence. Language modeling often calls for content-dependent retention: an association may need to survive intervening distractors, an irrelevant token may require little writing, and a context change may call for faster replacement of old information. Selective dynamics let the input regulate these operations.

\begin{figure}[!htbp]
    \centering
    \includegraphics[width=16.5cm]{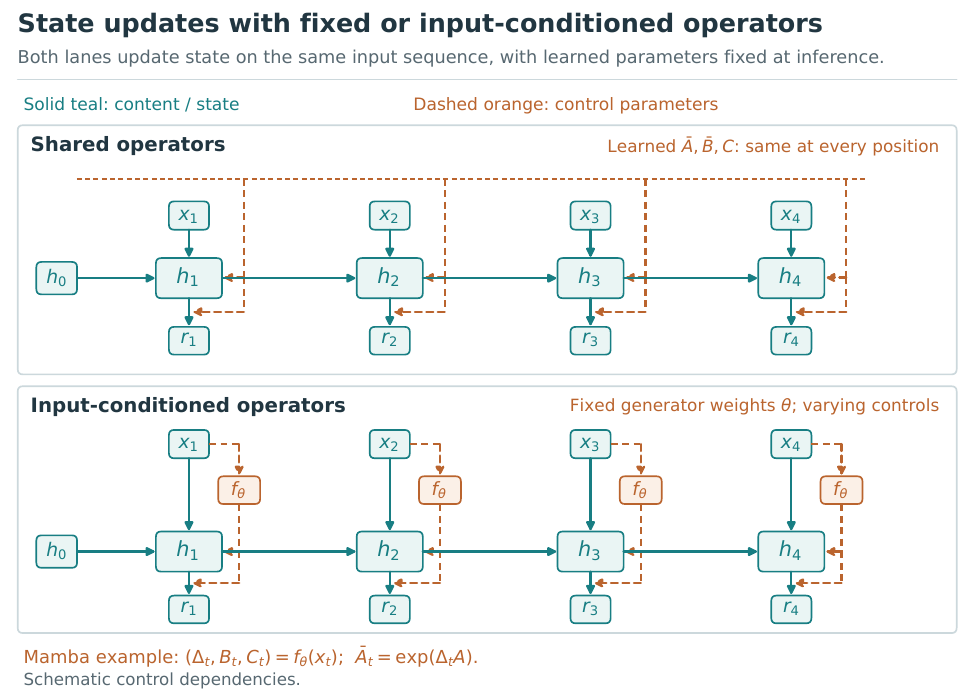}
    \caption{Shared and input-conditioned operators applied to the same input sequence. Both systems update their state at every position. The upper lane reuses learned transition, write and read operators; the lower lane generates control quantities from each input using fixed generator weights. In Mamba \citep{gu2023mamba}, the generated step size modulates the effective discrete transition. Solid paths carry content or state, and dashed orange paths indicate control dependencies.}
    \label{fig:ssm-v7-b}
\end{figure}

Mamba makes the step size $\Delta_t$, input parameter $B_t$, and readout parameter $C_t$ functions of the current input \citep{gu2023mamba}. Its underlying state matrix $A$ is learned during training; the effective discrete transition $\bar A_t=\exp(\Delta_t A)$ varies with the generated step size. The step size controls how far the dynamics advance, the input map controls writing, and the readout map modulates the contribution of state directions to the current representation. During inference, the state and generated controls vary across positions, while the weights of the networks producing those controls remain fixed. Figure~\ref{fig:ssm-v7-b} contrasts the resulting control paths with shared operators.

Input conditioning also changes the available computation. The contribution of an earlier token now depends on controls generated along the intervening sequence, so a single lag-based convolution is insufficient for the general selective kernel. Mamba instead uses a structured selective scan \citep{gu2023mamba}. Mamba-2 develops block-oriented computation through Structured State Space Duality (SSD), connecting a constrained state-space kernel to an attention-like matrix formulation \citep{dao2024mamba2}. This connection makes the recurrence's structure available to efficient matrix-based sequence computation.

For a Mamba-2 head, let $H_t\in\mathbb{R}^{N\times P}$ be a matrix state with memory dimension $N$ and value-channel width $P$. Writing the projected head input as $u_t\in\mathbb{R}^{P}$, its core recurrence and readout take the form
\begin{equation}
    H_t=a_tH_{t-1}+b_tu_t^{\top},
    \qquad
    r_t=H_t^{\top}c_t,
    \label{eq:ssm-ssd-v2}
\end{equation}
where $a_t$ is a scalar transition shared within the head, $b_t,c_t\in\mathbb{R}^{N}$ are write and read directions, and the write scale is absorbed into $b_t$.\footnote{Here inputs and readouts are column vectors. Equivalently, a row read map $C_t^{\mathrm{row}}=c_t^{\top}$ gives $r_t^{\top}=C_t^{\mathrm{row}}H_t$.} From a zero initial state,
\begin{equation}
    r_t=\sum_{i=1}^{t}
    \underbrace{\left(c_t^{\top}b_i\right)}_{\text{read--write match}}
    \underbrace{\left(\prod_{j=i+1}^{t}a_j\right)}_{\text{intervening decay}}
    u_i.
    \label{eq:ssm-ssd-expansion-v2}
\end{equation}
The contribution of an earlier input factors into its match with the current read direction and the intervening transitions. The head-wise scalar transition enables this factorization, relating the SSM to the associative computation used in Linear Attention. Denote the scalar coefficient of $u_i$ in Equation~\eqref{eq:ssm-ssd-expansion-v2} by $M_{ti}$. These coefficients form a lower-triangular matrix: row $t$ corresponds to a readout position and column $i$ to a source position. On the diagonal, the transition product is empty and equals one.

\begin{figure}[H]
    \centering
    \includegraphics[width=16.5cm]{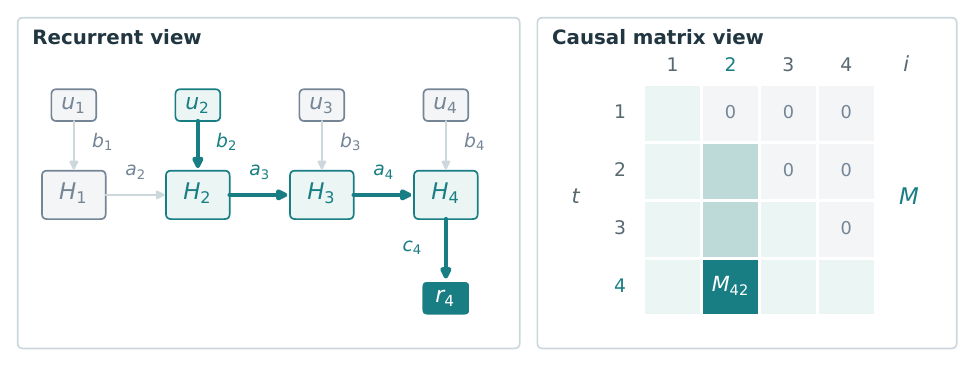}
    \caption{Two views of the SSD kernel in Equations~\eqref{eq:ssm-ssd-v2}--\eqref{eq:ssm-ssd-expansion-v2} \citep{dao2024mamba2}. The highlighted recurrent path and matrix entry $M_{42}$ identify the same contribution, from input $u_2$ to readout $r_4$.}
    \label{fig:ssm-v7-c}
\end{figure}

Figure~\ref{fig:ssm-v7-c} highlights the contribution of $u_2$ to $r_4$. The input is written along $b_2$, propagated by $a_3$ and $a_4$, and read along $c_4$, giving $M_{42}=(c_4^{\top}b_2)a_3a_4$. Multiplying this coefficient by $u_2$ gives that input's contribution to the readout; the intervening states also contain contributions from other inputs. Whole-sequence algorithms exploit the matrix structure in blocks, while incremental decoding evaluates the same kernel using only its current recurrent state.

Table~\ref{tab:ssm-kernel-choices-v2} compares the dynamics and computational interfaces of these representative kernels.

\begin{table}[!htbp]
    \centering
    \small
    \caption{Representative SSM dynamics and computational interfaces. The Mamba-2 SSD kernel is a selective parameterization with a head-wise scalar transition.}
    \label{tab:ssm-kernel-choices-v2}
    \begin{tabularx}{\linewidth}{@{}>{\raggedright\arraybackslash}p{0.22\linewidth}>{\raggedright\arraybackslash}p{0.34\linewidth}>{\raggedright\arraybackslash}X@{}}
        \toprule
        \textbf{Kernel choice} & \textbf{State control} & \textbf{Computational interface} \\
        \midrule
        Learned time-invariant dynamics: S4, S4D, S5 \citep{gu2021s4,gu2022s4d,smith2022s5} & Shared operators govern input writing, temporal propagation and readout. & Convolution or structured scans process sequences; recurrent decoding applies the same learned dynamics to the evolving state. \\
        Selective dynamics: Mamba \citep{gu2023mamba} & Input-generated step sizes and write/read maps regulate the effective operators. & A selective scan processes the input-dependent recurrence. \\
        SSD kernel: Mamba-2 \citep{dao2024mamba2} & A head-wise scalar transition combines with input-dependent write and read directions. & Recurrent and structured-matrix views support block computation and state-based decoding. \\
        \bottomrule
    \end{tabularx}
\end{table}

For a fixed number of heads and fixed state dimensions $N$ and $P$, the Mamba-2 kernel maintains context-length-independent state and per-step decoding work, while total sequence work grows linearly with the number of positions \citep{dao2024mamba2}. The $N\times P$ head state determines the retained storage, while the sequence algorithm determines how updates are scheduled and mapped to hardware. Increasing the state dimensions raises the storage and per-step arithmetic costs. The practical design therefore couples a useful state parameterization with an efficient implementation of its updates and readouts.

\FloatBarrier
\subsection{State Organization and Read--Write Refinements}
\label{subsec:ssm-refinements-v2}

State dynamics specify how information evolves; the surrounding interface determines what can be written and extracted at each position. H3 used associative-recall and induction-head tasks to study shortcomings of early SSMs and introduced multiplicative interactions to strengthen the relevant computation \citep{fu2022h3}. Zoology further showed that aggregate language-modeling quality can conceal substantial differences in associative recall \citep{arora2024zoology}. These results motivate closer examination of the operations connecting inputs, states and queries.

\subsubsection{Read--Write Interfaces}
\label{subsubsec:ssm-interfaces-v2}

The input and output ports of the specified recurrent unit give one useful description of this interface. In Equation~\eqref{eq:ssm-kernel-v2}, scalar ports $m=p=1$ define a single-input single-output (SISO) system with an $n$-dimensional internal state. A multi-input multi-output (MIMO) system has vector-valued ports. State size determines the internal representation, while port dimensions determine the number of input and output components connected to that representation.

S5 uses a joint MIMO system in the time-invariant setting \citep{smith2022s5}. By comparison, the $P$ columns in the Mamba-2 kernel of Equation~\eqref{eq:ssm-ssd-v2} can be read as $P$ scalar-port recurrences sharing coefficients. Each column receives one scalar component of $u_t$, maintains $N$ state values, and produces one component of $r_t$. The head has $NP$ state values in total. Stating the recurrent unit at this level separates the value-channel width from both its internal memory dimension and the extra read--write directions considered next.

The MIMO variant of Mamba-3 expands the read--write channels available to the principal state \citep{gu2026mamba3}. Whereas Equation~\eqref{eq:ssm-ssd-v2} uses one read vector $c_t$, a matrix $C_t^{\mathrm{multi}}\in\mathbb{R}^{N\times R}$ supplies $R$ directions,
\begin{equation}
    Y_t^{\mathrm{multi}}=H_t^{\top}C_t^{\mathrm{multi}}\in\mathbb{R}^{P\times R},
    \label{eq:ssm-multichannel-v2}
\end{equation}
and these channels are combined into the layer output. Writing is similarly extended by combining multiple outer-product contributions. The construction increases the work performed with a principal state: each read direction projects the same retained information differently, and the resulting channels provide a richer interface to the network. For the read projection shown above, direct multiplication requires work proportional to $NPR$, exposing how the number of directions changes the arithmetic. When state movement dominates decoding time, this additional arithmetic can improve hardware utilization; realized latency depends on the implementation, device and batch size.

MIMOMamba develops a different multichannel construction by generalizing scalar SSD to matrix-valued attention \citep{wang2026mimomamba}. Within a head, transition and input--output matrices are parameterized through a polynomial algebra generated from shared base matrices. Their commuting structure supports the associated matrix-valued factorization. Affine state-update composition supplies the associativity used by parallel scans, while the polynomial construction supplies the additional algebraic structure for this particular dual formulation. Mamba-3 MIMO and MIMOMamba thus enrich state interfaces through different choices of parameter sharing and operator structure.

Figure~\ref{fig:ssm-v7-d}(a,b) contrasts increasing the state dimensions with adding read--write directions to the same principal state. Together, the interfaces above determine the calculations performed on the retained state and the channels through which its information reaches the layer output. Their value depends on the information preserved by the dynamics: a richer projection exposes additional views of the state, while the transition and writing process determine the historical distinctions encoded there.

\FloatBarrier
\subsubsection{State Organization and Update Geometry}
\label{subsubsec:ssm-organization-v6}

\begin{figure}[!htbp]
    \centering
    \includegraphics[width=16.5cm]{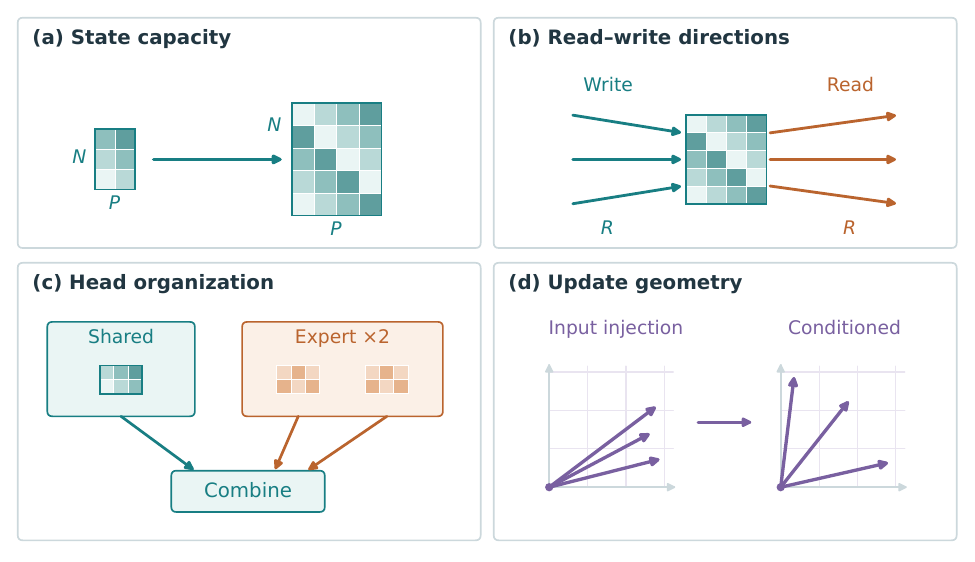}
    \caption{Four ways to refine an SSM layer: (a) enlarge the state, (b) add read--write directions, (c) combine shared and expert temporal components as in HADES \citep{shin2026hades}, and (d) condition input-injection directions as in MuonSSM \citep{nguyen2026muonssm}. Panels (c,d) are conceptual sketches of the mechanisms described in the text.}
    \label{fig:ssm-v7-d}
\end{figure}

The layer can further organize temporal processing across heads. HADES interprets Mamba-2's multi-head recurrences as an adaptive filter bank \citep{shin2026hades}. It combines shared filters for global low-pass behavior with expert filters for local high-pass behavior, as sketched in Figure~\ref{fig:ssm-v7-d}(c). The shared component carries more slowly varying information; expert components emphasize more rapidly changing contributions. This organizes complementary temporal characteristics within the SSM layer, alongside the choice of read--write interfaces.

The geometry of the injected updates provides another way to improve how the state evolves. MuonSSM augments an SSM with a momentum-based pathway and lightweight Newton--Schulz iterations on low-rank input injections \citep{nguyen2026muonssm}. The momentum pathway carries information across updates, while the orthogonalization procedure conditions their directional structure. Figure~\ref{fig:ssm-v7-d}(d) illustrates this change in the geometry of input injections. In the state-space recurrence, this intervention targets the contribution written into memory, keeping the learned temporal transition as a separate component. It therefore adds update processing to the choices of retention and input--output mapping already described.

These designs operate at different locations in the state computation. Head organization distributes temporal characteristics across interacting components; update conditioning shapes the signal entering the state. This distinction helps explain how an SSM layer can be refined while retaining recurrent execution. The transition's expressiveness remains important as well: state-tracking analyses identify limitations of particular transition classes under specified architectural and numerical assumptions \citep{merrill2024illusion}. Examining the transition, interface and update together is consequently useful for understanding the behavior of a concrete SSM.

\FloatBarrier
\subsection{Summary}
\label{subsec:ssm-summary-v2}

SSM sequence layers combine a retained state with learnable temporal dynamics. Structured time-invariant models provide tractable ways to represent and propagate historical features. Selective models let input-generated controls modulate state evolution, writing and reading, and SSD connects a constrained selective kernel to matrix-based computation. Subsequent designs enrich the channels connected to the state, organize its temporal processing across heads, and condition the geometry of incoming updates. Table~\ref{tab:ssm-design-summary-v7} summarizes these SSM-specific choices and their computational roles.

These mechanisms also supply components for hybrid architectures. Attamba uses SSMs to form block-level keys and values \citep{akhauri2024attamba}, while DART retains local SSM block-state contributions for query-time key and value generation \citep{qian2026dart}. Samba combines Mamba with sliding-window attention \citep{ren2024samba}. The following chapter examines how such state-based components are coordinated with other sequence-processing paths.

\begin{table}[!htbp]
    \centering
    \small
    \renewcommand{\arraystretch}{1.08}
    \caption{SSM design questions, mechanisms and roles. The fixed-size decoding state assumes a fixed head count and fixed state dimensions.}
    \label{tab:ssm-design-summary-v7}
    \begin{tabularx}{\linewidth}{@{}>{\raggedright\arraybackslash}p{0.23\linewidth}>{\raggedright\arraybackslash}p{0.34\linewidth}>{\raggedright\arraybackslash}X@{}}
        \toprule
        \textbf{Design question} & \textbf{SSM mechanism} & \textbf{Computational or state effect} \\
        \midrule
        How should the state propagate and adapt? & Structured dynamics and input-dependent controls. & Temporal propagation, selective writing and readout. \\
        \addlinespace[3pt]
        How should the recurrence be computed? & SSD recurrent and matrix views; block computation. & Structured sequence computation and a fixed-size decoding state. \\
        \addlinespace[3pt]
        How should the state communicate? & Multichannel interfaces: S5, Mamba-3 MIMO and MIMOMamba. & Richer input--output interaction with the retained state. \\
        \addlinespace[3pt]
        How should heads and updates be organized? & HADES head organization; MuonSSM update geometry. & Complementary temporal roles; conditioned input-injection directions. \\
        \bottomrule
    \end{tabularx}
\end{table}
\FloatBarrier

% !TEX root = ../main.tex
\section{Hybrid Architecture: Composing Heterogeneous Memory Mechanisms}
\label{sec:hybrid-architectures}

This section takes composition granularity as its organizing principle and examines how heterogeneous attention and state-based memory mechanisms are combined at the layer, head, branch, and token levels. Hybrid Architecture is treated as a compositional design space rather than as a fifth memory operator parallel to Softmax Attention, Sparse Attention, Linear Attention, and State Space Models. At a coarse granularity, a hybrid may arrange largely self-contained modules across network depth; at finer granularities, it may redesign the internal organization of heads, branches, token routing, memory transitions, and fusion interfaces. Accordingly, this section focuses on where heterogeneous mechanisms are combined and how their interactions change across different structural granularities.

Softmax Attention, Sparse Attention, Linear Attention, and State Space Models provide complementary capability--cost profiles. Explicit attention preserves direct token-level addressability, Sparse Attention restricts the candidate set, and Linear Attention and SSMs propagate history through bounded recurrent states. Hybrid designs combine these capabilities so that fine-grained retrieval, local interaction, sparse access, and efficient long-range propagation need not be provided by one sequence mixer.

We organize Hybrid Architectures by the structural granularity at which heterogeneous mechanisms are combined. \emph{Layer-wise Hybrid} assigns different mixers across network depth; \emph{Head-wise Hybrid} partitions mechanisms across heads or channel groups within a layer; \emph{Branch-wise Hybrid} maintains and coordinates multiple memory paths inside a block; and \emph{Token-wise Hybrid} assigns tokens or chunks to different memory representations or sequence-mixing operations. These granularities may coexist within one architecture and together extend contextual memory from a single substrate or processing path to a heterogeneous organization of explicit and recurrent memories. Under the five-dimensional view in Section~\ref{sec:unified-memory-centric-view}, such composition first broadens Memory Representation and, depending on how tightly the paths are coupled, may further coordinate how information is updated or transferred, which memory interfaces are exposed, how their contents are read, and how the resulting features are integrated. Composition granularity therefore identifies the structural scope of hybridization, while the five dimensions describe its consequences for memory processing.

\begin{figure}[t]
    \centering
    \includegraphics[width=\textwidth]{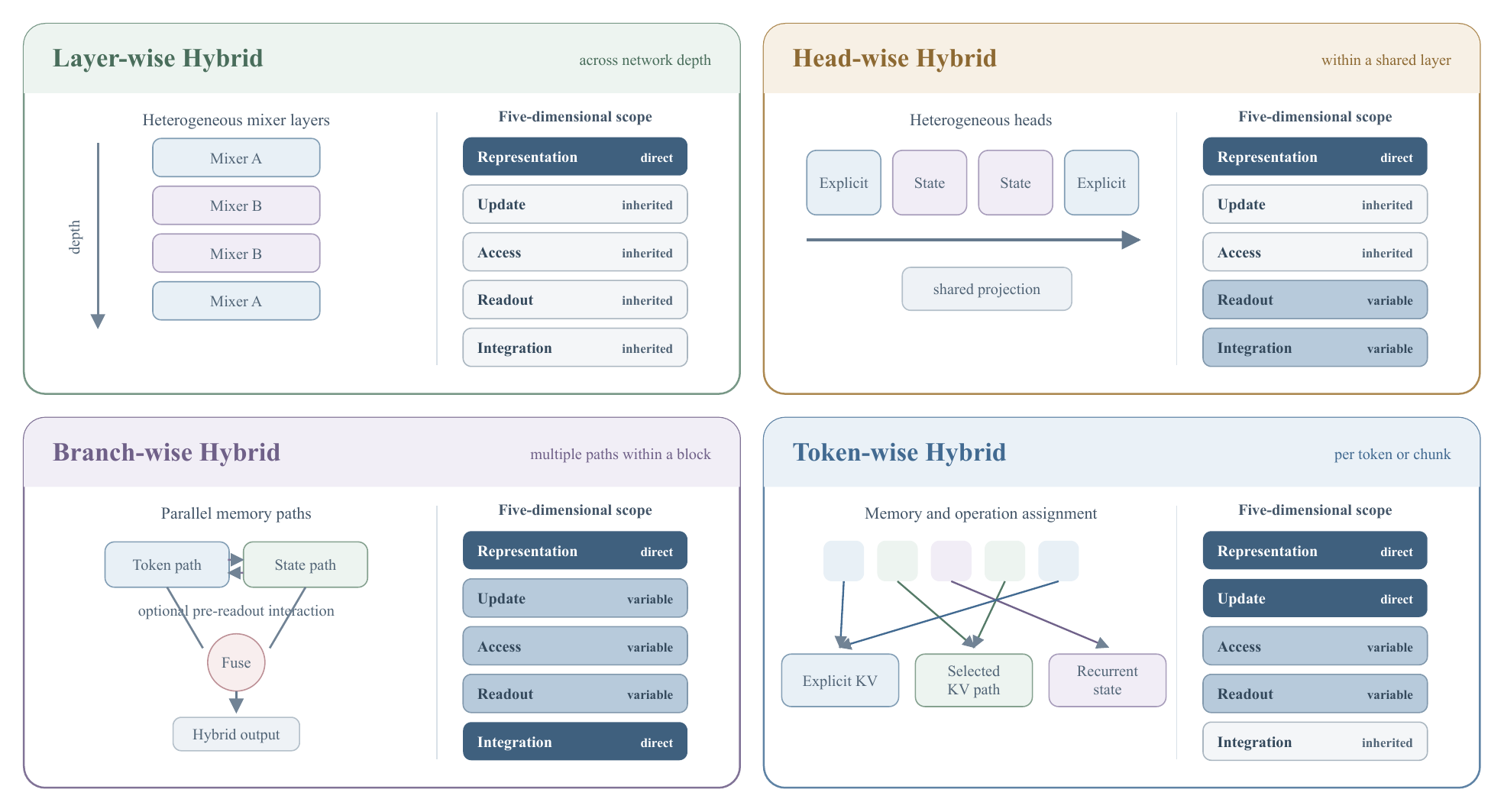}
    \caption{Four composition granularities of Hybrid Architecture and their typical intervention scopes under the five-dimensional memory-centric view. The schematics show where heterogeneous mechanisms are combined. The five-cell columns correspond to Memory Representation, Memory Update, Access, Readout, and Integration. Dark, medium, and light blocks denote dimensions commonly modified directly by hybrid composition, affected in a method-dependent manner, or primarily inherited from the constituent mechanisms, respectively. The profiles are qualitative summaries rather than quantitative literature counts.}
    \label{fig:hybrid-taxonomy}
\end{figure}

Figure~\ref{fig:hybrid-taxonomy} summarizes where heterogeneous mechanisms are combined and qualitatively relates each composition granularity to its typical intervention scope under the five-dimensional view. The following subsections compare Layer-wise designs by layer-allocation policy, Head-wise designs by head budget, Branch-wise designs by interaction interface, and Token-wise designs by routing signal. Tables~\ref{tab:hybrid-layer-wise}--\ref{tab:hybrid-token-wise} then report the corresponding categories, periods, and representative methods before the text examines their motivations and implementations.

\subsection{Layer-wise Hybrid}
\label{subsec:layer-wise-hybrid}

Layer-wise Hybrid assigns different sequence mixers to different depths of a network, combining mechanisms such as Softmax or Linear Attention, SSMs, local attention, and Sparse Attention within a single backbone. The principal distinction is the source of the layer-allocation decision. Some architectures determine module positions before training through a fixed ratio or periodic schedule. Others begin with a pretrained model and use layer behavior to decide which attention modules should be retained or replaced. In the latter group, the selection signal may come directly from diagnosis of the original layers, from representation differences after a candidate replacement module has been aligned, or from teacher--student discrepancy during iterative distillation. Table~\ref{tab:hybrid-layer-wise} therefore distinguishes Predefined Allocation, Direct Diagnosis Selection, Alignment-Based Selection, and Iterative Distillation-Guided Selection.

\begin{table}[!htbp]
    \centering
    \small
    \caption{Layer-wise Hybrid designs grouped by layer-allocation policy.}
    \label{tab:hybrid-layer-wise}
    \begin{tabularx}{\textwidth}{@{}>{\raggedright\arraybackslash}p{0.21\textwidth}>{\raggedright\arraybackslash}p{0.18\textwidth}>{\centering\arraybackslash}p{0.10\textwidth}>{\raggedright\arraybackslash}X@{}}
        \toprule
        \textbf{Layer allocation} & \textbf{Training strategy} & \textbf{Period} & \textbf{Representative methods} \\
        \midrule
        Predefined allocation & From-scratch training & 2024--2026 & Griffin \citep{de2024griffin}; Jamba \citep{lieber2024jamba}; Samba \citep{ren2024samba}; Zamba \citep{glorioso2024zamba}; Zamba2 \citep{zamba2024zamba2}; HySparse \citep{zhao2026hysparse}; HySparse2 \citep{wei2026hysparse2} \\
        Direct diagnosis selection & Pretrained-model transfer & 2024--2026 & LightTransfer \citep{zhang2024lighttransfer}; Priming \citep{priming2026hybrid} \\
        Alignment-based selection & Pretrained-model transfer & 2025--2026 & Jet-Nemotron / PostNAS \citep{gu2025jetnemotron}; HypeNet / HALO \citep{chen2026hypenet} \\
        Iterative distillation-guided selection & Pretrained-model transfer & 2025 & KL-Guided Layer Selection \citep{li2025klguided} \\
        \bottomrule
    \end{tabularx}
\end{table}

\subsubsection{Predefined Allocation}
\label{subsubsec:hybrid-predefined-layer}

Many native Hybrid models use manually specified layer structures. Griffin \citep{de2024griffin} configures gated linear recurrent blocks and local-attention blocks at an approximate ratio of $2{:}1$, whereas Jamba \citep{lieber2024jamba} uses Mamba as its dominant path and retains full-attention layers at an approximate ratio of $1{:}7$ relative to Mamba layers. Samba likewise adopts a predetermined layer-wise arrangement of Mamba, sliding-window attention, and MLP blocks, using recurrent propagation as the backbone while periodically providing direct local retrieval \citep{ren2024samba}. These designs control where different context-processing capabilities appear across depth without fusing their readouts inside one block.

Zamba and Zamba2 periodically invoke attention modules that are shared across depth within a Mamba backbone. Zamba repeatedly uses one shared attention block. Zamba2 retains parameter sharing but alternates between two shared attention blocks and adds LoRA projection modules to provide lightweight depth-specific adaptation at different invocation sites \citep{glorioso2024zamba,zamba2024zamba2}. HySparse similarly organizes the model with a fixed full/sparse layer ratio. Within its sparse layers, however, it computes global sparse attention and local sliding-window attention in parallel, while reusing the KV representations and selection indices of a preceding full-attention layer \citep{zhao2026hysparse}. HySparse2 keeps this predefined organization but partitions the backbone into a self-decoder that interleaves full attention with sliding-window attention and a cross-decoder that interleaves full attention with token-level sparse attention, connecting the two through full-attention KV bridging so that each cross-decoder full-attention layer projects its keys and values from self-decoder hidden states \citep{wei2026hysparse2}. Its detailed selection and cross-layer reuse mechanism is treated in Section~\ref{subsec:sparse-reuse}.

\subsubsection{Direct Diagnosis Selection}
\label{subsubsec:hybrid-direct-diagnosis}

Direct diagnosis methods estimate the replaceability of attention layers in a pretrained model without performing complete replacement-module alignment and repeated distillation before every selection decision. LightTransfer identifies ``lazy'' layers that depend less on global attention and prioritizes their conversion to streaming or local attention \citep{zhang2024lighttransfer}. Priming uses a block-importance score to select Transformer layers that should be retained and applies spectral initialization when converting the remaining attention blocks into SSM modules \citep{priming2026hybrid}. Both derive their selection signal directly from properties of the original model, but use different migration procedures: lightweight adaptation in LightTransfer and training-free conversion in Priming.

\subsubsection{Alignment-Based Selection}
\label{subsubsec:hybrid-alignment-selection}

Alignment-based methods first reduce the representation gap between Softmax Attention and a candidate efficient module, and then evaluate the effect of replacement at different depths. Jet-Nemotron performs hidden-state alignment before PostNAS jointly searches full-attention placement, the efficient attention operator, and hardware-aware hyperparameters; the resulting Hybrid model is subsequently fine-tuned \citep{gu2025jetnemotron}. HypeNet/HALO performs attention-weight transfer and hidden-state alignment before selecting the full-attention layers to retain, followed by distillation and fine-tuning to recover the converted model's capability \citep{chen2026hypenet}. Both follow an alignment--selection--adaptation sequence, differing mainly in the search space and subsequent optimization procedure.

\subsubsection{Iterative Distillation-Guided Selection}
\label{subsubsec:hybrid-iterative-selection}

KL-Guided Layer Selection alternates student adaptation with layer selection. It first performs attention-weight transfer and hidden-state alignment, and distills an initial Linear Attention student. Softmax Attention is then restored one layer at a time, with the resulting reduction in teacher--student KL divergence used to estimate that layer's importance; distillation continues after each selection round \citep{li2025klguided}. The selection signal therefore reflects the behavior of an already adapted student rather than statistics of the original model or a one-time alignment result.

This progression moves Layer-wise Hybrid design from uniform periodic schedules specified during architecture construction toward layer-specific allocation informed by pretrained-model behavior. In parallel, model construction expands beyond training from scratch to adaptation, distillation, fine-tuning, and training-free conversion. These training strategies support the allocation decision, but remain distinct from the architectural question of which mechanism is deployed at each depth and from the memory-level Access rule used inside that mechanism.

\subsection{Head-wise Hybrid}
\label{subsec:head-wise-hybrid}

Head-wise Hybrid assigns different heads or channel groups within the same layer to distinct sequence mixers, allowing full-context retrieval, streaming or sparse token interaction, and state-based propagation to coexist at every depth. Its central question is how the head budget is apportioned among these mechanisms and whether the assignment is predefined, selected from head function, adapted to the current input, or varied across network depth. Table~\ref{tab:hybrid-head-wise} groups existing methods into Fixed Head Allocation, Function-Aware Head Selection, Input-Adaptive Head Routing, and Depth-Adaptive Head Allocation.

\begin{table}[!htbp]
    \centering
    \small
    \caption{Head-wise Hybrid designs grouped by the policy used to determine the head allocation.}
    \label{tab:hybrid-head-wise}
    \begin{tabularx}{\textwidth}{@{}>{\raggedright\arraybackslash}p{0.22\textwidth}>{\centering\arraybackslash}p{0.10\textwidth}>{\raggedright\arraybackslash}p{0.39\textwidth}>{\raggedright\arraybackslash}X@{}}
        \toprule
        \textbf{Head allocation} & \textbf{Period} & \textbf{Allocation characteristic} & \textbf{Representative methods} \\
        \midrule
        Fixed head allocation & 2024--2025 & A predefined ratio is used across Hybrid layers. & Hymba \citep{dong2024hymba}; Falcon-H1 \citep{zhai2025falconh1} \\
        Function-aware head selection & 2024--2026 & Heads with retrieval or long-range functions retain full attention, while the remaining heads use streaming, sparse, or recurrent paths. & DuoAttention \citep{xiao2024duoattention}; HydraHead \citep{tan2026hydrahead} \\
        Input-adaptive head routing & 2026 & A test-time router assigns heads to full-attention or sparse-attention modes according to the input, allowing the overall sparsity ratio to vary dynamically. & Elastic Attention \citep{tang2026elasticattention} \\
        Depth-adaptive head allocation & 2026 & The hybrid ratio is varied across depth. & Head-wise Hybridization \citep{shi2026implicithybrids} \\
        \bottomrule
    \end{tabularx}
\end{table}

\subsubsection{Fixed Head Allocation}
\label{subsubsec:hybrid-fixed-head}

Fixed-allocation methods specify the ratio of the two head types during architecture design and keep it identical or approximately identical across the Hybrid layers of a given configuration. Hymba places attention and SSM heads in parallel within a symmetrically formulated fusion module, while its final configurations use an SSM-dominant allocation in which attention heads occupy no more than approximately one fifth of the Mamba heads \citep{dong2024hymba}. Falcon-H1 uses a more SSM-heavy asymmetric allocation; in its 7B configuration, the attention-to-SSM head budget is approximately $1{:}2$ \citep{zhai2025falconh1}. Both are native Head-wise Hybrid architectures rather than methods that choose particular positions from a pretrained set of full-attention heads.

\subsubsection{Function-Aware Head Selection}
\label{subsubsec:hybrid-function-head}

DuoAttention separates pretrained attention heads according to their long-context functions. It learns a scalar gate for each head by minimizing the output deviation between full attention and a gated mixture of full and streaming attention, and then binarizes the gates for deployment. Retrieval heads retain full attention and a complete KV cache, whereas streaming heads retain only attention sinks and recent tokens \citep{xiao2024duoattention}. DuoAttention therefore combines full-context and structurally sparse Softmax Attention within each layer, using an optimization-based estimate of which heads require global retrieval.

HydraHead extends function-aware selection across different sequence-mixing families. It analyzes the roles of attention heads in retrieval and long-range information processing, preserves the head positions whose functions depend more strongly on full attention, and converts the remaining heads to GDN paths \citep{tan2026hydrahead}. The resulting architecture can retain a small aggregate full-attention budget---for example, approximately one full-attention head for every seven GDN heads---but this ratio describes overall resource allocation. Both DuoAttention and HydraHead produce a fixed head assignment after the identification or conversion stage; they differ in whether the efficient heads use streaming attention or a recurrent GDN mechanism.

\subsubsection{Input-Adaptive Head Routing}
\label{subsubsec:hybrid-input-adaptive-head}

Elastic Attention combines full attention and streaming sparse attention within the same layer, but does not fix one head assignment for every input. A lightweight attention router uses the current sequence representation to assign individual heads to full-attention or sparse-attention computation modes at test time, allowing the model-level sparsity ratio to adapt to the input's sensitivity to sparse retrieval \citep{tang2026elasticattention}. This distinguishes Elastic Attention from offline function-aware selection: DuoAttention and HydraHead identify a deployment topology before inference, whereas Elastic Attention changes the active full/sparse head allocation across inputs.

\subsubsection{Depth-Adaptive Head Allocation}
\label{subsubsec:hybrid-depth-head}

Head-wise Hybridization further allows the full-attention/GDN budget to vary with depth. For layer $\ell$,
\begin{equation}
    H_{\ell}^{\mathrm{FA}}+H_{\ell}^{\mathrm{Eff}}=H,
    \qquad
    H_{\ell}^{\mathrm{Eff}}:H_{\ell}^{\mathrm{FA}}=k_{\ell}:1,
    \label{eq:hybrid-depth-head-budget}
\end{equation}
where $k_{\ell}$ changes across the network so that each depth receives a different amount of exact token-retrieval capacity according to its functional requirements \citep{shi2026implicithybrids}. Function-aware selection asks which head positions should preserve full attention, input-adaptive routing asks which computation mode each head should use for the current input, and depth-adaptive allocation asks how much full-attention capacity should be assigned at each depth.

\subsection{Branch-wise Hybrid}
\label{subsec:branch-wise-hybrid}

Branch-wise Hybrid maintains multiple comparatively complete information streams with distinct Memory Representations, Updates, or Readout rules inside the same block. Head-wise Hybrid instead partitions heads or channel groups within a common multi-head structure and typically concatenates their readouts before a shared output projection. We distinguish Branch-wise designs by whether the paths are combined after producing contextual readouts $r_t^{(p)}$ or after independently forming path outputs $o_t^{(p)}$. The fusion operator alone does not determine the category; the defining property is that the architecture preserves identifiable memory paths rather than partitioning head or channel capacity within one shared module. Table~\ref{tab:hybrid-branch-wise} summarizes the two interfaces.

\begin{table}[!htbp]
    \centering
    \small
    \caption{Branch-wise Hybrid designs grouped by the interface at which heterogeneous memory paths are combined.}
    \label{tab:hybrid-branch-wise}
    \begin{tabularx}{\textwidth}{@{}>{\raggedright\arraybackslash}p{0.24\textwidth}>{\centering\arraybackslash}p{0.24\textwidth}>{\centering\arraybackslash}p{0.10\textwidth}>{\raggedright\arraybackslash}X@{}}
        \toprule
        \textbf{Fusion interface} & \textbf{Unified form} & \textbf{Period} & \textbf{Representative methods} \\
        \midrule
        Readout-level branch fusion & $r_t^{\mathrm{hyb}}=\operatorname{Fuse}_r(\{r_t^{(p)}\})$ & 2022--2026 & Block-Recurrent Transformer \citep{hutchins2022blockrecurrent}; Memorizing Transformer \citep{wu2022memorizing}; Infini-attention \citep{mansukhani2024infini}; DART \citep{qian2026dart} \\
        Output-level branch fusion & $o_t^{\mathrm{hyb}}=\operatorname{Fuse}_o(\{o_t^{(p)}\})$ & 2025 & Titans-MAG \citep{behrouz2025titans} \\
        \bottomrule
    \end{tabularx}
\end{table}

\subsubsection{Readout-Level Branch Fusion}
\label{subsubsec:hybrid-readout-fusion}

Readout-level designs combine contextual representations produced by distinct paths before a shared output transformation:
\begin{equation}
    r_t^{\mathrm{hyb}}
    =
    \operatorname{Fuse}_{r}
    \left(\{r_t^{(p)}\}_{p\in\mathcal{P}_{t}};x_t\right),
    \qquad
    o_t^{\mathrm{hyb}}
    =
    \operatorname{Integration}
    \left(r_t^{\mathrm{hyb}};x_t\right).
    \label{eq:hybrid-readout-fusion}
\end{equation}
The methods in this category differ in the memories read by each path, the degree of cross-path dependence, and the fusion rule applied to their readouts.

Block-Recurrent Transformer maintains a token stream and a fixed set of recurrent state vectors. In the token direction, token self-attention and token-to-state cross-attention produce parallel readouts that are concatenated before a shared projection. The state direction mirrors this structure: state self-attention and state-to-token cross-attention are combined before gated recurrent updating. It therefore performs readout-level fusion in both directions while jointly producing the current token representations and the recurrent state for the next block \citep{hutchins2022blockrecurrent}. Unlike the associative matrix used by standard Linear Attention, its recurrent memory consists of multiple explicit state vectors that are read through ordinary attention.

Memorizing Transformer provides a boundary case for the same interface. It separately reads the local KV context and a top-$k$ external kNN memory, then combines the two per-head attention results through a learned gate before the usual multi-head output transformation \citep{wu2022memorizing}. Although the external datastore lies outside the model-internal memory emphasized by the core taxonomy, its gated readout interface is directly comparable to internal branch fusion.

Infini-attention implements the same ordering with model-internal memories. For each head, it first produces a local causal attention context $A_{\mathrm{dot}}$ and a compressive-memory readout $A_{\mathrm{mem}}$, and then computes
\begin{equation}
    A
    =
    \sigma(\beta)\odot A_{\mathrm{mem}}
    +
    \left(1-\sigma(\beta)\right)\odot A_{\mathrm{dot}}.
    \label{eq:infini-readout-fusion}
\end{equation}
The fused head contexts are subsequently concatenated and projected by the shared multi-head output matrix, making this a clear readout-level fusion \citep{mansukhani2024infini}.

DART couples its paths more tightly because the State-Memory Attention (SMA) branch retrieves from chunk-state memories produced by the Mamba-2 scan and shares the native SSM read vector. Its final combination nevertheless remains at the readout interface: the SMA readout $R_t$ is added as a scalar-gated residual correction to the SSM readout $C_tH_t$,
\begin{equation}
    r_t^{\mathrm{DART}}
    =
    C_tH_t
    +
    G_tR_t,
    \qquad
    G_t=\operatorname{SiLU}(U_tW_G).
    \label{eq:dart-readout-fusion}
\end{equation}
The resulting representation then continues through the remaining block transformations \citep{qian2026dart}. Thus, Block-Recurrent Transformer, Memorizing Transformer, Infini-attention, and DART use different memory substrates and fusion operators, but all combine path readouts before the enclosing module completes its output transformation.

\subsubsection{Output-Level Branch Fusion}
\label{subsubsec:hybrid-output-fusion}

When each branch first forms its own output, Hybrid-level Integration instead acts on $\{o_t^{(p)}\}$:
\begin{equation}
    o_t^{\mathrm{hyb}}
    =
    \operatorname{Fuse}_{o}
    \left(\{o_t^{(p)}\}_{p\in\mathcal{P}_{t}};x_t\right).
    \label{eq:hybrid-output-fusion}
\end{equation}
Titans-MAG follows this pattern. Its short-term branch applies sliding-window attention, while its neural long-term memory branch independently produces a memory output from the same input. The two branch outputs are normalized with learned vector-valued weights and combined through a nonlinear gate,
\begin{equation}
    o_t^{\mathrm{MAG}}
    =
    \operatorname{Gate}
    \left(
        o_t^{\mathrm{SWA}},
        o_t^{\mathrm{mem}}
    \right),
    \label{eq:titans-output-fusion}
\end{equation}
so the fusion directly defines the output of the Hybrid module rather than an intermediate readout awaiting a shared output projection \citep{behrouz2025titans}.

\subsection{Token-wise Hybrid}
\label{subsec:token-wise-hybrid}

Token-wise Hybrid uses token position, a content score, or a learned policy to decide which memory or sequence-mixing path receives a token. Let $z_i$ denote this architecture-level routing decision for token $i$. Depending on the method, the token may remain in recent KV memory, move to sparse historical KV, be compressed into a recurrent state, or be assigned to a different sequence-mixing operation. The routing decision changes memory-level Access only when it changes which represented memory units or state interface are exposed to the current query; routing that merely selects an operator remains mechanism allocation. Table~\ref{tab:hybrid-token-wise} groups the methods by routing signal into Temporal-Boundary Allocation, Content-Score Allocation, and Learned Operation Allocation.

\begin{table}[!htbp]
    \centering
    \footnotesize
    \caption{Token-wise Hybrid designs grouped by the signal used to determine token allocation.}
    \label{tab:hybrid-token-wise}
    \begin{tabularx}{\textwidth}{@{}>{\raggedright\arraybackslash}p{0.19\textwidth}>{\raggedright\arraybackslash}p{0.43\textwidth}>{\centering\arraybackslash}p{0.10\textwidth}>{\raggedright\arraybackslash}X@{}}
        \toprule
        \textbf{Token allocation} & \textbf{Unified selection rule} & \textbf{Period} & \textbf{Representative methods} \\
        \midrule
        Temporal-boundary allocation & $z_i=f_{\mathrm{time}}(t-i,w)$: deterministic allocation by token age, position, or local-window boundary & 2024--2025 & LoLCATs \citep{zhang2024lolcats}; Native Hybrid Attention \citep{du2025nha} \\
        Content-score allocation & $z_i\in\{\mathrm{KV}_{\mathrm{recent}},\mathrm{KV}_{\mathrm{salient}},S_{\mathrm{compressed}}\}$: recent tokens remain local; departing tokens enter salient historical KV or a compressed state according to self-recall error or self-saliency & 2025--2026 & LoLA \citep{mcdermott2025lola}; STILL \citep{meng2026still} \\
        Learned operation allocation & $z_i=\arg\max_{o\in\mathcal O}p_{\theta}(o\mid x_i,\mathcal H_i)$: a learned or search-derived policy assigns a Softmax or Linear/Recurrent operation & 2026 & NAtS-L \citep{deng2026natsl} \\
        \bottomrule
    \end{tabularx}
\end{table}

\subsubsection{Temporal-Boundary Allocation}
\label{subsubsec:hybrid-temporal-boundary}

Temporal-boundary methods make a deterministic allocation according to token age, position, or a local-window boundary. LoLCATs applies exact Softmax Attention to the most recent $w$ tokens and accumulates tokens that leave the window into a Linear Attention state; local tokens and the long-term state are then read through a jointly normalized expression \citep{zhang2024lolcats}. Native Hybrid Attention similarly preserves exact KV representations for recent tokens, updates older history into a fixed number of long-term slots, and applies one Softmax readout over the local KV and the slots \citep{du2025nha}. Both use a temporal boundary to determine memory lifecycle, but represent remote history with an associative matrix state and gated slots, respectively.

\subsubsection{Content-Score Allocation}
\label{subsubsec:hybrid-content-score}

Content-score methods further distinguish which tokens leaving the local window should remain explicitly addressable. Their allocation can be summarized as
\begin{equation}
    z_i
    \in
    \left\{
        \mathrm{KV}_{\mathrm{recent}},
        \mathrm{KV}_{\mathrm{salient}},
        S_{\mathrm{compressed}}
    \right\}.
    \label{eq:hybrid-content-allocation}
\end{equation}
All recent tokens first remain in local KV memory. When a token leaves the window, LoLA uses self-recall error to determine whether it should enter sparse historical KV or be compressed into a Linear Attention state \citep{mcdermott2025lola}. STILL performs a similar division using self-saliency: high-saliency history remains accessible through Sparse Attention, while the remaining information is written into a linear state \citep{meng2026still}. These designs extend a fixed temporal lifecycle into content-dependent allocation of memory fidelity.

\subsubsection{Learned Operation Allocation}
\label{subsubsec:hybrid-learned-operation}

NAtS-L does not only decide how a token should be stored after it leaves a local window. It searches, at token or chunk granularity, whether the unit should use a Softmax Attention or GDN operation:
\begin{equation}
    z_i
    =
    \arg\max_{o\in\mathcal O}
    p_{\theta}(o\mid x_i,\mathcal H_i),
    \qquad
    \mathcal O=\{\mathrm{Softmax},\mathrm{GDN}\}.
    \label{eq:hybrid-learned-operation}
\end{equation}
Tokens assigned to the Softmax operation retain explicit KV representations, whereas those assigned to GDN propagate through a recurrent state \citep{deng2026natsl}. Routing thus progresses from a predetermined time boundary and hand-designed content score to learned or search-derived assignment of sequence-mixing operations.

\subsection{Summary}
\label{subsec:hybrid-summary}

The preceding comparison shifts the question of efficient sequence modeling from whether one mechanism can fully replace Softmax Attention to how heterogeneous mechanisms can divide and coordinate contextual processing. Hybrid Architecture is therefore not an additional sequence mixer, but a compositional design paradigm in which explicit token retrieval, sparse access, local interaction, and recurrent propagation can be assigned to different structural units within the same model.

\noindent\textbf{Hybrid designs distribute complementary capabilities across progressively finer and more tightly coordinated granularities.}
Layer-wise methods arrange largely self-contained mixers across network depth. Head-wise methods place heterogeneous retrieval and state-propagation capabilities within a shared layer, Branch-wise methods coordinate comparatively complete memory paths at the readout or output interface, and Token-wise methods determine how individual tokens or chunks are represented or processed. These granularities are composable rather than mutually exclusive, allowing one architecture to coordinate complementary mechanisms across several structural axes.

\noindent\textbf{The central technical development is a shift from module coexistence toward deeper coordination of heterogeneous memory paths.}
Coarse Layer-wise composition primarily controls where each mechanism is used, whereas finer-grained designs can additionally couple memory representations, lifecycle decisions, readout rules, and output formation. Head-wise allocation divides representational and retrieval capacity inside a layer; Branch-wise fusion combines multiple contextual readouts or independently formed path outputs; and Token-wise allocation determines whether particular information remains explicitly addressable or is incorporated into a compressed recurrent state. The scope of complementarity therefore expands from arranging separate modules to jointly organizing how heterogeneous memories are formed, accessed, and combined.

% !TEX root = ../main.tex
\section{Attention Designs in Publicly Documented LLM Architectures: Evolution, Coordination, and Frontier Adoption}
\label{sec:attention-designs-in-publicly-documented-architectures}

The preceding sections examine attention and related sequence-mixing mechanisms at the method level. This section turns to their use in publicly documented autoregressive language-model backbones and asks three architecture-level questions: whether attention design is converging toward one dominant form, how heterogeneous mechanisms are combined within individual models, and which attention structures are represented among high-performing open-weight models.

Our first evidence set is a longitudinal inventory of 59 release-level architecture records spanning 14 major model lineages from 2022 through September~22, 2026. Models released together are grouped when they share the same language-model attention backbone, whereas separately released versions remain separate records; a documented change in the attention backbone also defines a separate record. The inventory is based primarily on official papers, technical reports, model cards, and released configurations. Although several included releases are natively multimodal, the unit of analysis remains the autoregressive language-model backbone. We include such models only when their causal language backbone is publicly documented and relevant to the evolution of attention design, and exclude vision or audio encoders, modality projectors, cross-modal interfaces, and other modality-specific components from the classification. The chapter therefore remains an analysis of LLM attention architectures rather than a survey of multimodal architectures. The complete inventory and source mapping are provided in Supplementary Table~\ref{tab:supplementary-architecture-inventory}.

Our second evidence set is a frozen comparison of high-performing open-weight models from the Artificial Analysis Intelligence Index v4.3.2, captured on September~22, 2026 \citep{artificialanalysis2026intelligenceindex,artificialanalysis2026methodology,artificialanalysis2026selectedopenweight}. It provides a cross-sectional view of attention designs near the open-weight performance frontier, while the longitudinal inventory characterizes architectural development over time. Both analyses are descriptive: the inventory is purposively curated rather than exhaustive or market-share weighted, and leaderboard scores depend on model scale, training, post-training, reasoning budget, and systems implementation in addition to attention design. We therefore use these evidence sets to characterize adoption and coexistence, not to infer that a particular attention mechanism causes higher model quality.

\begin{figure}[t]
    \centering
    \includegraphics[width=\textwidth]{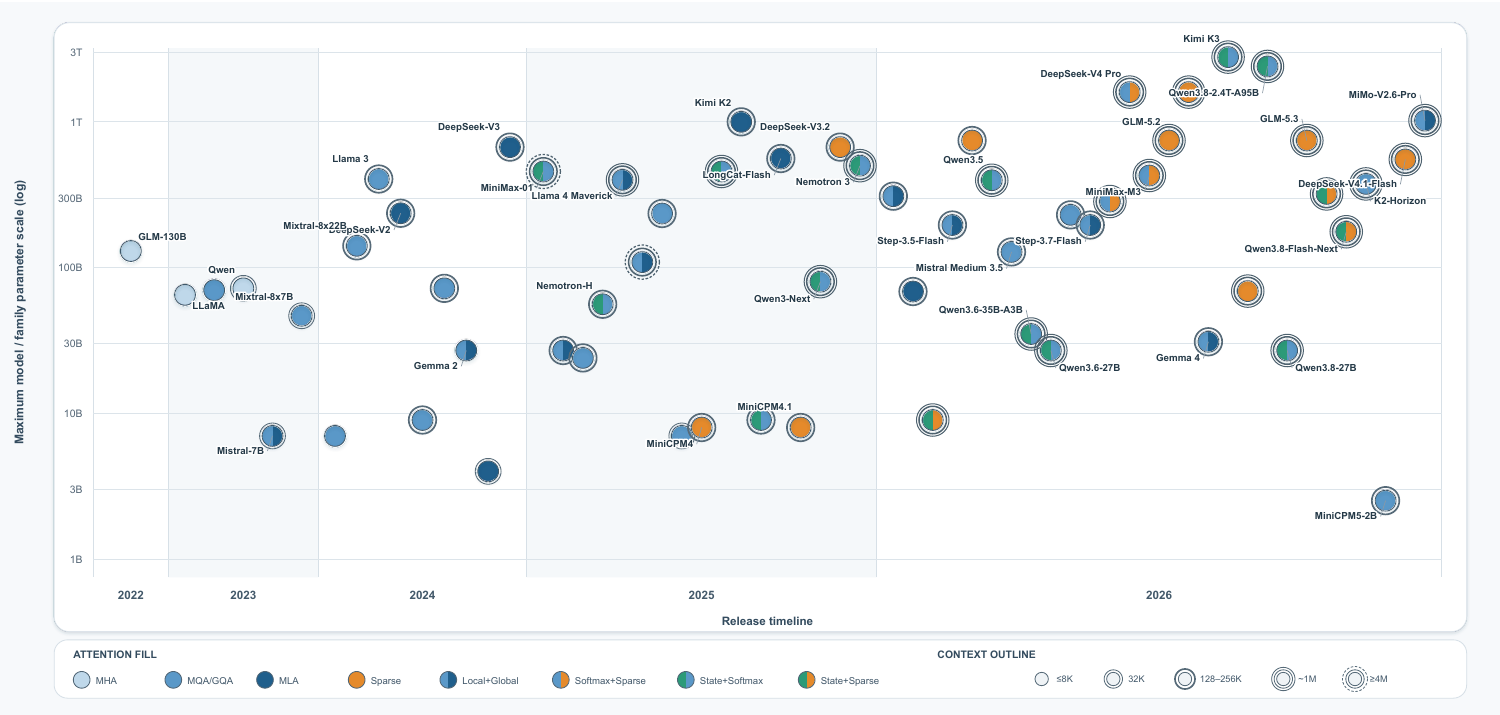}
    \caption{Attention designs across the classifiable architecture inventory. For natively multimodal systems, only the autoregressive language-model backbone is classified. Fill colors encode attention composition, split fills denote heterogeneous designs, and outlines indicate broad context-length tiers. The increasing concentration of split-fill nodes in recent years indicates the growing use of Hybrid architectures; the figure is descriptive and does not imply performance rankings or causal relationships.}
    \label{fig:publicly-documented-attention-architecture-landscape}
\end{figure}

Figure~\ref{fig:publicly-documented-attention-architecture-landscape} visualizes the 58 classifiable records in the longitudinal inventory; Kimi~k1.5 is omitted because its attention architecture is not separately disclosed. Family-level records with distinct scale-specific variants may appear as multiple nodes, so the figure is not a one-to-one count of release records. Node fills encode attention composition, outlines summarize context-length tiers, and labels are limited to representative releases to preserve readability.

\subsection{Diversification Rather Than Convergence in Attention Design}
\label{subsec:attention-cost-control-evolution}

\noindent\textbf{The longitudinal inventory shows diversification rather than convergence on one attention type.}
Dense MHA evolved toward MQA and GQA to reduce KV-head redundancy while preserving full token-level retrieval, and GQA remains a widely used backbone in recent Llama \citep{dubey2024llama3}, Qwen \citep{yang2024qwen2}, Mistral \citep{mistral2025small31}, MiniMax \citep{minimax2026m2}, MiniCPM \citep{openbmb2026minicpm5}, and K2-Horizon releases \citep{ifm2026k2horizon375b}. In parallel, MLA \citep{deepseek2024v2,meituan2025longcatflash,liu2026longcatflashlite} compresses per-token KV representations, local and sparse attention \citep{deepseek2025v32,zan2026longcatsparse} restrict explicit access, and recurrent mechanisms such as Gated DeltaNet \citep{qwen2025qwen3next}, Lightning Attention \citep{minimax2025m1}, KDA \citep{kimi2026k3}, and Mamba \citep{nvidia2025nemotron3} replace part of repeated token retrieval with fixed-size state. These approaches reduce different costs and retain different memory interfaces, so they continue to coexist rather than forming a single replacement sequence.

\noindent\textbf{Model lineages also change direction rather than following a monotonic progression.}
DeepSeek develops from MLA to MLA-based sparse retrieval and then to CSA/HCA and CSA2 \citep{deepseek2024v2,deepseek2025v32,deepseek2026v4,deepseek2026v41flash}; Qwen moves from MHA and GQA to recurrent-state architectures combined with full attention in Qwen3-Next, Qwen3.5, and Qwen3.6, and with sparse attention in Qwen3.8-Flash-Next \citep{bai2023qwen,qwen2025qwen3,qwen2025qwen3next,qwen2026qwen35,qwen2026qwen3635ba3b,qwen2026qwen3627b,qwen2026qwen38next}; MiniMax moves from Lightning/Softmax hybrids to full GQA and then to dense/sparse attention \citep{minimax2025m1,minimax2026m2,minimax2026m3}; and LongCat moves from MLA in LongCat-Flash and LongCat-Flash-Lite to LSA in LongCat-2.0 and LongCat-Flash-Lite-Sparse \citep{meituan2025longcatflash,liu2026longcatflashlite,zan2026longcatsparse,meituan2026longcat2}. The resulting landscape is therefore best understood as several persistent design routes---shared-KV dense attention, latent KV compression, restricted explicit access, and recurrent-state propagation---that are increasingly recombined in different models.

\subsection{Increasing Hybridization and Emerging Cross-Layer Reuse}
\label{subsec:coordination-of-heterogeneous-mechanisms}

\noindent\textbf{Hybrid composition becomes more common in the recent inventory, primarily through predetermined layer-wise schedules.}
Of the 59 release-level records, 58 disclose enough information for classification: 32 retain one principal attention or memory family, while 26 are classified as Hybrid. These are release-level records rather than 26 distinct architectural patterns. Table~\ref{tab:architecture-evolution-periods} shows that Hybrid records increase from none in 2022--2023 to 9 of 25 in 2024--2025 and 17 of 27 in 2026. The dominant forms are local/global explicit-attention schedules and recurrent-state/explicit-attention schedules, in which complementary capabilities are assigned to different layers and coordinated through the hidden-state stream.

\begin{table}[!htbp]
    \centering
    \small
    \setlength{\tabcolsep}{7pt}
    \renewcommand{\arraystretch}{1.18}
    \caption{Temporal distribution of Single-family, Hybrid, and cross-layer-coordinated architectures in the publicly documented model inventory. Single-family and Hybrid are mutually exclusive labels; cross-layer coordination is an overlapping attribute.}
    \label{tab:architecture-evolution-periods}
    \begin{tabularx}{\textwidth}{@{}
        >{\raggedright\arraybackslash}X
        >{\centering\arraybackslash}p{0.16\textwidth}
        >{\centering\arraybackslash}p{0.21\textwidth}
        >{\centering\arraybackslash}p{0.18\textwidth}
        >{\centering\arraybackslash}p{0.21\textwidth}@{}}
        \toprule
        \textbf{Period}
        & \textbf{Records}
        & \textbf{Single-family}
        & \textbf{Hybrid}
        & \textbf{Cross-layer} \\
        \midrule

        2022--2023
        & 6
        & 6 (100\%)
        & 0
        & 0 \\

        2024--2025
        & 25
        & 16 (64\%)
        & 9 (36\%)
        & 0 \\

        2026\textsuperscript{*}
        & 27
        & 10 (37\%)
        & 17 (63\%)
        & 5 (19\%) \\

        \bottomrule
    \end{tabularx}

    \vspace{3pt}
    \parbox{0.96\textwidth}{
        \footnotesize
        \textsuperscript{*}Records available through September~22, 2026. Percentages use the classifiable records within each period as the denominator. Kimi~k1.5 is excluded because its attention architecture is not separately documented. Cross-layer coordination should not be added to the Single-family and Hybrid columns. The five cross-layer records are GLM-5.2, GLM-5.3, DeepSeek-V4.1-Flash, LongCat-2.0, and LongCat-Flash-Lite-Sparse; the two GLM releases and the two LongCat releases each share a base architectural pattern. DeepSeek-V4.1-Flash is counted as Hybrid in the mutually exclusive Single-family/Hybrid classification and also carries the overlapping cross-layer attribute.
    }
\end{table}

\noindent\textbf{Cross-layer reuse introduces a distinct form of coordination by extending the depth-wise lifetime of memory and routing artifacts.}
Across the documented cases, GLM-5.2 and GLM-5.3 reuse retrieval indices or top-$k$ candidate decisions through IndexShare/IndexCache \citep{glm2026glm52,glm2026glm53,bai2026indexcache}; DeepSeek-V4.1-Flash additionally reuses selected global KV representations and indexer keys through the Full, Reindex, and Reuse modes of CSA2 \citep{deepseek2026v41flash}; and LongCat-2.0 and LongCat-Flash-Lite-Sparse use LSA's Cross-Layer Indexing to reuse one layer's selected token set across subsequent layers \citep{zan2026longcatsparse,meituan2026longcat2}. Hybrid composition describes where complementary mechanisms are placed; cross-layer reuse describes whether artifacts produced at one depth remain available to later layers. The latter is currently documented in five records, but it extends coordination from module placement to the depth-wise persistence of memory and routing artifacts.

\subsection{Attention Designs among High-Performing Open-Weight Models}
\label{subsec:top-ranked-open-weight-models}

\noindent\textbf{The frozen Artificial Analysis comparison provides a cross-sectional view of the attention structures represented near the open-weight performance frontier.}
Table~\ref{tab:top-ranked-open-weight-attention} reports the eleven high-performing open-weight endpoints retained in the September~22, 2026 snapshot. Reasoning settings such as ``max'' and ``xhigh'' are treated as endpoint configurations rather than separate architectures. The endpoint selection follows the Artificial Analysis Intelligence Index v4.3.2 snapshot, while release dates, parameter scales, and attention structures are based on the corresponding public model documentation and the source mapping in Supplementary Table~\ref{tab:supplementary-architecture-inventory} \citep{artificialanalysis2026intelligenceindex,artificialanalysis2026selectedopenweight}.

\begin{table}[!htbp]
    \centering
    \footnotesize
    \setlength{\tabcolsep}{3.5pt}
    \renewcommand{\arraystretch}{1.12}
    \caption{Attention structures of the eleven selected high-performing open-weight model endpoints in the Artificial Analysis snapshot captured on September~22, 2026. Rows are ordered by release date in reverse chronological order. Model size reports total parameters; disclosed active parameters are shown in parentheses \citep{artificialanalysis2026selectedopenweight}.}
    \label{tab:top-ranked-open-weight-attention}
    \begin{tabularx}{\textwidth}{@{}
        >{\raggedright\arraybackslash}p{0.19\textwidth}
        >{\centering\arraybackslash}p{0.085\textwidth}
        >{\centering\arraybackslash}p{0.16\textwidth}
        >{\raggedright\arraybackslash}X@{}}
        \toprule
        \textbf{Model endpoint} & \textbf{Release} & \textbf{Model size} & \textbf{Attention structure} \\
        \midrule
        MiMo-V2.6-Pro & 2026-09 & 1.02T (42B active) & 60 SWA-GQA + 10 global-GQA layers \citep{mimo2026v26} \\
        DeepSeek-V4.1-Flash (max) & 2026-09 & 552B (8B active prefill; 16B active decode) & CSA2 with SWA; cross-layer KV/index/top-$k$ reuse \\
        K2-Horizon-375B-A23B & 2026-09 & 375B (23B active) & full GQA \\
        GLM-5.3-Flash & 2026-08 & 320B (18B active) & 34 KDA + 11 compressed-indexer DSA layers \\
        GLM-5.3 (max) & 2026-08 & 744B (40B active) & MLA-based DSA; IndexShare/IndexCache \\
        Qwen3.8-Flash-Next & 2026-08 & 176B; 125B main (6B active) & 3 GDN : 1 QSA \\
        DeepSeek-V4 Pro 0813 (max) & 2026-08 & 1.6T (49B active) & CSA/HCA layer-wise hybrid \\
        Qwen3.8-2.4T-A95B & 2026-08 & 2.4T (95B active) & 3 Gated DeltaNet : 1 gated full-GQA layer \\
        Qwen3.8-27B (xhigh) & 2026-08 & 27B & 3 Gated DeltaNet : 1 gated full-GQA layer \citep{qwen2026qwen3827b} \\
        Kimi K3 (max) & 2026-07 & 2.8T (104B active) & 3 KDA : 1 Gated MLA \\
        MiniMax-M3 & 2026-06 & 428B (23B active) & 3 dense-attention + 57 MSA layers \\
        \bottomrule
    \end{tabularx}
\end{table}

\noindent\textbf{The frontier comparison remains architecturally heterogeneous.}
The selected models include full GQA, latent-compressed sparse attention, block-sparse attention, local/global or compressed dense/sparse hybrids, and recurrent-state layers combined with full, latent, or sparse explicit retrieval. Hybrid structures are prominent but not universal: K2-Horizon retains full GQA and GLM-5.3 uses a single DSA family, while the other endpoints combine distinct access or state-propagation routes. At the same time, every selected architecture retains an explicit token-retrieval path, indicating that recurrent state and sparsity currently complement rather than uniformly replace token-addressable attention. The table demonstrates coexistence at the performance frontier; its chronological ordering does not rank attention mechanisms or imply that any particular mechanism causes higher model quality.

\subsection{Summary}
\label{subsec:publicly-documented-attention-architecture-summary}

The longitudinal inventory and the frozen frontier comparison support the same conclusion: current LLM attention design is diversifying rather than converging. Hybrid records have become more common and are still organized mainly through fixed layer-wise composition, while cross-layer reuse remains a limited but conceptually distinct extension that increases the depth-wise persistence of selected memory and routing artifacts. High-performing open-weight models preserve this diversity and continue to combine efficient state propagation or sparse access with explicit token retrieval.

% !TEX root = ../main.tex
\section{Synthesis Across Mechanisms, Architectures, and Future Directions}
\label{sec:landscape-and-future-directions}

The preceding chapters reveal that the evolution of attention-centered sequence architectures can be understood, at least in part, as an expanding redesign of contextual memory rather than simply as a succession of attempts to replace one operator. Explicit token memories, recurrent associative states, structured state-space models, and hybrid architectures begin from different technical formulations, yet repeatedly confront a related set of design questions: which memory units are maintained, how new information updates them, which represented information is made available to a query, how that information is read, and how multiple readouts are coordinated. Across the literature reviewed in this survey, the design focus consequently expands from optimizing an isolated interaction rule toward organizing the representation, lifecycle, and use of contextual memory as a whole.

Building on this perspective, this section develops a three-level synthesis that traces the evolution of memory control from mechanism design to architectural coordination and future memory systems. At the \emph{mechanism level}, explicit-memory and recurrent-state methods begin from different memory representations and computational forms, yet both gradually
extend the scope of control from their respective core concerns to a broader set of functions spanning Memory Representation, Memory Update, Access, Readout, and Integration. The ranges of memory functions explicitly controlled by the two routes consequently show a growing degree of overlap. At the \emph{architecture level}, an important direction is to integrate memory mechanisms with complementary capabilities within a single model. This complementarity is currently realized primarily through predetermined layer-wise composition, while recent systems begin to extend coordination beyond module placement to the cross-layer sharing, reuse, and refresh of selected memory and routing artifacts. At the \emph{forward-looking level}, we further synthesize these mechanism- and architecture-level developments into a stateful multidimensional memory-routing hypothesis spanning temporal scope, network depth, substrate type, and representation resolution.

\subsection{Mechanism Level: Distinct Starting Points and Expanding Control Scopes}
\label{subsec:two-memory-routes}

At the mechanism level, the surveyed literature suggests two broad routes for organizing contextual memory. The \emph{explicit-memory route} retains an enumerable memory interface while reducing the cost of representing or accessing its units. The \emph{state-based route} compresses history into
recurrently maintained states while improving their capacity, update control, and recoverability. These routes represent different predominant memory forms and design pressures rather than mutually exclusive model classes.

\noindent\textbf{Explicit-memory route.}
As reviewed in Sections~\ref{sec:softmax-attention} and~\ref{sec:sparse-attention}, this route represents contextual memory through token KVs, shared or latent token representations, bounded slots, blocks, or lower-resolution summaries. Its primary emphasis lies in Memory Representation and Access: methods reduce redundancy across heads, channels, or layers, alter the resolution of stored history, restrict the candidate set exposed to a query, or reuse representations and retrieval decisions. Its control scope also extends to Memory Update when bounded or multiresolution memories merge, replace, or consolidate historical information; to Readout when routing scores, summary estimates, or modified attention maps affect the aggregation of eligible memory; and to Integration through head coordination, output gating, or the combination of multiple readout components. The route therefore retains an enumerable memory interface while extending control from representation and access efficiency to the broader lifecycle and use of the represented memory.

\noindent\textbf{State-based route.}
As reviewed in Sections~\ref{sec:linear-attention} and~\ref{sec:state-space-models}, this route replaces a growing list of token-wise memories with one or more recurrently maintained distributed states. Its primary emphasis lies in Memory Representation and Memory Update: methods determine how historical information is compressed, accumulated, retained, corrected, overwritten, or selectively propagated. Its control scope expands through additional rows, slots, partitions, temporal summaries, and state groups that differentiate memory capacity and temporal resolution; selection among these units makes Access more explicit; conditioned, multi-channel, or multi-state projections enrich Readout; and gating or coordination among completed state readouts makes Integration a direct design consideration. The route therefore extends from constructing and updating a compressed state toward organizing how multiple state units are represented, maintained, exposed, read, and coordinated.

\begin{figure}[H]
    \centering
    \includegraphics[width=\linewidth]{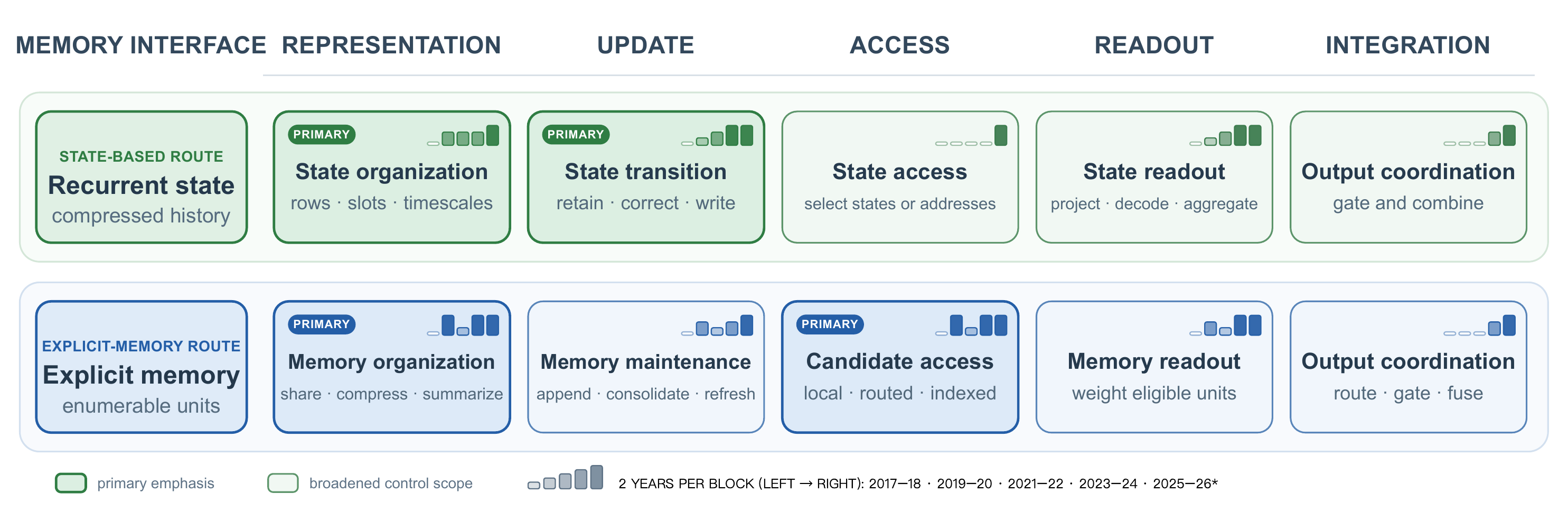}
    \caption{Mechanism-level synthesis of the expanding control scopes of explicit-memory and state-based methods. For the five functional dimensions, the upper-right mini-bars represent 2017--2018, 2019--2020, 2021--2022, 2023--2024, and 2025--2026 from left to right; taller and darker bars indicate greater qualitative activity among the representative methods discussed in Sections~\ref{sec:softmax-attention}--\ref{sec:state-space-models}. The 2026 coverage extends through September.}
    \label{fig:parallel-memory-routes}
\end{figure}

Figure~\ref{fig:parallel-memory-routes} summarizes the distinct starting points and expanding control scopes of the two routes. Darker cells indicate each route's primary emphasis, whereas lighter cells mark functions that become more explicit in subsequent designs. The columns are functional dimensions rather than a mandatory execution sequence.

\begin{insightbox}{Synthesis 1: Growing Overlap in the Scope of Memory Control}
Explicit-memory and state-based methods retain different memory interfaces and computational forms, yet both expand from their respective initial emphases toward a broader and increasingly overlapping set of memory functions. The convergence lies in the scope of explicit design control across Memory Representation, Memory Update, Access, Readout, and Integration, rather than in a shared memory substrate or computational mechanism.
\end{insightbox}

This growing overlap in control scope provides the mechanism-level basis for the architecture-level development examined next. As Representation, Update, Access, Readout, and Integration become more explicit and coordinated design variables, hybrid architectures can combine complementary memory capabilities across layers, heads, branches, and tokens rather than requiring one mechanism to satisfy every quality--cost trade-off. The next subsection examines how this complementarity is realized in publicly documented LLM architectures and how coordination begins to extend from module placement to the cross-layer lifecycle of selected memory and routing artifacts.

\subsection{Architecture-Level Synthesis: From Complementary Placement to Memory-Artifact Lifecycles}
\label{subsec:unified-memory-logic}

The inventory analysis in Section~\ref{subsec:coordination-of-heterogeneous-mechanisms}, summarized in Table~\ref{tab:architecture-evolution-periods}, reveals two ways in which contextual memory is increasingly organized along network depth: the layer-wise composition of heterogeneous memory mechanisms and the cross-layer reuse of selected memory and routing artifacts. Layer-wise composition determines what form of memory processing occurs at each representational stage, whereas cross-layer reuse determines which artifacts produced at one stage remain available to later stages. Together, these developments bring the construction, maintenance, and reuse of contextual memory along network depth into the architecture-level design space.

Within the current sample, Hybrid records become more common and are organized mainly through predetermined layer-wise schedules. Explicit attention supports fine-grained query-dependent retrieval, Sparse Attention restricts the candidate budget, and recurrent Linear Attention or SSM layers propagate compressed history with bounded or slowly growing decoding state. Coordinated through the hidden-state stream, these mechanisms apply different memory operations as a representation moves through the network. Layer-wise composition therefore organizes complementary memory capabilities across depth, allowing different representational stages to contribute different forms of retrieval, compression, and state propagation.

Layer-wise composition changes the memory processing performed at each depth; cross-layer reuse changes the lifetime of the resulting artifacts. In the documented cases, selected KV representations, index states, or candidate decisions remain available across depth instead of being independently reconstructed at every layer. This introduces a distinction between \emph{capability placement}---which memory functions operate at different depths---and \emph{artifact-lifecycle coordination}---which memory or routing artifacts persist across layers and when they are reused or refreshed. In the current inventory, layer-wise Hybrid composition is already common, whereas explicit cross-layer artifact reuse is confined to five release-level records corresponding to three architectural patterns documented across GLM \citep{glm2026glm52,glm2026glm53,bai2026indexcache}, DeepSeek \citep{deepseek2026v41flash}, and LongCat \citep{zan2026longcatsparse,meituan2026longcat2} systems.

\begin{insightbox}{Synthesis 2: Network Depth as an Emerging Dimension of Memory Organization}
Layer-wise composition determines what memory processing occurs at each depth, while cross-layer reuse determines what memory artifacts persist across depth. Together, they make network depth an emerging dimension along which contextual memory is constructed and managed.
\end{insightbox}

Viewed together, these developments shift the architecture-level question from which mechanism each layer should use to how memory should be organized across the network as a whole. As representations move through successive layers, memory may be retrieved, compressed, transformed, or propagated by different mechanisms, while selected intermediate artifacts may remain available to later stages. Network depth thus becomes an additional coordinate for organizing where contextual memory is produced, transformed, maintained, reused, and refreshed. This perspective motivates the forward-looking multidimensional memory organization considered next.

\subsection{Forward-Looking Hypothesis: Stateful Multidimensional Memory Routing}
\label{subsec:stateful-memory-moe}

The mechanism- and architecture-level syntheses above motivate a broader design space for contextual memory. At the mechanism level, explicit-memory and state-based methods increasingly control overlapping sets of memory functions. At the architecture level, complementary capabilities are commonly distributed across layers, while a small number of recent systems begin to extend the lifecycle of selected memory and routing artifacts across network depth. Taken together, these developments suggest that persistent memory could be organized and controlled jointly across multiple architectural coordinates, rather than being defined only by temporal position or by the mechanism of one layer.

\subsubsection{Multidimensional Memory as an Address Space}

We represent the admissible addresses of persistent memory by four coordinates:
\begin{equation}
    \mathcal{A}_t
    =
    \mathcal{T}_t
    \times
    \mathcal{L}
    \times
    \mathcal{S}
    \times
    \mathcal{G},
    \qquad
    \mathcal{M}_t
    =
    \left\{
        M_t[a]
        \,\middle|\,
        a\in\mathcal{A}_t
    \right\}
    \in
    \mathfrak{M}_{\rho},
    \label{eq:multidimensional-memory}
\end{equation}
where an address $a=(\tau,\ell,s,g)$ identifies the temporal scope, network depth, memory substrate, and representation granularity of a memory unit, and $M_t[a]$ denotes the memory unit stored at address $a$. The temporal coordinate $\tau\in\mathcal{T}_t$ may refer to a token, segment, or longer historical interval. The depth coordinate $\ell\in\mathcal{L}$ identifies the layer or representational stage at which the unit is produced or maintained. The substrate coordinate $s\in\mathcal{S}$ distinguishes forms such as explicit KV, an associative state, or a structured recurrent state. The granularity coordinate $g\in\mathcal{G}$ distinguishes fine-resolution representations from coarser summaries of the same or related temporal scope.

These coordinates describe an address space for memory organization rather than requiring one physically materialized Cartesian tensor. A model could instantiate only a selected set of addresses and could associate different Update and Readout rules with different substrates. The same historical interval might, for example, be represented by fine-grained token KVs, a block-level representation, or a compressed recurrent summary. Memories produced at different depths could likewise preserve information associated with different levels of representation and remain available to selected later computations.

Such an organization allows persistent memory units to assume differentiated roles. Fine-resolution explicit representations may support precise query-dependent retrieval, while compressed summaries or recurrent states may provide broader historical coverage at lower storage or access cost. Architecture design would consequently determine not only which memory forms are present, but also where they are produced, how long they persist, and at what granularity they remain available.

\subsubsection{Sparse Write and Sparse Read}

Uniformly updating and reading every admissible memory unit would diminish the efficiency gained from differentiated memory roles. We therefore distinguish a selected write-address set $\mathcal{W}_t$ from the eligible memory view $\mathcal{C}_t$ exposed to the current query. Sparse Write can be expressed as
\begin{equation}
    \mathcal{W}_t
    \subseteq
    \mathcal{A}_t,
    \qquad
    \mathcal{M}_t^{+}
    =
    \operatorname{Update}_{\rho}
    \left(
        \mathcal{M}_t^{-},
        x_t;
        \mathcal{W}_t
    \right),
    \label{eq:sparse-memory-write}
\end{equation}
where $\mathcal{M}_t^{-}$ and $\mathcal{M}_t^{+}$ denote the memory before and after the current update. The selected addresses may correspond to particular temporal scopes, network depths, substrates, or representation granularities. The Update rule remains specific to each selected substrate: it may append an explicit representation, modify a recurrent state, merge information into a summary, or apply a substrate-specific retention or correction rule. Sparse Write therefore controls where incoming information is maintained without requiring heterogeneous memory units to share one update mechanism.

Following the analytical framework in Section~\ref{sec:unified-memory-centric-view}, the complete Sparse Read pathway is written as
\begin{equation}
    \mathcal{C}_t
    =
    \operatorname{Access}_{\rho}
    \left(
        q_t,
        \widehat{\mathcal{M}}_t
    \right),
    \qquad
    r_t
    =
    \operatorname{Readout}_{\rho}
    \left(
        q_t,
        \mathcal{C}_t
    \right),
    \qquad
    o_t
    =
    \operatorname{Integration}_{\rho}
    \left(
        r_t;
        x_t
    \right).
    \label{eq:sparse-memory-read}
\end{equation}
Here, $\widehat{\mathcal{M}}_t$ denotes the memory visible to the read path under the mechanism's read--write convention, and $\mathcal{C}_t$ is the eligible memory view produced by Access. It may contain selected units from different temporal scopes, depths, substrates, and granularities, together with the local metadata required by their Readout rules. Readout converts this eligible view into contextual information, while Integration transforms or coordinates one or more completed readouts into the module output.

Sparse Write and Sparse Read are coupled through Memory Representation. Write routing determines which memory addresses receive new information and therefore constrains what can remain available to later queries. Read routing allocates retrieval over the resulting memory organization. Writing to units that are rarely accessed wastes storage and update effort, whereas repeatedly accessing units that receive insufficient or unsuitable updates weakens retrieval quality. Their budgets and routing policies may benefit from joint coordination over the same multidimensional address space.

\subsubsection{Stateful Routing of Memory Lifecycles}

This organization resembles a multidimensional Memory-MoE in which persistent memory units assume different functional roles and routing determines which units participate in writing and reading. The central distinction from a conventional feed-forward MoE is persistence. Conventional expert routing primarily selects the computation applied to the current representation; memory routing would additionally determine which information is retained, modified, and made available to future queries. A write-routing decision could therefore affect both the current update and the future state of the memory system.

%Within the five-dimensional framework, Memory Representation defines the available address space and memory substrates; Memory Update determines how the units selected by $\mathcal{W}_t$ evolve; Access constructs the eligible view $\mathcal{C}_t$ for the current query; Readout extracts information according to the rules of the selected substrates; and Integration transforms or coordinates the completed readouts into the module output. The hypothesis therefore concerns the coordinated lifecycle of heterogeneous memory rather than the replacement of these functions by one universal routing operator.

\begin{insightbox}{Hypothesis: Stateful Multidimensional Memory Routing}
A future architecture may organize persistent memory jointly across temporal scope, network depth, substrate type, and representation granularity. Separate write and read routing would determine which memory addresses preserve incoming information and which represented units contribute to a query. Because write decisions change the information available to later computation, routing would coordinate not only current resource allocation, but also the future lifecycle of contextual memory.
\end{insightbox}

Such a system would introduce several coupled design questions. Memory units would require compatible interfaces across representational levels, while routing would need to balance memory occupancy, update frequency, retrieval fidelity, interference, and hardware efficiency. Persistent write decisions would also create delayed dependencies between information stored at one step and its value to later queries. Feature alignment, memory ownership, staleness control, recovery from harmful writes, load balancing, and long-horizon credit assignment would consequently become central considerations. The hypothesis would ultimately need to be evaluated by whether multidimensional memory routing improves the practical quality--efficiency frontier under matched memory, computation, and routing budgets.

\subsubsection{Concrete and Testable Research Directions}
\label{subsubsec:concrete-research-directions}

Building on the hypothesis above, we outline four feasible directions that progressively expand the scope of memory control. They begin with choosing suitable representations for heterogeneous context, then separate large-scale memory construction from query-time retrieval, extend memory control to incremental lifecycle management, and finally coordinate memory and computation according to the current task state. Existing systems provide partial evidence for each step, while the proposals below identify concrete extensions that could be implemented and tested without realizing the complete multidimensional architecture at once.

\paragraph{Workload-aware differentiated memory representation.}
Different input regions need not share a single memory representation. Long contexts may combine large documents, code repositories, tool specifications, recent interactions, and execution logs, which differ in stability, structural hierarchy, update frequency, and required retrieval precision. Existing systems already expose parts of this design space: Qwen Sparse Attention uses compressed micro-block addresses to select original-token content, HCA and CSA combine coarser global representations with fine-grained local access, and Engram assigns stable patterns to deterministic conditional lookup rather than repeatedly reconstructing them through neural computation \citep{qwen2026qwen38next,qwen2026qwen38flashnext,deepseek2026v4,cheng2026engram}. A concrete next step is a typed memory constructor that maps each input region to one of a small number of representations---exact, compressed, indexed, or recurrent---using observable attributes such as source type, expected stability, hierarchy, and precision requirements. A repository, for example, could retain a coarse repository summary, file- and symbol-level indexes, and exact KVs only for active code regions. Restricting the initial routing space to these representation choices under an explicit capacity budget would make the substrate and granularity coordinates directly testable.

\paragraph{Asymmetric memory construction and retrieval.}
Large-scale memory construction and query-time retrieval need not use the same computation. Input-heavy workloads may provide extensive background context but require only short or incremental outputs, making uniform processing across prefill and decoding inefficient. DeepSeek-V4.1-Flash offers direct evidence for this asymmetry: its Causal Encoder--Decoder activates fewer parameters during prefill than decoding and combines CSA2 cross-layer KV/index reuse with aggressive cache compression; QSA similarly lowers long-input scanning cost by compressing routing keys before selecting original-token KVs \citep{deepseek2026v41flash,qwen2026qwen38flashnext}. A practical extension is a multiresolution construction pipeline in which a lightweight prefill path produces summaries and routing addresses while retaining exact payloads only for precision-sensitive regions. During decoding, each query would first inspect coarse memory and escalate to finer blocks or original tokens only when the coarse readout is insufficient. This coarse-to-fine interface would make prefill cost, persistent-memory size, and retrieval precision separately controllable, extending differentiated Representation into Memory Update, Access, and Readout.

\paragraph{Incremental memory lifecycle and reusable artifacts.}
Memory maintenance should scale with the amount of changed context rather than with total context length. A document collection may add one section, a repository may modify one function, and an execution trace may append only a few observations; rebuilding all associated memory would repeat work over unchanged content. Cross-layer systems provide initial evidence that memory artifacts can persist: IndexCache reuses retrieval indices, CSA2 defines Full, Reindex, and Reuse modes for KV and index artifacts, and YOIO/CLSA shares one routing decision across multiple downstream layers \citep{bai2026indexcache,deepseek2026v41flash,sun2026yoio}. The next step is conditional lifecycle control. At each layer or context update, a lightweight controller could choose \emph{reuse}, \emph{refresh}, \emph{transform}, or \emph{invalidate} for each representation or routing artifact. Reuse could apply when hidden-state drift and candidate-set change remain small; refresh when query--index agreement falls below a threshold; transformation when an artifact must be adapted to a new representation space; and invalidation after a source-version change. For a code edit, only the modified function, its file summary, and affected index entries would be refreshed, while unrelated repository memory remains reusable. This would turn temporal scope and depth from static address labels into dynamic coordinates of memory maintenance.

\paragraph{Task-conditioned joint memory and computation routing.}
Memory and computation policies should ultimately adapt together to the current task state. Broad planning may favor repository- or document-level summaries, precise editing may require local exact tokens, execution may emphasize recent logs and tool outputs, and verification may require multi-source evidence. Elastic Attention and depth-adaptive Hybrid Architecture proposals already vary the allocation of attention and recurrent paths according to input or depth, while Mixture-of-Depths dynamically allocates computation under a fixed budget \citep{tang2026elasticattention,shi2026implicithybrids,raposo2024mixturedepths}. A tractable realization could begin with a finite menu of configurations rather than an unrestricted controller. Each configuration would specify a memory substrate, retrieval granularity, attention mode, computation-depth budget, and write-back action; a compact task-state representation would then select among them. Planning could invoke broad coarse-grained retrieval, editing fine-grained local memory, and verification multi-source Readout and Integration. The policy could first be distilled from task annotations or fixed heuristics and later optimized under memory and computation constraints. By jointly selecting how memory is represented, updated, accessed, read, and integrated, this direction provides a constrained path toward complete stateful multidimensional memory routing.

% !TEX root = ../main.tex
\section{Conclusion}
\label{sec:conclusion}

This survey has examined four mechanism-centered research lines---Softmax Attention, Sparse Attention, Linear Attention, and State Space Models---together with Hybrid Architecture that organizes these mechanisms within larger models. Rather than treating these lines only as alternative approaches to reducing computational complexity, we have analyzed how they balance fine-grained addressability, compressed historical coverage, memory capacity, update control, and practical computation.

To provide a comparable basis for this analysis, we introduced a five-dimensional memory-centric lens: \textbf{Memory Representation}, \textbf{Memory Update}, \textbf{Access}, \textbf{Readout}, and \textbf{Integration}. These dimensions identify what historical information remains represented, how the represented memory changes, what becomes eligible for a query, how eligible memory is read, and how one or more completed readouts are transformed or coordinated into the module output. They are analytical roles rather than a universal computational factorization: one operation may affect several dimensions, and the chapter-level research lines remain historically and technically overlapping. Within this view, Softmax Attention reduces redundancy or historical resolution while retaining a separately addressable explicit-memory interface and normalized query--key Readout; Sparse Attention controls which explicit memory units become eligible for a query; Linear Attention develops recurrent associative states through more precise update, capacity, and temporal organization; SSMs structure compressed history through learned and input-conditioned dynamics; and Hybrid Architectures allocate complementary capabilities across layers, heads, branches, and tokens.

At the mechanism level, explicit-memory and recurrent-state methods retain different predominant memory interfaces, but the ranges of memory functions they explicitly control increasingly overlap. Explicit-memory methods extend beyond representation and access efficiency toward bounded update, readout modulation, and output coordination. State-based methods extend beyond state representation and recurrent update toward multiple memory units, selective access, richer readout, and coordinated integration. The shared development is therefore not convergence on one memory substrate or operator, but an expansion in the scope over which memory representation, update, access, readout, and integration are jointly considered.

At the architecture level, our curated inventory of 59 release-level architecture records spanning 14 major model lineages---drawn primarily from text-centered LLMs and, where relevant, from the autoregressive language backbones of natively multimodal models---shows continued coexistence rather than a universal replacement path. Single-family GQA, MLA, and Sparse Attention backbones remain in use, while Hybrid designs increasingly combine complementary memory capabilities. Within the documented Hybrid architectures, predetermined layer-wise composition remains the principal organizational pattern. A smaller set of recent Sparse Attention systems additionally extends coordination across network depth by sharing or reusing selected KV representations, index states, or candidate decisions. This cross-layer reuse is an additional architectural attribute rather than a subtype of Hybrid composition, and the current evidence remains concentrated in a small number of architecture records.

Taken together, the mechanism-level review and architecture-level analyses also show why attention-centered architectures should not be evaluated through asymptotic complexity alone. Explicit memories retain fine-grained addressability but incur storage and data-movement costs; sparse retrieval depends on index quality, candidate recall, and irregular execution; recurrent states face capacity, interference, and recoverability limits; and heterogeneous designs introduce placement, routing, synchronization, and kernel-design challenges. Meaningful comparison should therefore consider model quality, persistent memory, prefill and decoding cost, retrieval behavior, update stability, routing overhead, and realized hardware utilization under matched budgets and implementation conditions.

The mechanism-level and architecture-level syntheses jointly motivate a forward-looking hypothesis in which persistent contextual memory is organized across temporal scope, network depth, substrate type, and representation granularity. Layer-wise composition makes depth a coordinate of capability placement, while emerging cross-layer reuse suggests that depth may also become a coordinate of memory- and routing-artifact persistence. Coordinated Sparse Write and Sparse Read would determine which memory addresses receive information and which represented units contribute to a query, thereby controlling not only current computation but also the future lifecycle of memory. This remains a design hypothesis whose value depends on whether such organization improves the practical quality--efficiency frontier without introducing prohibitive routing, optimization, memory-management, or hardware costs. The memory-centric lens developed in this survey provides a common vocabulary for relating historically distinct research lines and formulating such testable questions. Overall, efficient sequence architecture design is increasingly concerned with the joint organization, lifecycle, and selective use of contextual memory rather than the optimization of an isolated attention operator.

\clearpage
\appendix
% !TEX root = ../main.tex
\section{Supplementary Materials: Publicly Documented Architecture Inventory}
\label{sec:supplementary-architecture-inventory}

This appendix reports the 59 release-level architecture records used in Section~\ref{sec:attention-designs-in-publicly-documented-architectures}. \emph{Release record} is the unit counted in Table~\ref{tab:architecture-evolution-periods}: models released together are grouped when they share the same language-model sequence-mixing architecture, whereas separately released versions remain separate records. \emph{Model lineage} identifies the broader model family used to obtain the 14-lineage count. The \emph{Attention / mixer composition} column gives only the principal sequence-mixing mechanisms and, where useful, their layer ratio. A record is labeled Hybrid when its language backbone instantiates two or more distinguishable sequence-mixing or access regimes in separate layers, branches, heads, or token paths; multiple support components jointly forming a single sparse-attention candidate set do not by themselves constitute Hybrid composition. Records with insufficient architectural disclosure are marked Undisclosed and excluded from the Non-Hybrid/Hybrid percentages. Cross-layer reuse is reported separately in Table~\ref{tab:architecture-evolution-periods} and the accompanying discussion. For multimodal systems, only the autoregressive language-model backbone is classified. The inventory is purposively curated rather than exhaustive or market-share weighted.

\setcounter{table}{0}
\renewcommand{\thetable}{S\arabic{table}}
\setlength{\LTleft}{0pt}
\setlength{\LTright}{\fill}
\setlength{\tabcolsep}{2.2pt}
\renewcommand{\arraystretch}{1.12}
\footnotesize
\begin{longtable}{@{}
    >{\raggedright\arraybackslash}p{0.065\textwidth}
    >{\raggedright\arraybackslash}p{0.085\textwidth}
    >{\raggedright\arraybackslash}p{0.19\textwidth}
    >{\raggedright\arraybackslash}p{0.325\textwidth}
    >{\centering\arraybackslash}p{0.09\textwidth}
    >{\raggedright\arraybackslash}p{0.13\textwidth}@{}}
\caption{The 59 publicly documented release-level architecture records analyzed in Section~\ref{sec:attention-designs-in-publicly-documented-architectures}, with the lineage and Hybrid labels used to derive Table~\ref{tab:architecture-evolution-periods}.}
\label{tab:supplementary-architecture-inventory}\\
\toprule
\textbf{Release} & \textbf{Model lineage} & \textbf{Release record} & \textbf{Attention / mixer composition} & \textbf{Hybrid} & \textbf{Source} \\
\midrule
\endfirsthead

\multicolumn{6}{l}{\footnotesize\tablename~\ref{tab:supplementary-architecture-inventory} continued from the previous page}\\
\toprule
\textbf{Release} & \textbf{Model lineage} & \textbf{Release record} & \textbf{Attention / mixer composition} & \textbf{Hybrid} & \textbf{Source} \\
\midrule
\endhead

\midrule
\multicolumn{6}{r}{\footnotesize Continued on the next page}\\
\endfoot

\bottomrule
\endlastfoot

2026-09 & MiMo & MiMo-V2.6-Pro-RL & SWA-GQA + global GQA (6:1) & Yes & \citep{mimo2026v26} \\

2026-09 & DeepSeek & DeepSeek-V4.1-Flash & SWA + CSA2 & Yes & \citep{deepseek2026v41flash} \\

2026-09 & MiniCPM & MiniCPM5-2B family & Full GQA & No & \citep{openbmb2026minicpm5} \\

2026-09 & K2 & K2-Horizon-375B-A23B & Full GQA & No & \citep{ifm2026k2horizon375b} \\

2026-08 & Qwen & Qwen3.8-2.4T-A95B & Gated DeltaNet + full GQA (3:1) & Yes & \href{https://huggingface.co/Qwen/Qwen3.8-2.4T-A95B}{Model card}; \href{https://huggingface.co/Qwen/Qwen3.8-2.4T-A95B/blob/main/config.json}{config} \\

2026-08 & Qwen & Qwen3.8-27B & Gated DeltaNet + gated full GQA (3:1) & Yes & \citep{qwen2026qwen3827b} \\

2026-08 & Qwen & Qwen3.8-Flash-Next & Gated DeltaNet + QSA (3:1) & Yes & \citep{qwen2026qwen38next} \\

2026-08 & GLM & GLM-5.3 & MLA-based DSA & No & \citep{glm2026glm53,bai2026indexcache} \\

2026-08 & GLM & GLM-5.3-Flash & KDA + compressed-indexer DSA (approximately 3:1) & Yes & \citep{glm2026glm53flash} \\

2026-08 & LongCat & LongCat-Flash-Lite-Sparse & LSA & No & \citep{zan2026longcatsparse} \\

2026-07 & Kimi & Kimi K3 & KDA + Gated MLA & Yes & \citep{kimi2026k3,chen2026attnres} \\

2026-07 & Gemma & Gemma 4 & SWA + global full Attention (4:1 or 5:1) & Yes & \citep{gemma2026gemma4} \\

2026-06 & LongCat & LongCat-2.0 & LSA & No & \citep{zan2026longcatsparse,meituan2026longcat2} \\

2026-06 & MiniMax & MiniMax-M3 & Dense Attention + MSA & Yes & \citep{minimax2026m3,minimax2026msa} \\

2026-06 & DeepSeek & DeepSeek-V4 & CSA + HCA (compressed dense Attention + SWA) & Yes & \citep{deepseek2026v4} \\

2026-06 & GLM & GLM-5.2 & MLA-based DSA & No & \citep{glm2026glm52,bai2026indexcache} \\

2026-05 & Step & Step-3.7-Flash & Full GQA + SWA-GQA (1:3) & Yes & \href{https://static.stepfun.com/blog/step-3.7-flash/}{Model page}; \href{https://huggingface.co/stepfun-ai/Step-3.7-Flash/blob/main/config.json}{config} \\

2026-05 & MiniMax & MiniMax-M2 & Full GQA & No & \citep{minimax2026m2} \\

2026-04 & Qwen & Qwen3.6-27B & Gated DeltaNet + gated full Attention (3:1) & Yes & \citep{qwen2026qwen3627b} \\

2026-04 & Qwen & Qwen3.6-35B-A3B & Gated DeltaNet + gated full Attention (3:1) & Yes & \citep{qwen2026qwen3635ba3b} \\

2026-04 & Mistral & Mistral Medium 3.5-128B & Full GQA & No & \citep{mistral2026medium35} \\

2026-02 & Step & Step-3.5-Flash & Full GQA + SWA-GQA (1:3) & Yes & \href{https://arxiv.org/abs/2602.10604}{Tech. report}; \href{https://huggingface.co/stepfun-ai/Step-3.5-Flash/blob/main/config.json}{config} \\

2026-02 & Qwen & Qwen3.5 & Gated DeltaNet + gated full Attention (3:1) & Yes & \citep{qwen2026qwen35} \\

2026-02 & MiniCPM & MiniCPM-SALA & Lightning Attention + InfLLM-v2 Sparse Attention (3:1) & Yes & \citep{minicpm2026sala} \\

2026-02 & GLM & GLM-5 & MLA-based DSA & No & \citep{glm2026glm5} \\

2026-01 & LongCat & LongCat-Flash-Lite & MLA & No & \citep{liu2026longcatflashlite} \\

2026-01 & MiMo & MiMo-V2-Flash & SWA + global Attention (5:1) & Yes & \citep{mimo2026v2flash} \\

2025-12 & Nemotron & Nemotron 3 & Mamba + full Attention + MLP-only blocks & Yes & \citep{nvidia2025nemotron3} \\

2025-12 & DeepSeek & DeepSeek-V3.2 & MLA-based DSA & No & \citep{deepseek2025v32} \\

2025-09 & LongCat & LongCat-Flash family & MLA & No & \citep{meituan2025longcatflash} \\

2025-09 & MiniCPM & MiniCPM4.1 & InfLLM-v2 Sparse GQA & No & \citep{openbmb2025minicpm41,zhao2025infllmv2} \\

2025-09 & Qwen & Qwen3-Next & Gated DeltaNet + gated full Attention (3:1) & Yes & \citep{qwen2025qwen3next,yang2024gdn,qiu2025gatedattention} \\

2025-08 & Nemotron & Nemotron Nano 2 & Mamba-2 + full Attention & Yes & \citep{nvidia2025nemotronnano2} \\

2025-07 & Kimi & Kimi K2 & MLA & No & \citep{kimi2025k2} \\

2025-06 & MiniCPM & MiniCPM4 & InfLLM-v2 Sparse GQA & No & \citep{minicpm2025minicpm4,zhao2025infllmv2} \\

2025-06 & MiniMax & MiniMax-M1 & Lightning Attention + Softmax Attention (7:1) & Yes & \citep{minimax2025m1} \\

2025-05 & Qwen & Qwen3 & Full GQA & No & \citep{qwen2025qwen3} \\

2025-05 & MiMo & MiMo-7B & Full GQA & No & \citep{mimo2025mimo7b} \\

2025-04 & Llama & Llama 4 Scout / Maverick & RoPE chunked GQA + NoPE full GQA (3:1) & Yes & \citep{meta2025llama4} \\

2025-04 & Nemotron & Nemotron-H & Mamba + full Attention + MLP-only blocks & Yes & \citep{nvidia2025nemotronh} \\

2025-03 & Mistral & Mistral Small 3.1-24B & Full GQA & No & \citep{mistral2025small31} \\

2025-03 & Gemma & Gemma 3 & Local Attention + global Attention (5:1) & Yes & \citep{gemma2025gemma3} \\

2025-01 & MiniMax & MiniMax-01 / Text-01 & Lightning Attention + Softmax Attention (7:1) & Yes & \citep{minimax2025text01} \\

2025-01 & Kimi & Kimi k1.5 & Undisclosed & Undisclosed & \citep{kimi2025k15} \\

2024-12 & DeepSeek & DeepSeek-V3 & MLA & No & \citep{deepseek2024v3} \\

2024-09 & MiniCPM & MiniCPM3-4B & MLA & No & \citep{openbmb2024minicpm3} \\

2024-08 & Gemma & Gemma 2 & Local Attention + global Attention (1:1) & Yes & \citep{gemma2024gemma2} \\

2024-07 & Qwen & Qwen2 & Full GQA & No & \citep{yang2024qwen2} \\

2024-06 & GLM & GLM-4 family & Full GQA & No & \citep{glm2024chatglm} \\

2024-05 & DeepSeek & DeepSeek-V2 & MLA & No & \citep{deepseek2024v2} \\

2024-04 & Mistral & Mixtral-8x22B-v0.1 & Full GQA & No & \citep{mistral2024mixtral8x22b} \\

2024-04 & Llama & Llama 3 family & Full GQA & No & \citep{dubey2024llama3} \\

2024-03 & Gemma & Gemma 1 & MQA / MHA (scale-dependent) & No & \citep{gemma2024gemma1} \\

2023-12 & Mistral & Mixtral-8x7B-v0.1 & Full GQA & No & \citep{mistral2023mixtral8x7b} \\

2023-09 & Qwen & Qwen & Full MHA & No & \citep{bai2023qwen} \\

2023-09 & Mistral & Mistral-7B-v0.1 & SWA-GQA & No & \citep{mistral2023mistral7b} \\

2023-07 & Llama & Llama 2 & MHA / GQA (scale-dependent) & No & \citep{touvron2023llama2} \\

2023-02 & Llama & LLaMA & Full MHA & No & \citep{touvron2023llama} \\

2022-10 & GLM & GLM-130B & Full MHA & No & \citep{zeng2022glm130b} \\

\end{longtable}
\normalsize

\clearpage
\printbibliography

\end{document}